\def\year{2021}\relax
\documentclass[letterpaper]{article} 
\usepackage{aaai}  
\usepackage{times}  
\usepackage{helvet} 
\usepackage{courier}  
\usepackage[hyphens]{url}  
\usepackage{graphicx} 
\usepackage{natbib}  
\usepackage{caption} 
\usepackage{booktabs}       
\usepackage{amsfonts}       
\usepackage{nicefrac}       
\usepackage{subcaption}
\usepackage{algorithm} 
\usepackage{algpseudocode} 
\everypar{\looseness=-1 } 
\usepackage{tikz}
\usepackage{pgfplotstable}
\usepackage{pgfplots}
\title{An Adversarial Approach for Explaining the Predictions of Deep Neural Networks}
\author{Arash Rahnama\\
	Modzy\\
	arash.rahnama@modzy.com\\
	\And
	Andrew Tseng\\
	Modzy\\
	andrew.tseng@modzy.com\\}

\begin{document}
	\maketitle
	
	\begin{abstract}
		Machine learning models have been successfully applied to a wide range of applications including computer vision, natural language processing, and speech recognition. A successful implementation of these models however, usually relies on deep neural networks (DNNs) which are treated as opaque black-box systems due to their incomprehensible complexity and intricate internal mechanism. In this work, we present a novel algorithm for explaining the predictions of a DNN using adversarial machine learning. Our approach identifies the relative importance of input features in relation to the predictions based on the behavior of an adversarial attack on the DNN. Our algorithm has the advantage of being fast, consistent, and easy to implement and interpret. We present our detailed analysis that demonstrates how the behavior of an adversarial attack, given a DNN and a task, stays consistent for any input test data point proving the generality of our approach. Our analysis enables us to produce consistent and efficient explanations. We illustrate the effectiveness of our approach by conducting experiments using a variety of DNNs, tasks, and datasets. Finally, we compare our work with other well-known techniques in the current literature.
	\end{abstract}
	
	\section{Introduction} \label{sec:intro}
	Explaining the outcomes of complex machine learning models is a prerequisite for establishing trust between the machines and users. As humans increasingly rely on DNNs to process large amounts of data and make decisions, it is crucial to develop solutions that can interpret the predictions of DNNs in a user-friendly manner. Explaining the outcomes of a model can help reduce bias and contribute to improvements in model design, performance, and accountability by providing beneficial insights into how models behave \cite{fidel2019explainability}. Consequently, the field of explainable artificial intelligence systems, XAI, has gained traction in recent years, where researchers from different disciplines have come together to define, design and evaluate explainable systems \cite{vstrumbelj2014explaining, datta2016algorithmic, mohseni2018survey}. The majority of current explainability algorithms for DNNs produce an explanation for a single input-output pair: an input data point fed into the DNN and the respective prediction made by the DNN. The algorithm usually finds the most important features in the input contributing the most to the model's predictions and selects those as explanations for the model's behavior \cite{alvarez2018robustness}. The majority of these algorithms find the important features using either a \emph{perturbation-based} approach or a \emph{saliency-based} approach \cite{lundberg2017unified}. The saliency-based approaches rely on gradients of the outputs in relation to the inputs to find the important features \cite{simonyan2013deep,selvaraju2017grad}. Perturbation-based methods on the other hand apply small local changes to the input, track the changes in the output, and find and rank the important input features \cite{ribeiro2016should, alvarez2017causal}.
	
	One main problem with current state-of-the-art explainability tools is their reliance on a large set of hyper-parameters. This leads to local instability of explanations and can negatively affect the user's experience \cite{alvarez2018robustness}. An explainability algorithm should satisfy 3 properties: 1- It has to produce human-understandable explanations which are loyal to the decision making process of the DNN, 2- It has to be locally consistent and efficient, 3- It should be user-friendly, easy to apply and quick in providing explanations. In this work, we propose a new algorithm, explanations via adversarial attacks, which satisfies these 3 important properties and more. We call our method \textbf{A}dversarial \textbf{E}xplanations for \textbf{A}rtificial \textbf{I}ntelligence systems or AXAI \footnote{Code will be readily available.}. AXAI inherits from the nature of adversarial attacks to automatically find and select important features affecting the model's prediction to produce explanations. The idea behind our work comes from the natural behavior of adversarial attacks. The attacks tend to manipulate important features in the input to deceive a DNN. The logic is simple, rather than trying to build a model that learns to explain the DNN's behavior, why don’t we utilize the nature of attacks to learn this behavior? One who knows how to fool a model, certainly knows what the model may be thinking. Another benefit of our approach is that certain attacks, such as the Projected Gradient Descent (PGD) method \cite{madry2017towards}, are fast, efficient, and consistent in their adversarial behavior. Our work further aims to solve at least 2 problems:  1- Provide fast explanations without a need for model training, 2- Reduce the need for selecting a large set of hyper-parameters to produce consistent results.
	
	Obviously, one needs to first show how adversarial attacks link to explainbility, i.e., how an attack can point to the important features in the input and how one can filter out the unimportant ones to produce explanations. Further, one needs to show how an adversary behaves similarly in its approach across models, tasks and datasets so that the explanations are consistent, stable, and applicable to a large group of models. Here, we present a novel algorithm for explaining the DNN's predictions in multiple domains including text, audio and image. In particular, this paper makes the following contributions:
	\begin{itemize}
		\item We show that given an $\ell_2$ PGD attack and a trained DNN, the distribution of attack magnitudes vs. frequency across all unseen test inputs follows a beta distribution, regardless of the task and dataset. We also show that these distributions are symmetric and the differences between their means, medians, and quantiles are not statistically significant.
		
		\item We show that the most important input features, i.e., features with the largest effect on the model's predictions, can be found using a consistent rule across different DNN architectures, datasets, and tasks. This rule leverages the properties of the distributions explained above.
		
		\item We propose a novel algorithm for explaining the outcomes of DNNs and provide a detailed  analysis of our algorithm's performance for different DNN architectures, datasets and tasks.
		\item We benchmark our algorithm against methods such as LIME and SHAP  \cite{ lundberg2017unified, ribeiro2016should} and show that our algorithm performs faster while producing similar or better explainability results. 
	\end{itemize}
	
	\section{Related Work} \label{sec:related}
	One of the popular explainability solutions called LIME \cite{ribeiro2016should} assumes that DNNs are linear locally. LIME trains weighted linear models on the top of the DNN for perturbed samples around a target input to produce explanations. The computational bottleneck in LIME is caused by the training part where a selected number of perturbed samples are sent through the DNN for learning the explanation. Certain combination of LIME's hyper-parameters can produce unstable results \cite{alvarez2018robustness}. DeepLIFT produces explanations by modeling the slope of gradient changes of output with respect to the input \cite{shrikumar2017learning}. Grad-CAM is a saliency-based method that uses the gradients of the input at the final convolutional layer to produce coarse localization maps pointing to important regions in the input \cite{selvaraju2017grad}. The majority of approaches  based on sensitivity maps fail to produce explanations that only rely on important features. Creators of DeepLIFT associate this lack of stability to the behavior of activation functions such as ReLU. \cite{smilkov2017smoothgrad} proposed Smooth Grad which uses gradients and Gaussian based de-noising methods to produce stable explanations. The authors of the paper  mention that large outlier values in the gradient maps produced by gradient differentiation may cause instability. In our algorithm, we overcome the problem of instability by utilizing the density of attacks, which are created iteratively on segments. Some other important works in this area are given in \cite{sundararajan2017axiomatic,jacovi2018understanding, zhao2018respond, bach2015pixel, becker2018interpreting, erhan2009visualizing, letham2015interpretable}.
	
	DNNs are vulnerable to subtle adversarial perturbations applied to their input. The basic idea behind most adversarial attacks revolves around solving a maximization problem with a constraint that keeps the distance between the original input and adversarial input small, so that the adversarial input, while capable of fooling the DNN,  is not perceptually recognizable by humans.  The connection between model interpretation and attacks has recently gravitated the interest of researchers. \cite{ilyas2019adversarial} and \cite{tsipras2018robustness} showed that one benefit of adversarial examples is that they reveal useful insights into the salient features of input data and their effects on DNNs' predictions. Our solution relies on the nature of adversarial attacks to select and produce important and explainable features given a specific input and DNN. Our work puts more emphasis on model interpretability, where we make use of the information obtained from an adversarial attack on a DNN to de-noise the sensitivity maps and produce stable explanations. We de-noise the gradient map by utilizing the iterative nature of the PGD attack and by considering only a minimum number of highly influential gradients that contribute the most to the predictions. We use the density of gradients in a number of segments to remove the noise that was not filtered out in the previous steps and produce human-interpretable explanations. 
	\section{Main Results} \label{sec:main}
\begin{figure}[!htp]
		\centering
		\begin{subfigure}[!]{0.13\textwidth}
			\includegraphics[width=\linewidth]{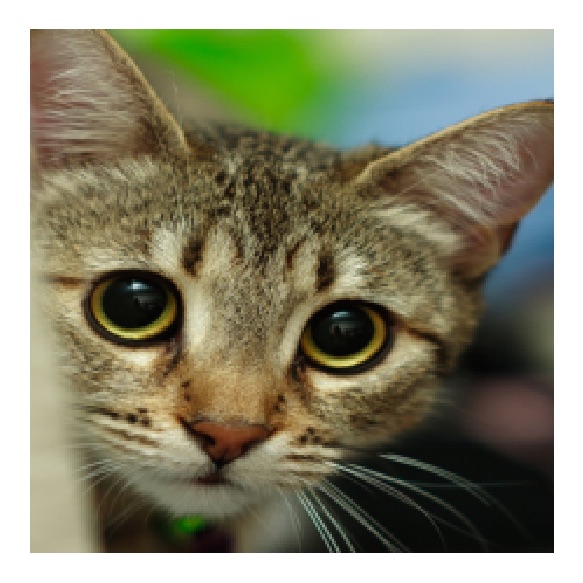} 
			\caption{Original image} \label{fig:1a}
		\end{subfigure}
		\quad
		\begin{subfigure}[!]{0.14\textwidth}
			\includegraphics[width=\linewidth]{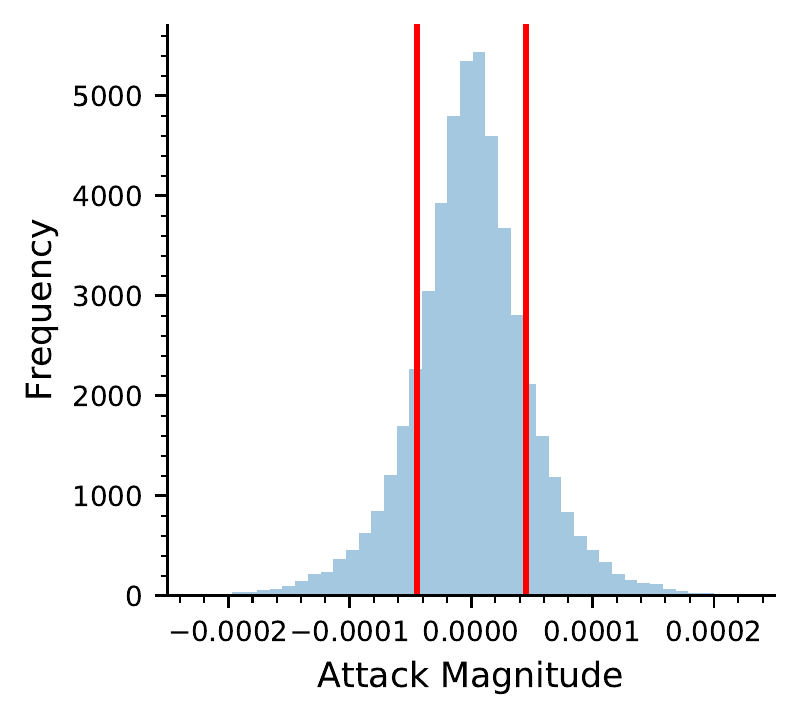} 
			\caption{Attack Magnitude vs. Freq.} \label{fig:1b}
		\end{subfigure}
		\quad
		\begin{subfigure}[!]{0.15\textwidth}
			\includegraphics[width=\linewidth]{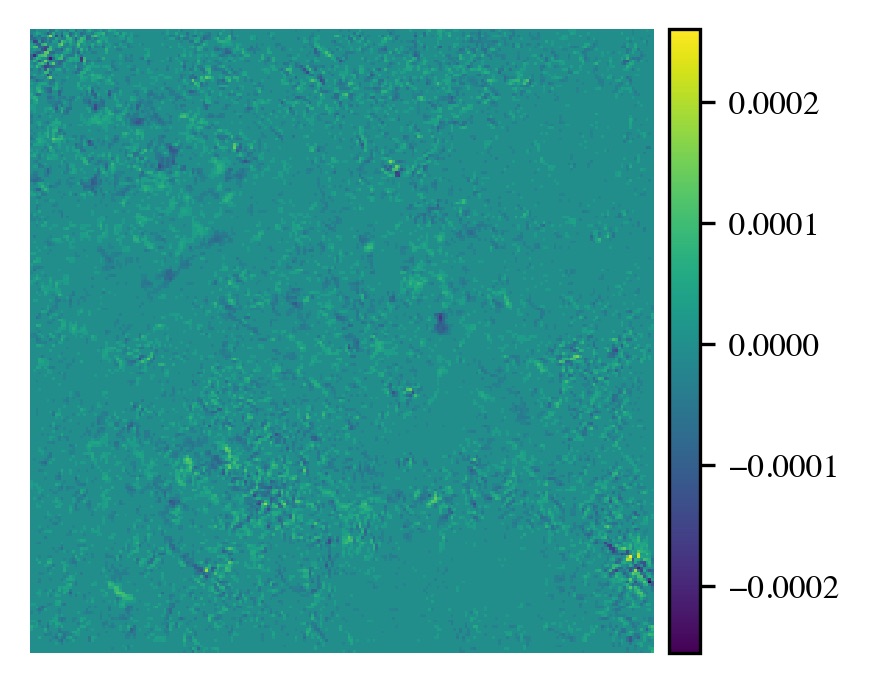} 
			\caption{Adversarial changes in pixels} \label{fig:1c}
		\end{subfigure}
		\quad
		\begin{subfigure}[!]{0.15\textwidth}
			\includegraphics[width=\linewidth]{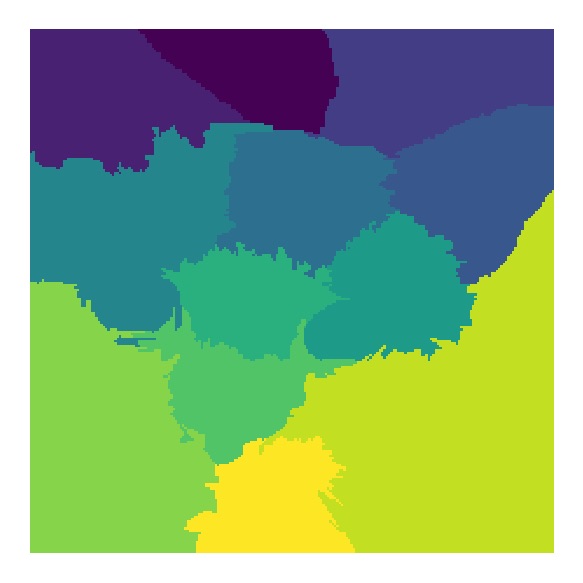} 
			\caption{Image segments} \label{fig:1d}
		\end{subfigure}
		\quad
		\begin{subfigure}[!]{0.14\textwidth}
			\includegraphics[width=\linewidth]{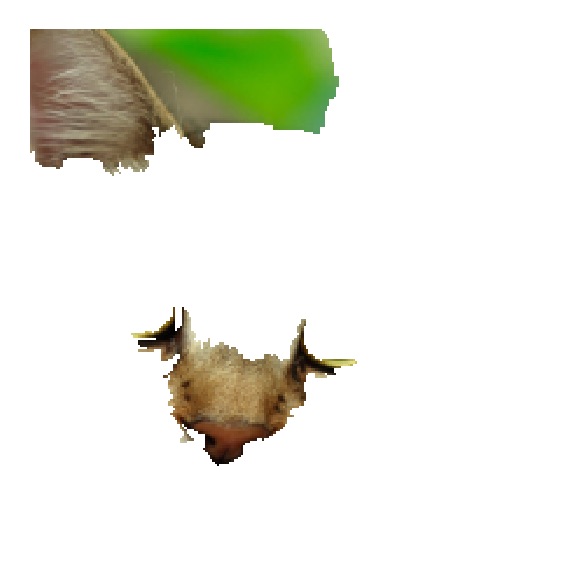} 
			\caption{Explanation} \label{fig:1e}
		\end{subfigure}
		\caption{A simple example depicting the steps taken in AXAI to produce explanations.}  \label{fig:1}
	\end{figure}

	The core idea behind our approach, AXAI, is to utilize the knowledge gained from an adversarial attack on a DNN and an input, to find the important features in the input in order to produce good explanations. This is done by mapping ``carefully filtered attacked inputs'' onto predefined segments and filtering out the unimportant features. This will be discussed in more detail in later sections. First let's look at an example in Fig. \ref{fig:1} to see how our approach works. Given an image classification DNN, the $\ell_2$ adversarial attack changes the pixels in the entire image, as seen in Fig. \ref{fig:1c}. The reason for this is simple: each pixel value is changed by the adversary so that the accumulated loss value can increase enough to fool the DNN. Fig. \ref{fig:1b} shows the distribution of the attack on this image. The x-axis  represents the magnitude of the pixel changes and the y-axis represents the number of pixels given each value on x-axis.  AXAI  maps the strongly attacked pixels to the image segments of the original image and filters out the segments with highest density of attacked pixels which meet certain criteria to produce explanations. Fig. \ref{fig:1c} shows the value changes for the important attacked pixels. As we will show, the important features used for explanations are located at specific sections in the tails of the distribution given in Fig. \ref{fig:1b}. These are the pixels that directly affect the classification decision made by the model. We use QuickShift \cite{vedaldi2008quick} for segmenting the input image (Fig. \ref{fig:1d}). It is important to note that the segmentation step in our algorithm is general and any type of input segmentation method may be utilized for this step depending on the model and input type, e.g., language, signal or imagery. Fig. \ref{fig:1e} shows the explanation produced by our algorithm. 
	
	Algorithm 1 details the steps taken by AXAI to produce an explanation $E$ for the output of a selected model $f$. Suppose that input $X$ is segmented into $p$ groups using a segmentation method and that the attack magnitudes for the input $X$  and DNN  $f$ are obtained.  Let $X_{diff}$ be the difference between the original $X$ and adversarial $X^\prime$. We filter out the low intensity attack magnitudes $X_{diff}$ and create a Boolean array $X_{difft}$, where values larger than a threshold, are only set to True. Let $Su$ be the set of unique segments, $Su=\{Su_{1},…,Su_{p}\}$. Next, we map the filtered attack $X_{difft}$ to the segments $Su$, and create a new list of filtered attack groups, $Su_{x}=\{Su_{x_1},…, Su_{x_p}\}$. The mapping function, \emph{Map} in Algorithm 1, simply stacks the filtered attacks on the segments and groups the filtered attack $X_{difft}$ based on the segments.  Finally, the attack density of each unique segment can be written as $Su_{d}=\{\frac{card(Su_{x_1})}{card(Su_{1})},…,\frac{card(Su_{x_p})}{card(Su_{p})}\}$ (\emph{Calculate\_density} in Algorithm 1). We then extract the indices $j$'s of the top $K$ maximum values in $Su_{d}$ (\emph{TopK\_indices} in Algorithm 1), and produce $Su(j)$ as explanation $E$ for the input $X$.  In next sections, we explain each step in details.
	\begin{algorithm}
		\caption{AXAI} 
		\begin{algorithmic}[1]
			\Require {Model $f$, input $X$}
			\State $X' \leftarrow Attack(f,X)$\Comment{i.e. PGD attack}
			\State $X_{diff} = x' - x$\Comment{The attack magnitudes}
			\State $X_{difft}  \leftarrow Threshold(X_{diff})$\Comment{Filtered attack magnitudes}
			\State $Su \leftarrow Segment(X)) $ 
			\State $Su_{x} \leftarrow Map(X_{difft}, Su)$\Comment{Group attack magnitudes based on segmentation}
			\State $Su_{d} \leftarrow Calculate\_density(Su_{x})$\Comment{Calculate attacks per segment}
			\State \textbf{return} $Su(TopK\_indices(Su_{d}))$
		\end{algorithmic} 
	\end{algorithm}
	\subsection{White-box adversarial attacks}  \label{subsec:adv}
	Adversary can  attack a DNN by adding engineered noise to the input to increase the associated loss value, if it has some prior knowledge of the DNN including the weights and biases. AXAI utilizes Projected Gradient Descend (PGD) attack \cite{madry2017towards}, although any $\ell_2$ adversarial attack can replace PGD in our algorithm (Appendix \ref{app_2}). However, PGD provides specific benefits such as stability and gradient smoothness that other attacks do not. PGD can be thought of as an iterative version of $\ell_2$ Fast Gradient Method (FGM) attack  \cite{goodfellow2014explaining}, where in each iteration, the adversarial changes are clipped into an $\ell_2$ ball of some $\epsilon$ value. PGD is generally considered a strong stable attack and is defined as,
	\begin{equation}
	x^{t+1}=\sqcap_{x+S}(x^{t}+\epsilon\nabla_x L(\Theta,x,y)),
	\end{equation}
	where for $t$ iterations, $x$ and $y$ are the inputs and outputs, and $\Theta$ are the weights and biases.
	\subsection{Statistical analysis of  attack magnitudes vs. frequency distributions} \label{subsec:stat}
	Here, we briefly report our statistical analysis of  attack magnitudes vs. frequency distributions for a fixed DNN, dataset and an adversarial attack. We can  show that the distributions are similar in their ``shapes,'' ``means,'' ``mean ranks,'' ``medians,'' and ``quantiles,'' and follow a Beta distribution with specific parameters. Given that there is no significant difference in the distributions, we can provide a universal threshold using quantiles which separates the important features from the rest to produce explanations.
	
	To be able to show that highly perturbed regions can be chosen to produce explanations for a single input, we should first show analytically that the results are consistent for all inputs, i.e., adversarial attacks are consistent in their adversarial behavior and the manner in which they attack the most influential input segments . Our analyses prove this point and consequently we can show that our proposed rule to find the important input segments for a single input holds. Finally, we  empirically show that these segments are indeed the most important parts of the input by analyzing the effects of them on the test error rate.
	
\begin{figure}[!]
		\centering
		\begin{subfigure}[!]{0.2\textwidth}
			\includegraphics[width=\linewidth]{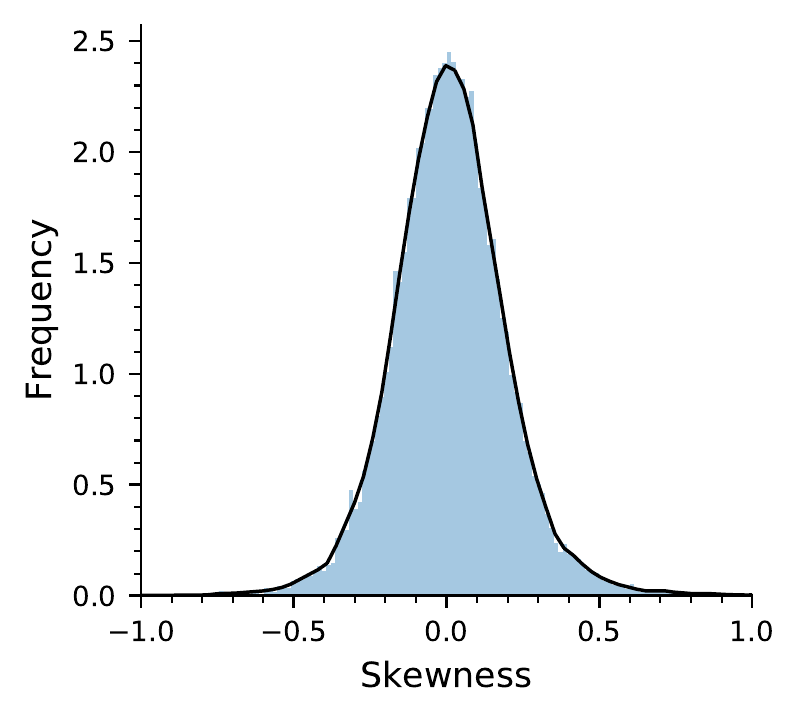} 
			\caption{PGD, AlexNet, CIFAR10} \label{fig:10a}
		\end{subfigure}
		\quad
		\begin{subfigure}[!]{0.2\textwidth}
			\includegraphics[width=\linewidth]{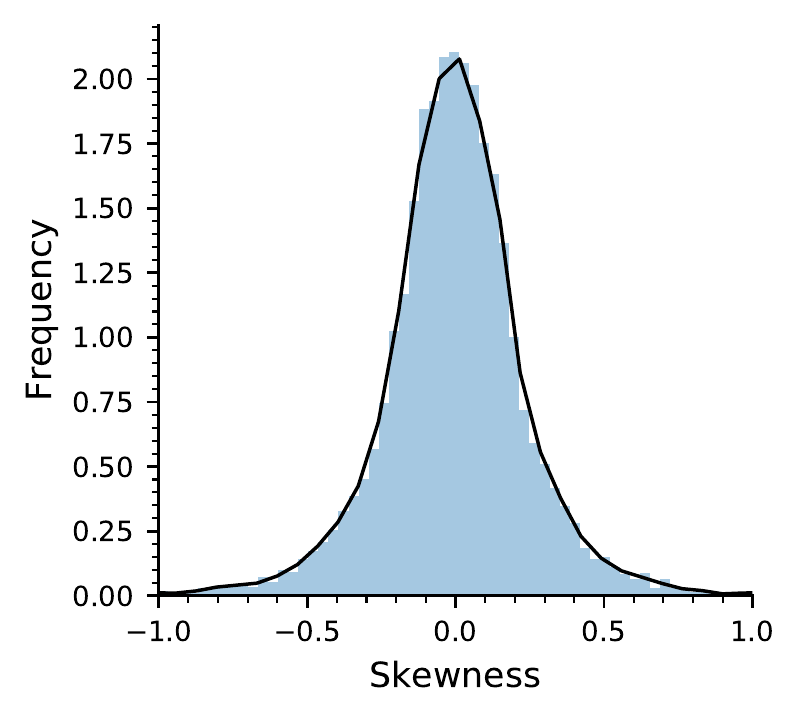} 
			\caption{PGD, VGG16, CIFAR100} \label{fig:10c}
		\end{subfigure}
		\quad
		\begin{subfigure}[!]{0.2\textwidth}
			\includegraphics[width=\linewidth]{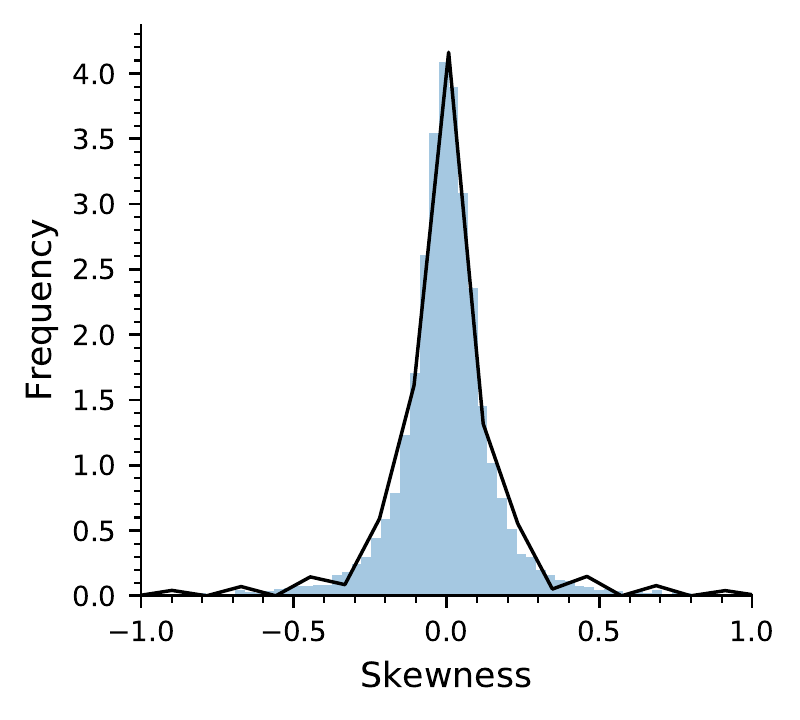} 
			\caption{PGD, ResNet34, ImageNet} \label{fig:10b}
		\end{subfigure}
		\caption{The Fisher-Pearson coefficient of attack magnitudes vs. frequency  distributions.}  \label{fig:10}
	\end{figure}

	We can measure the symmetricity of distributions using the Fisher-Pearson coefficient of skewness. We present the results for AlexNet on  CIFAR10 \cite{kaur2018convolutional}, VGG16 on CIFAR100 \cite{krizhevsky2009cifar} and ResNet34 on ImageNet \cite{deng2009imagenet}.  The Fisher-Pearson coefficients of the attack magnitudes vs. frequency  distributions for all cases are shown in Fig. \ref{fig:10}. It is seen that the skewness of all distributions falls within the $[-0.5, 0.5]$ range showing strong evidence that they are approximately symmetric \cite{bulmer1979principles}. Only 0.9\% of CIFAR10, 3.3\% of CIFAR100 and 1.9\% of ImageNet test datasets lie outside of  $[-0.5, 0.5]$ range.
	
\begin{figure}[h]
		\centering
		\begin{subfigure}[!]{0.22\textwidth}
			\includegraphics[width=\linewidth]{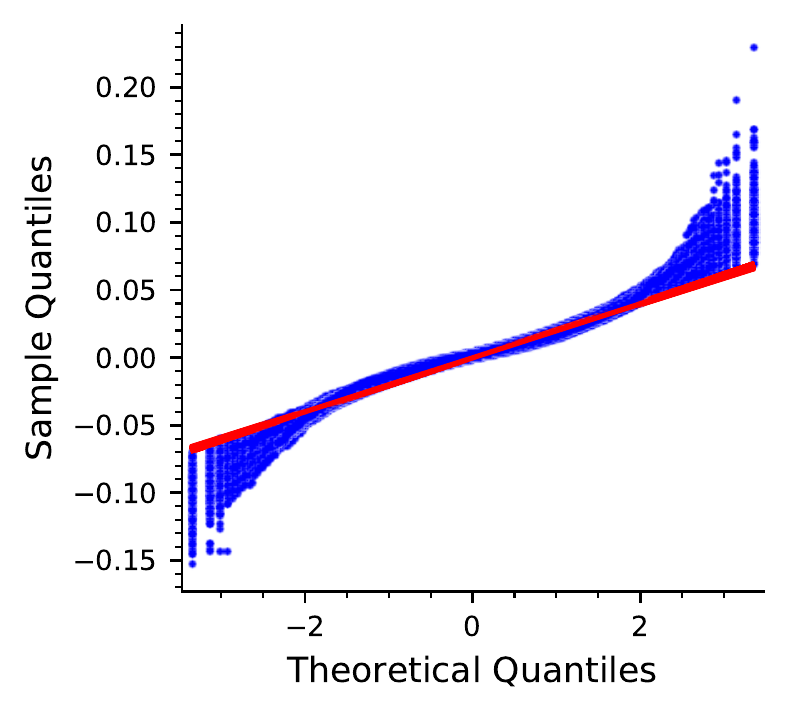} 
			\caption{PGD, AlexNet, CIFAR10} \label{fig:11a}
		\end{subfigure}
		\quad
		\begin{subfigure}[!]{0.22\textwidth}
			\includegraphics[width=\linewidth]{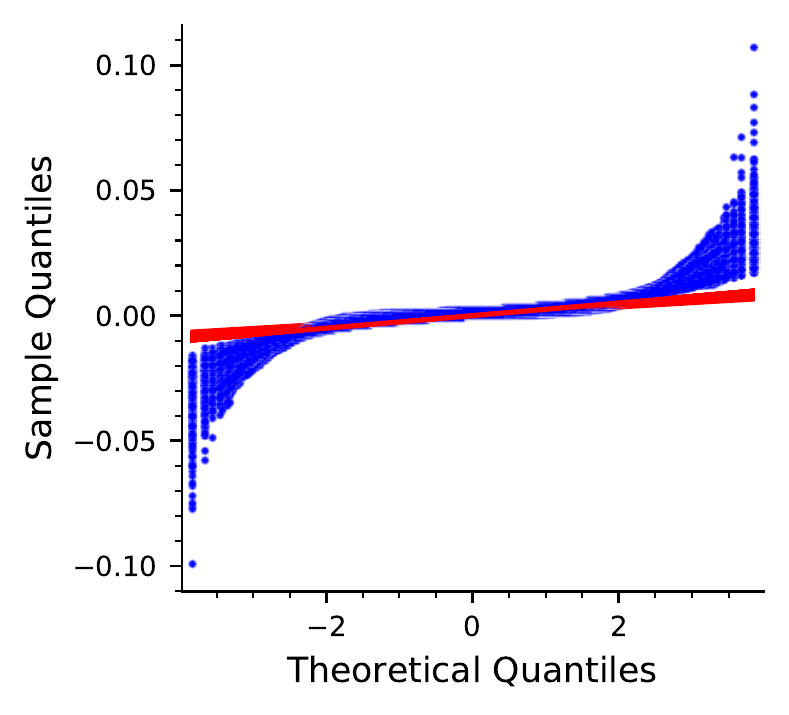} 
			\caption{PGD, ResNet34, ImageNet} \label{fig:12b}
		\end{subfigure}
		\caption{The Q-Q plot of  sample distributions vs. theoretical normal distribution (mean=0, std=1).}  \label{fig:11}
	\end{figure}

	Quantile-Quantile (Q-Q) plot allows us to understand how the quantiles of a distribution deviate from a specified theoretical distribution. The theoretical distribution selected is the normal distribution. The x-axis and y-axis represent the quantile values of the theoretical and sample distributions, respectively. While it is unlikely to have identical distributions that perfectly match, one can look at different parts of the Q-Q plot to distinguish between the similar and dissimilar locations in the distributions. Fig. \ref{fig:11} shows the Q-Q plots for random subsets of ImageNet and CIFAR10 test datasets each containing 1000 images. It is seen that the distributions follow a fairly straight line in the middle portion of the curve, while deviating at the upper and lower parts. This provides some evidence supporting the hypothesis that distributions are symmetrical with heavier tails. 
	\begin{table}[!]
		\centering
		\resizebox{1.\columnwidth}{!}{
			\begin{tabular}{|l|l|l|l|l|}
				\hline
				& t-test (CIFAR10) & Mann-Whitney (CIFAR10) & t-test (ImageNet) & Mann-Whitney (ImageNet) \\ \hline
				p-value & 0.70             & 0.58                      & 0.64              & 0.55                         \\ \hline
		\end{tabular}}
		\caption{p-values for the mean similarity statistical tests at significance level 0.05.}
		\label{tab:table3}
	\end{table}
	
	We perform the two-sample location t-test and Mann-Whitney U test to determine if there is a significant difference between the hypotheses where the null hypothesis is the equality of the means. Carrying out pair t-tests on all samples allows us to be conservative in confirming the mean similarity of the distributions. A sample here is defined as the attack magnitudes vs. frequency distribution for a data point in the test adversarial dataset created by the PGD attack on a DNN trained on the training dataset. The results reported in Table \ref{tab:table3} indicate no significant difference between the means. Further, the Mann-Whitney U test results indicate that all pairs are similar to each other on the mean ranks. Under the assumption of two distributions having similar shapes, one could further state that Mann-Whitney test can be considered as a test of medians \cite{mcdonald2009handbook}. Since, we have shown that the shapes are similar, we can conclude that there are no significant difference between the medians of the distributions. Further details in addition to the results for the ANOVA test are given in Appendix \ref{app_3}.
	\begin{table}[]
		\centering
		\resizebox{1.\columnwidth}{!}{
			\begin{tabular}{|l|l|l|l|}
				\hline
				& AlexNet, CIFAR10, PGD & VGG16, CIFAR100, PGD & ResNet34, ImageNet, PGD\\ \hline
				15th Quantile &     $ (-1.807e-02, -1.805e-02)$&$	(-1.419e-02, -1.414e-02)$&$	(-1.785e-03, -1.777e-03)   $      \\ \hline
				25th Quantile &     $ (-1.145e-02, -1.071e-02)$&$	(-8.153e-03, -8.110e-03)$&$	(-1.015e-03, -1.101e-03)   $      \\ \hline
				Mean   & $ (1.775e-05, 2.295e-05)$	& $(-6.850e-06, -3.624e-06)$ &	$(-1.090e-07, -6.000e-08)    $        \\ \hline
				Median   & $ (2.115e-06, 1.127e-05)	$&$(-2.842e-06, 4.467e-06)$&$	(-2.155e-07, -9.381e-08)$		\\ \hline
				75th Quantile &     $ (1.071e-02, 1.073e-02)$&$	(8.102e-03, 8.146e-03)$&$	(1.011e-03, 1.016e-03)   $      \\ \hline
				85th Quantile &         $   (1.809e-02, 1.812e-02)$&$	(1.413e-02, 1.418e-02)$&$	(1.777e-03, 1.785e-03)            $               \\ \hline
			\end{tabular}
		}
		\caption{Estimations for mean, median, 15th , 25th, 75th and 85th quantiles at 95\% confidence level.}
		\label{tab:table4}
	\end{table}
	\begin{table}[]
		\centering
		\resizebox{1.\columnwidth}{!}{
			\begin{tabular}{|l|l|l|l|}
				\hline
				& AlexNet, CIFAR10, PGD & VGG16, CIFAR100, PGD & ResNet34, ImageNet, PGD \\ \hline
				$p$   & $ (1.124e+01, 1.132e+01)$	& $(2.129e+01, 2.171e+01)$ &	$(1.306e+02, 1.329e+02)    $        \\ \hline
				$q$   & $ (1.136e+01, 1.145e+01)	$&$(2.124e+01, 2.164e+01)$&$	(1.303e+02, 1.326e+02)$		\\ \hline
			\end{tabular}
		}
		\caption{Statistical estimations for parameters of  beta distribution at 95\% confidence level.}
		\label{tab:table5}
	\end{table}
	
	Next, to show consistency across distributions for a given model, dataset and attack, we estimate the values of quantiles, means and medians. We do this by estimating the statistics of the distributions and constructing confidences intervals. For each experiment, we estimate the mean, median, 15th, 25th, 75th and 85th quantiles of each attack magnitude vs. frequency distribution for the entire test dataset. The statistical confidence interval estimations at confidence level of $95\%$ are reported in Table \ref{tab:table4}. Our results show that the confidence intervals have narrow ranges and the estimations are consistent. The estimates for the 15th, 25th, 75th and 85th quantiles indicate a strong  symmetricity with respect to the origin in all cases. This matches the results of the skewness test in Fig. \ref{fig:10}. Another observation is that the confidence interval of the mean and medians are pretty narrow, supporting the results of the t-tests and Mann-Whitney U test. Finally, we can  show with high confidence that the distributions consistently follow a beta distribution. The beta distribution is a family of  distributions defined by two positive shape parameters, denoted by $p$ and $q$.  The estimated $p$ and $q$ of the beta distribution are reported in Table \ref{tab:table5}. Further technical details on our analyses presented in this section, in addition to further experiments with audio and text input types,  are provided in Appendix \ref{app_3}.
	
	\subsection{Quantile selection for the explanations} \label{subsec:quan}
\begin{figure}[!]
		\centering
		\begin{subfigure}[!]{0.14\textwidth}
			\includegraphics[width=\linewidth]{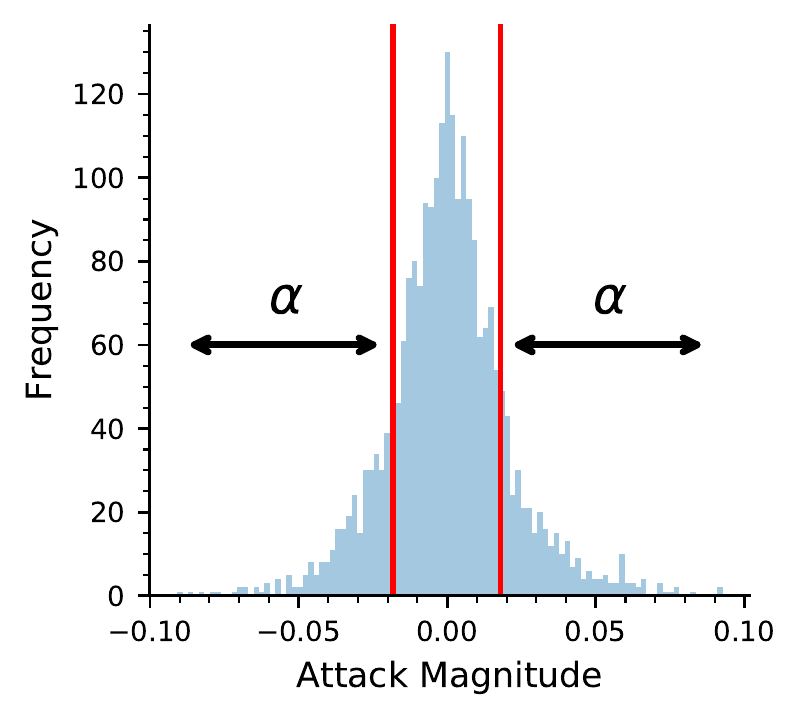} 
			\caption{} \label{fig:12a}
		\end{subfigure}
		\quad
		\begin{subfigure}[!]{0.14\textwidth}
			\includegraphics[width=\linewidth]{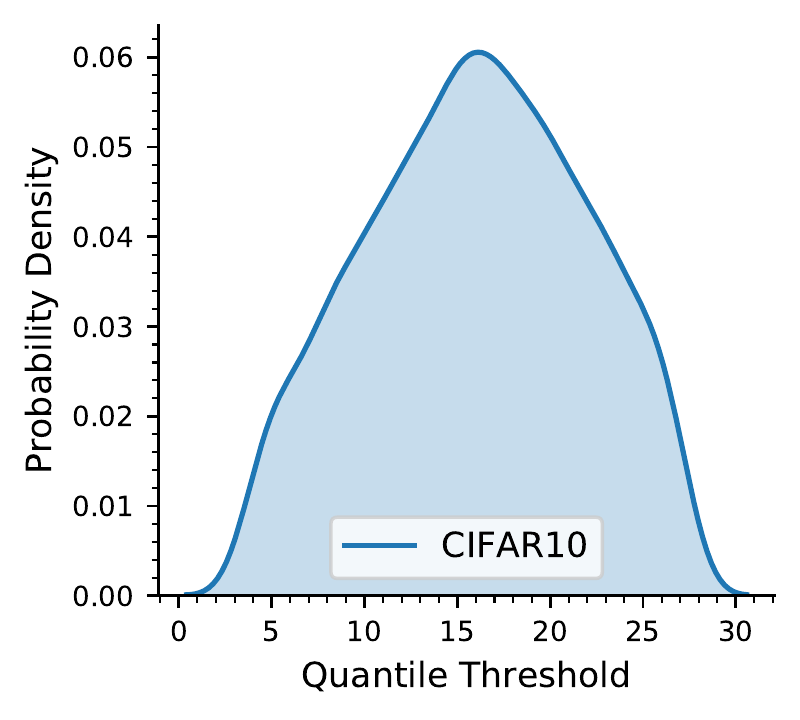} 
			\caption{} \label{fig:12b}
		\end{subfigure}
		\quad
		\begin{subfigure}[!]{0.14\textwidth}
			\includegraphics[width=\linewidth]{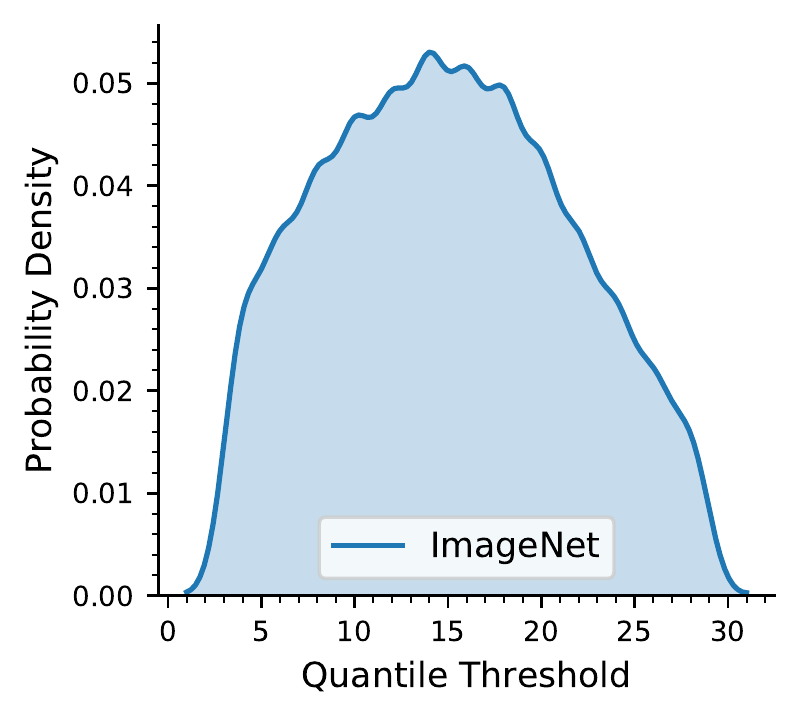} 
			\caption{} \label{fig:12c}
		\end{subfigure}
		\caption{Visualization of the re-attacking process where only portions of inputs lying outside the red lines are attacked ($[0\%, \alpha\%]$,  $[(100-\alpha)\%, 100\%]$ ) (b) AlexNet, CIFAR10 (c) ResNet34, ImageNet}  \label{fig:12}
	\end{figure}

	Our  algorithm produces explanations that rely only on the features in the input that have the largest effect on the predictions. While the majority of  the input is attacked, our belief is that only important features are strongly attacked. We show how one can select the boundary threshold between ``explainable features'' and the rest based on attack magnitudes. We demonstrate this with 2 experiments: 1) AlexNet trained on CIFAR10, 2) ResNet34 trained on ImageNet, both attacked by PGD with 20 iterations. In each case, we select the successfully attacked inputs from the adversarial test dataset, i.e., the inputs that fool the DNN. We then only re-attack specific features of the original clean inputs within the $[0\%$, $\alpha\%]$ and $[(100-\alpha)\%, 100\%]$  percentile of the distributions, where $\alpha$ is the percentage threshold. The re-attacking process starts from $\alpha=0$, where none of the input features are attacked, and then we gradually increase the value of $\alpha$ until the attack successfully changes the prediction, and then we save the value of $\alpha$ (Fig. \ref{fig:12a}). We repeat this for every input. The probability density distribution of $\alpha$'s are given in Fig. \ref{fig:12b} and Fig. \ref{fig:12c} with an estimated mean of $\alpha=15$.
	\begin{table}[!]
		\centering
		\resizebox{1.\columnwidth}{!}{
			\begin{tabular}{|l|l|l|l|l|l|l|l|}
				\hline
				& \multicolumn{2}{l|}{CIFAR10, AlexNet} & \multicolumn{2}{l|}{ImageNet, ResNet34} &                          & CIFAR10, AlexNet & ImageNet, ResNet34 \\ \hline
				Attack Percentile & \multicolumn{2}{l|}{}                 & \multicolumn{2}{l|}{}                   & Attack Percentile        &                  &                    \\ \hline
				$15\%-85\%$       & \multicolumn{2}{l|}{0.78}             & \multicolumn{2}{l|}{0.88}               & $0\%-15\% \& 85\%-100\%$ & 0.16             & 0.07               \\ \hline
				$10\%-90\%$       & \multicolumn{2}{l|}{0.26}             & \multicolumn{2}{l|}{0.79}               & $0\%-10\% \& 90\%-100\%$ & 0.26             & 0.13               \\ \hline
				$5\%-95\%$        & \multicolumn{2}{l|}{0.50}             & \multicolumn{2}{l|}{0.63}               & $0\%-5\% \& 95\%-100\%$  & 0.45             & 0.25               \\ \hline
				$1\%-99\%$        & \multicolumn{2}{l|}{0.07}             & \multicolumn{2}{l|}{0.12}               & $0\%-1\% \& 99\%-100\%$  & 0.92             & 0.80               \\ \hline
			\end{tabular}
		}
		\caption{Adversarial test accuracy where only features within a certain percentile of the attack magnitudes vs. frequency distributions are attacked (PGD with 20 Iterations).}
		\label{tab:table1}
	\end{table}
	
	Further, we report the test accuracies of the DNNs on  the adversarial test datasets that are created based on different attack percentiles. Given an attack percentile range, the adversarial test dataset consists of adversarial test inputs which are created by attacking only portions of the input features that lie withing a specific percentile range of the attack magnitudes vs. frequency distributions similar to above. This allows us to understand how  the features lying in the middle area, tails and outliers of the distributions affect the DNN's predictions. Our findings are reported in Table \ref{tab:table1}. Our results show that the majority of  the input features including  those within the first two standard deviations and the outliers of the distributions do not have a strong effect on the predictions. A smaller portion of the input features which are also those attacked with the highest intensity, i.e., within the $[0\%,15\%]$  and $[85\%, 100\%]$  percentiles of the distributions have the largest effect on the DNN's predictions, confirming our hypothesis. We see the same trend across different DNNs and datasets (Appendix \ref{app_3}).  
	\section{Experiment Results} \label{exp}
	Earlier, we provided a sample explanation created by AXAI for an image classifier. Appendix \ref{app_5} contains more experiments for image classification and object detection DNNs. Further, Appendix \ref{app_5} contains an ablation study and an interesting comparison between explanations produced by a non-robust model and an adversarially robust model. Here, we provide sample explanations produced by our algorithm for speech recognition and language-based tasks. 
	\subsection{Explaining a speech recognition model}
\begin{figure}[!htp]
		\centering
		\begin{subfigure}[!]{0.22\textwidth}
			\includegraphics[width=\linewidth]{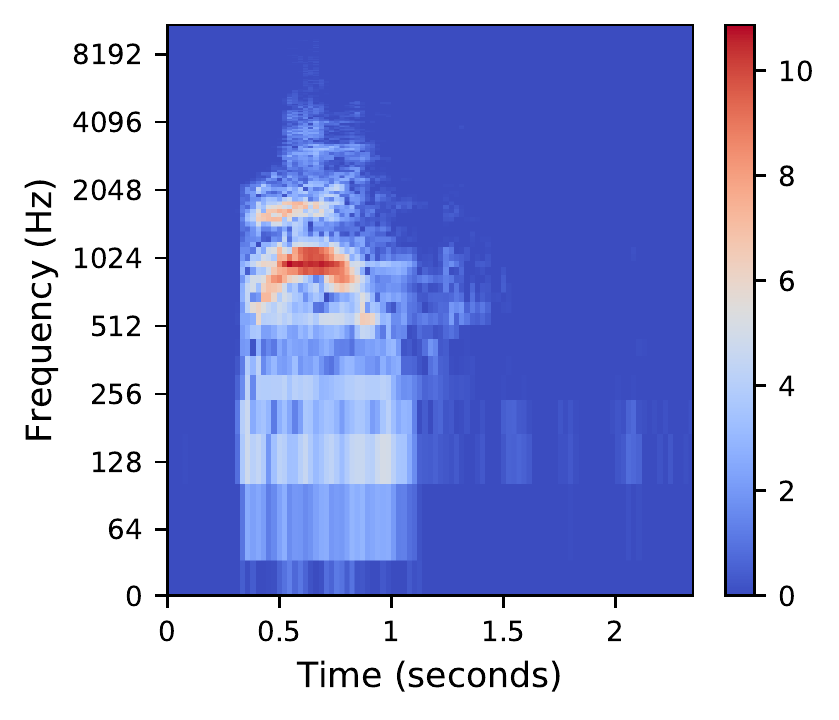} 
			\caption{"Right" } \label{fig:13a}
		\end{subfigure}
		\quad
		\begin{subfigure}[!]{0.22\textwidth}
			\includegraphics[width=\linewidth]{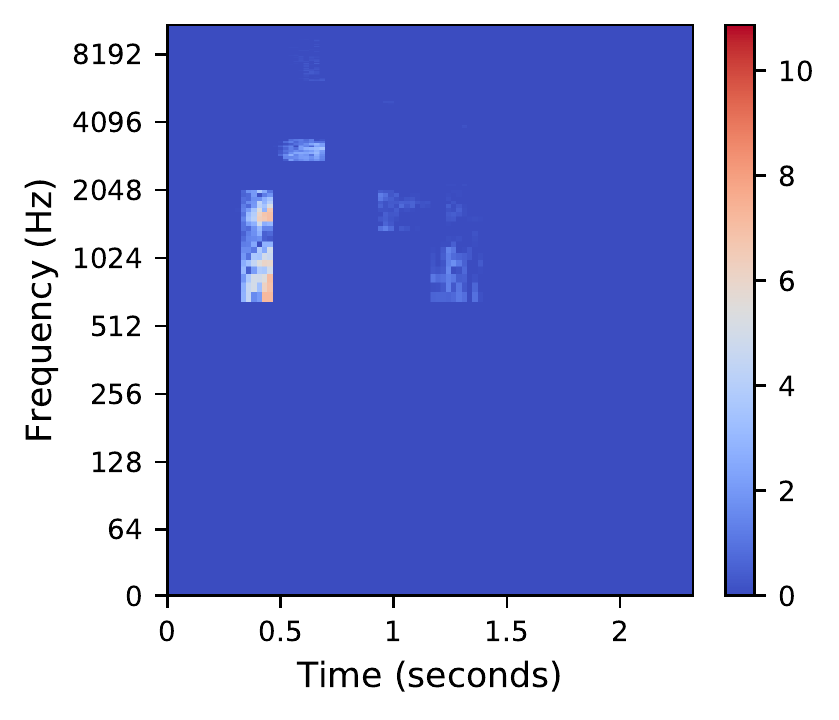} 
			\caption{Explanation} \label{fig:13b}
		\end{subfigure}
		\quad
		\begin{subfigure}[!]{0.22\textwidth}
			\includegraphics[width=\linewidth]{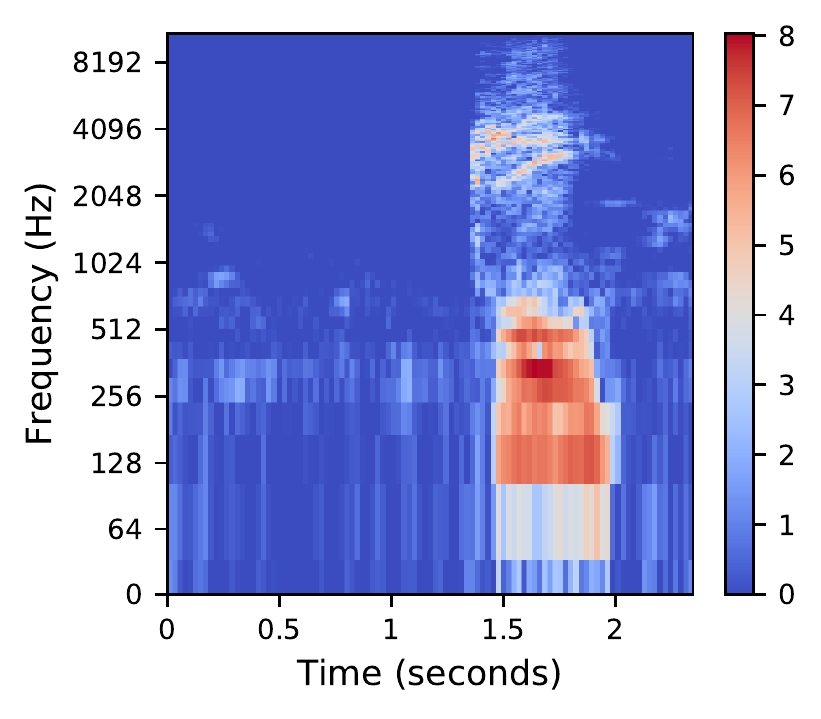} 
			\caption{"Three"} \label{fig:13c}
		\end{subfigure}
		\quad
		\begin{subfigure}[!]{0.22\textwidth}
			\includegraphics[width=\linewidth]{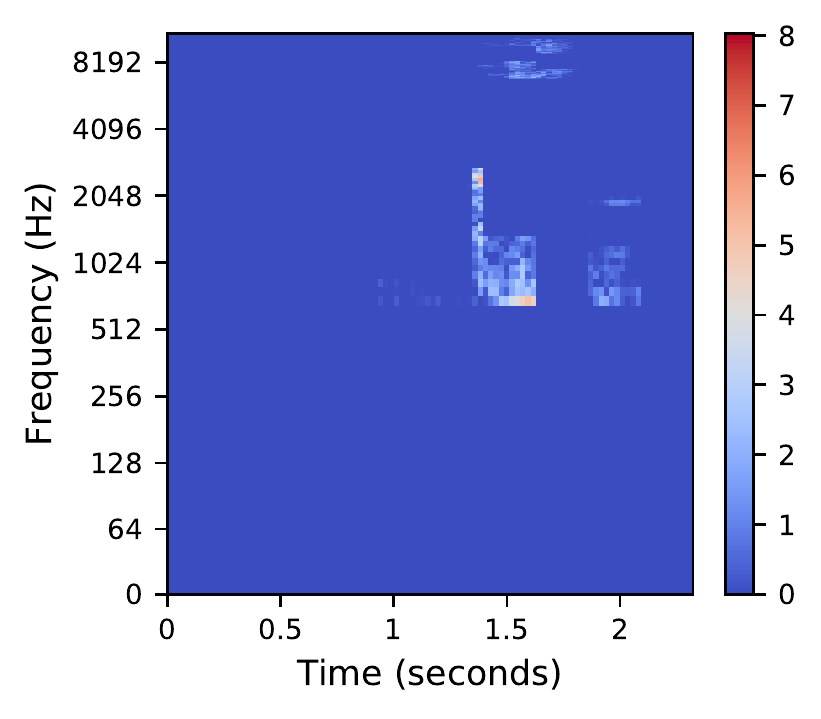} 
			\caption{Explanation} \label{fig:13d}
		\end{subfigure}
		\quad
		\caption{The AXAI explanations for the LeNet speech recognition model.}  \label{fig:13}
	\end{figure}

	The Speech Commands Dataset  \cite{warden2018speech} is an audio dataset of short spoken words. Here, we have converted the audio files to spectrograms and used them to train a LeNet model to identify ``speech commands.'' We have created time-frequency segments by dividing the spectrogram into time-frequency grids similar to \cite{Mishra2017LocalIM}. The x-axis and y-axis indicate the time-scale and log-scale frequency of the spectograms respectively, and the color bar indicates the magnitude. This kind of segmentation results in equal sized rectangular blocks where the height of the segment covers the range of frequencies (y-axis) and the width of the segment covers the range of the time (x-axis) associated with the spoken word. The spectrogram of the first word "Right" and its explanation are shown in Fig. \ref{fig:13a} and Fig. \ref{fig:13b}. The explanation shows that the first and last character in the spoken word ``Right'' stand out as important features ($[0.4s, 0.6s]$ and $[1.0s , 1.2s]$ intervals). This is reasonable because ``Five'' is the neighboring class of  ``Right'' in the dataset (Appendix \ref{app_4}) and ``Right'' and ``Five'' differ in the pronunciation of ``r'' and ``f'' and ``t'' and ``v.''  The second example is  for the word ``Three'' (Fig. \ref{fig:13c} and Fig. \ref{fig:13d}).  The produced explanation indicates the importance of ``Thr'' ($[1.4s, 1.7s]$ interval). This is reasonable because ``Three'' and its neighbor ``Tree'' differ in the letter ``h'' in ``Thr,'' and this difference is learned by the model during training to identify the two words correctly. More examples are shown in \ref{fig:26}. Details on this experiment are given in Appendix \ref{app_5}.
	
	\subsection{Explaining a text classification model}
\begin{figure}[!]
		\centering
		\begin{subfigure}[!]{0.3\textwidth}
			\includegraphics[width=\linewidth]{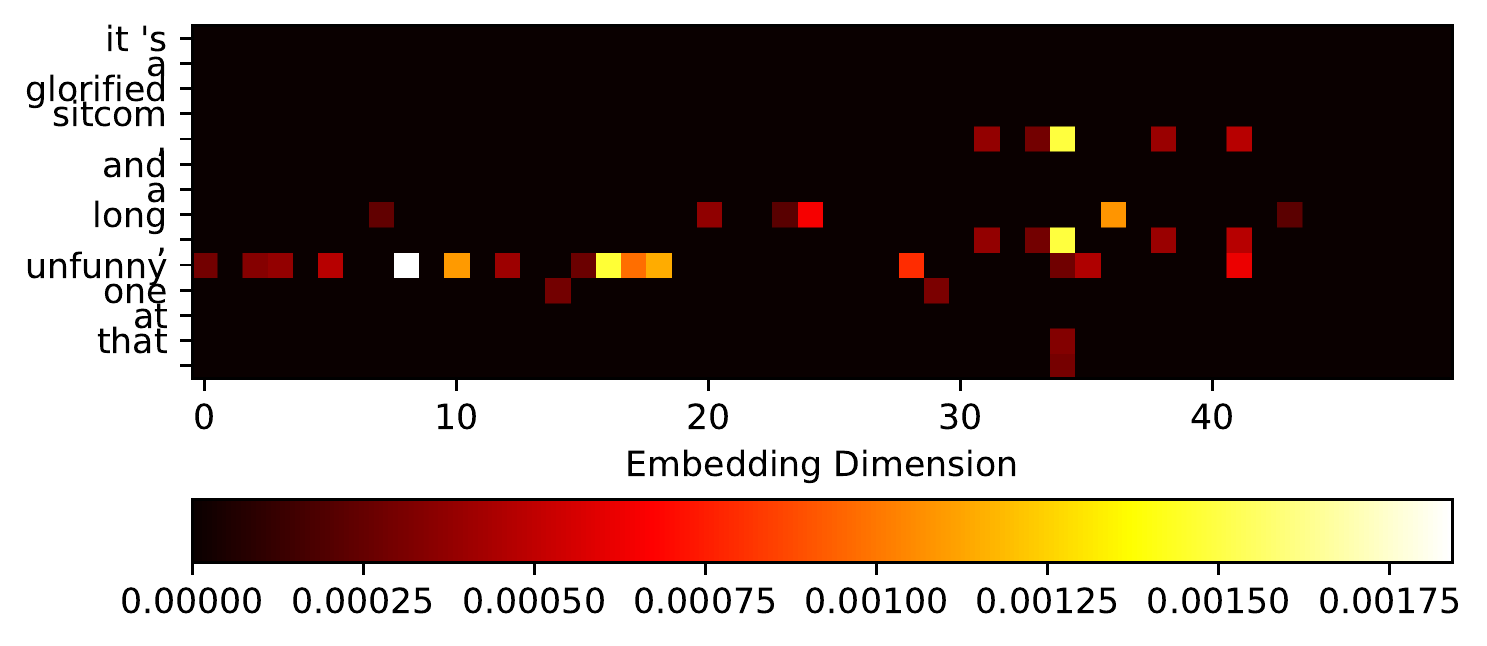} 
			\caption{Text example 1} \label{fig:144a}
		\end{subfigure}
		\quad
		\begin{subfigure}[!]{0.4\textwidth}
			\includegraphics[width=\linewidth]{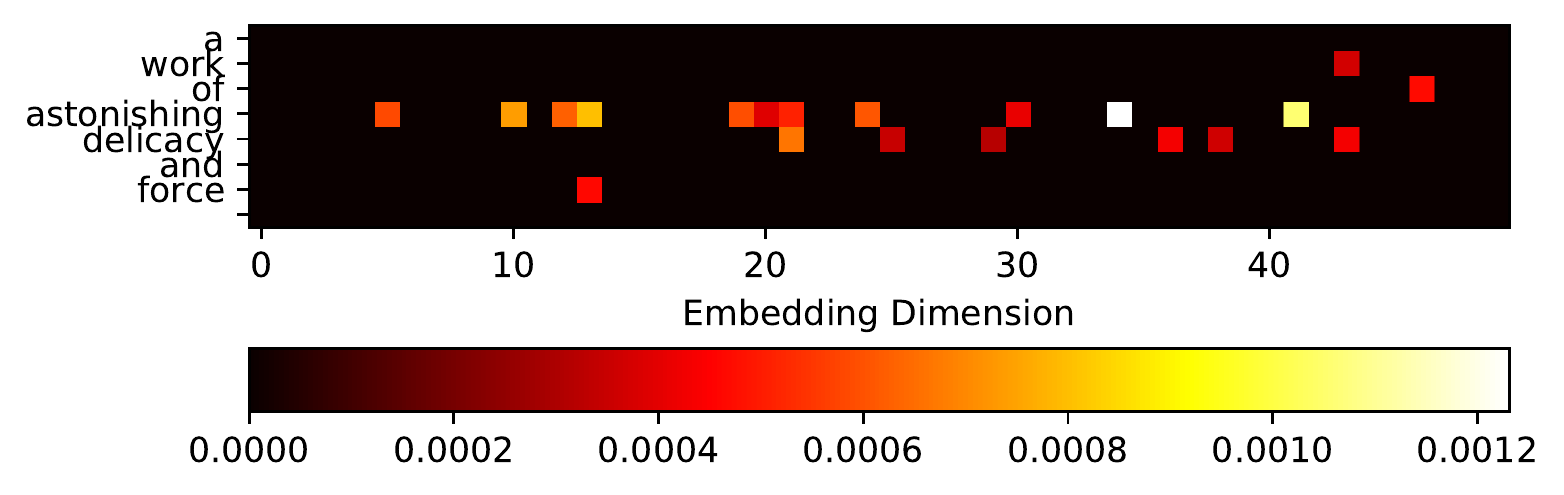} 
			\caption{Text example 2} \label{fig:144b}
		\end{subfigure}
		\quad
		\caption{The AXAI explanations for the sentence classification model.}  \label{fig:144}
	\end{figure}

\begin{figure}[!]
		\centering
		\begin{subfigure}[!]{0.3\textwidth}
			\includegraphics[width=\linewidth]{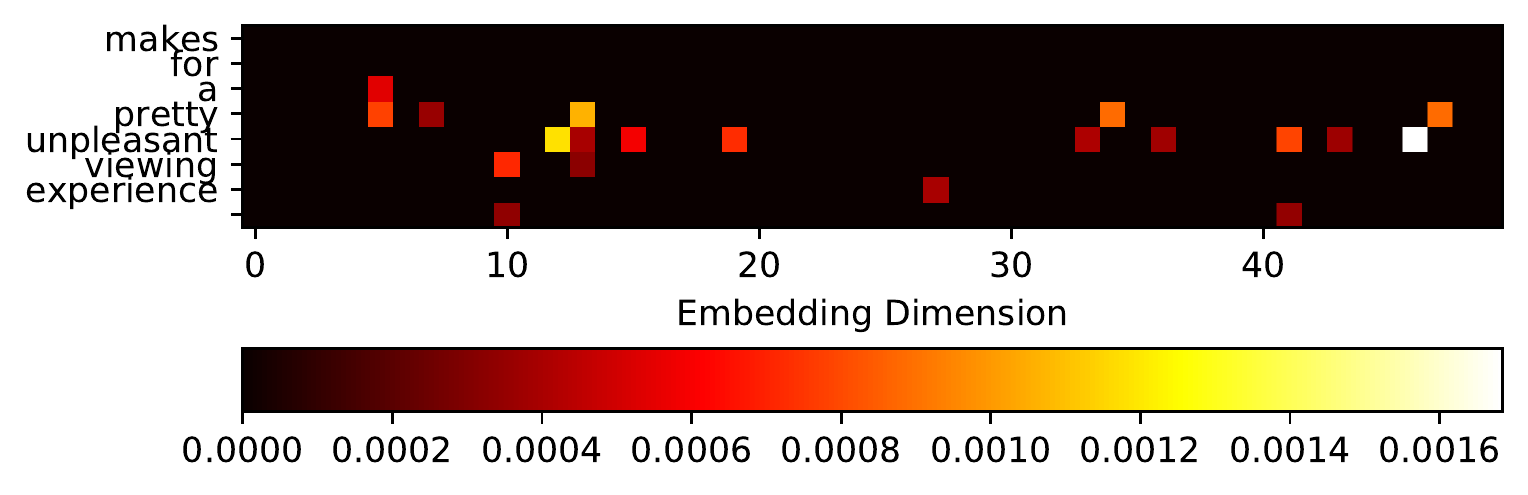} 
			\caption{Text example 1} \label{fig:144c}
		\end{subfigure}
		\quad
		\begin{subfigure}[!]{0.3\textwidth}
			\includegraphics[width=\linewidth]{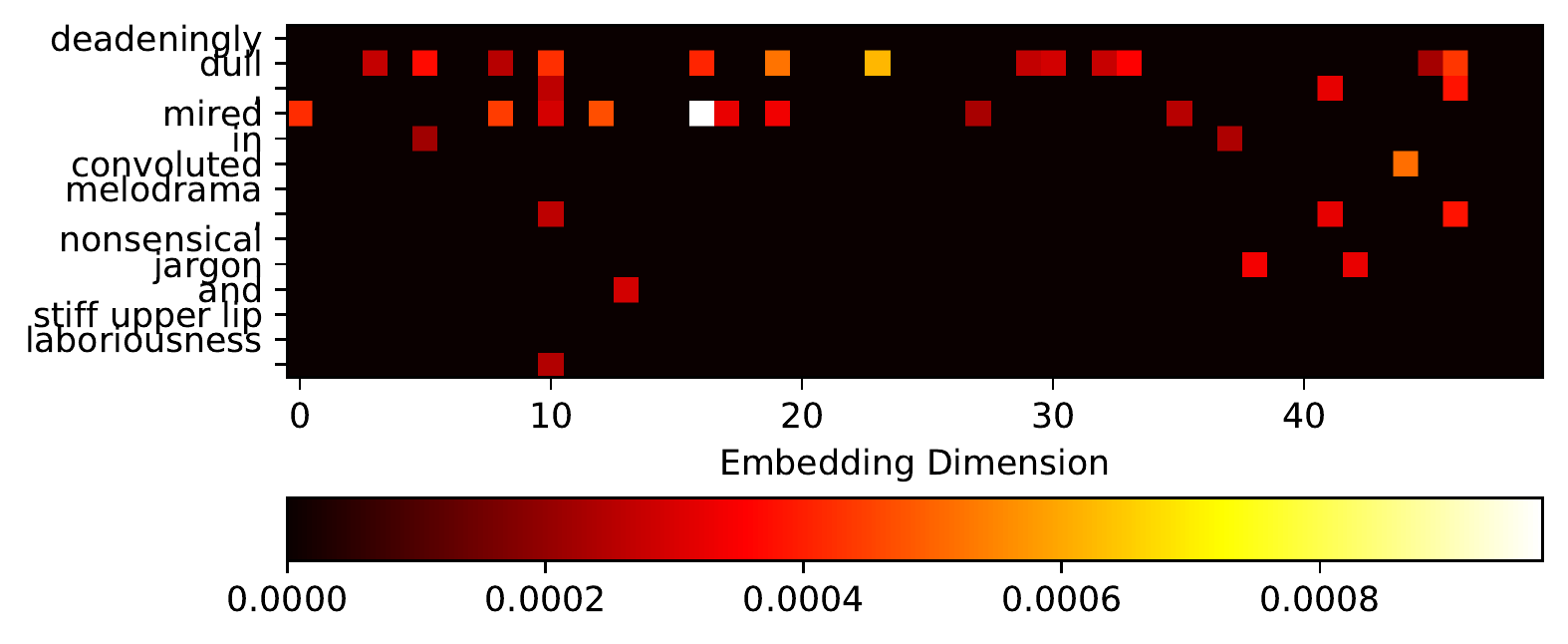} 
			\caption{Text example 2} \label{fig:144d}
		\end{subfigure}
		\quad
		\begin{subfigure}[!]{0.3\textwidth}
			\includegraphics[width=\linewidth]{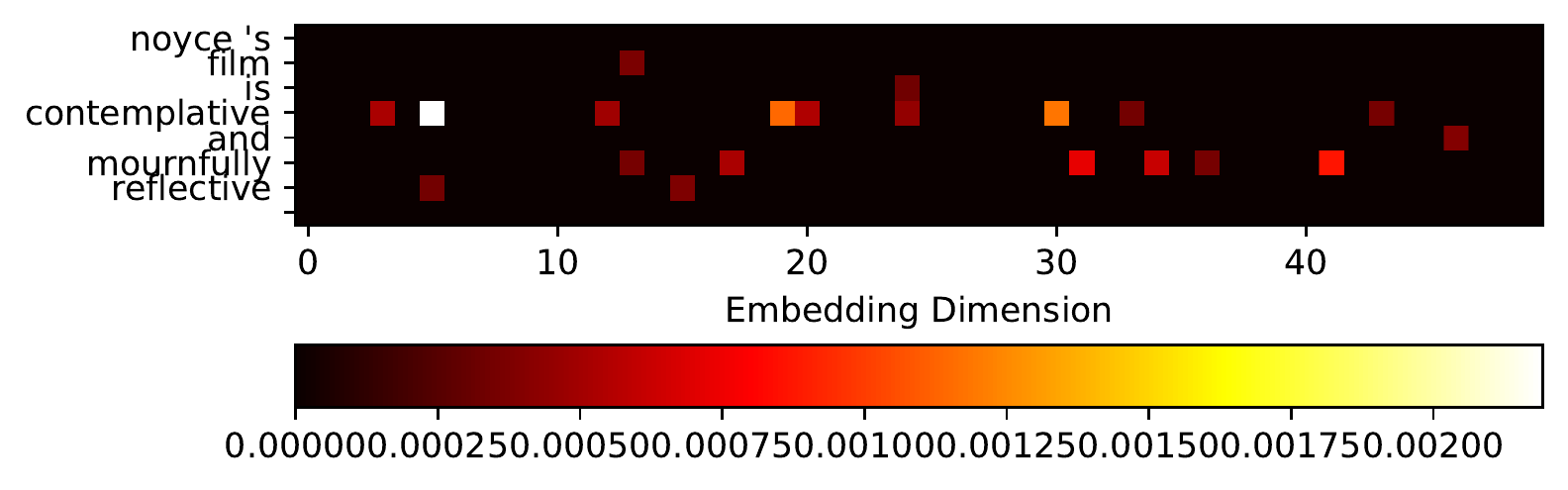} 
			\caption{Text example 3} \label{fig:144e}
		\end{subfigure}
		
		\caption{Examples for the sentence classification model.}  \label{fig:27}
	\end{figure}

	The Sentence Polarity Dataset \cite{Pang+Lee:05a} is a collection of movie-review documents labeled with respect to their overall sentiment polarity. Here, we will look at a negative and positive example (Fig. \ref{fig:144a} and Fig. \ref{fig:144b}) where the rows are the word tokens in the sentence, and the columns are the embedding dimensions. The NLP model used in our experiment is taken from \cite{kim2014convolutional} and trained on the dataset. As part of the pre-processing, the words in the dataset are tokenized and mapped to an embedding matrix.  The word embedding matrix is also used as the segments in our algorithm.  \cite{li2015visualizing} mentions that the saliency map of an NLP model can be visualized using the embedding layer similar to saliency maps used for image-based models. Consequently, one can apply our algorithm to NLP models in a similar manner, i.e., we can utilize the first order derivative of the loss with respect to the word embedding. This technique is similar to what was used in \cite{miyato2016adversarial}. The first example, ``it's a glorified sitcom, and a long, unfunny one at that.'' is classified as a negative review by the model. Fig. \ref{fig:144a} shows that the word ``unfunny'' is strongly highlighted as the main explanation for this prediction. For the positive example ``a work of astonishing delicacy and force,'' it is seen that the word ``astonishing'' has the most significant influence on  model's prediction. More examples are shown in Fig. \ref{fig:27}.
	\subsection{Benchmark tests}
\begin{figure}[h]
		\centering
		\begin{subfigure}[t]{0.14\textwidth}
			\includegraphics[width=\linewidth]{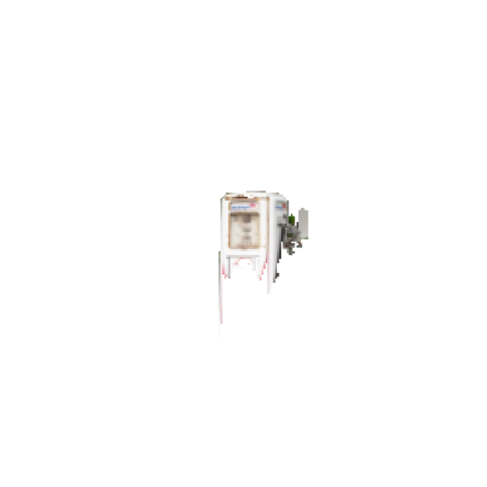} 
			\caption{} \label{fig:25a}
		\end{subfigure}
		\quad
		\begin{subfigure}[t]{0.14\textwidth}
			\includegraphics[width=\linewidth]{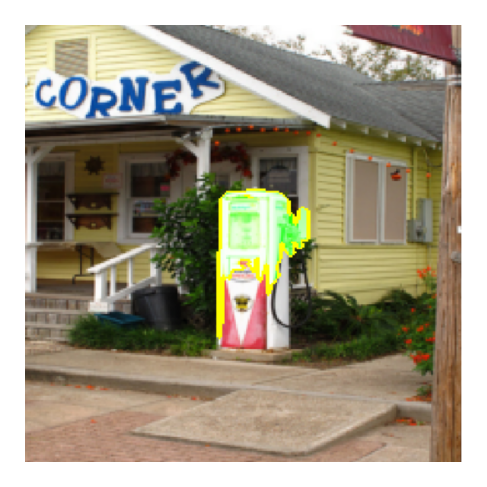} 
			\caption{} \label{fig:25b}
		\end{subfigure}
		\quad
		\begin{subfigure}[t]{0.14\textwidth}
			\includegraphics[width=\linewidth]{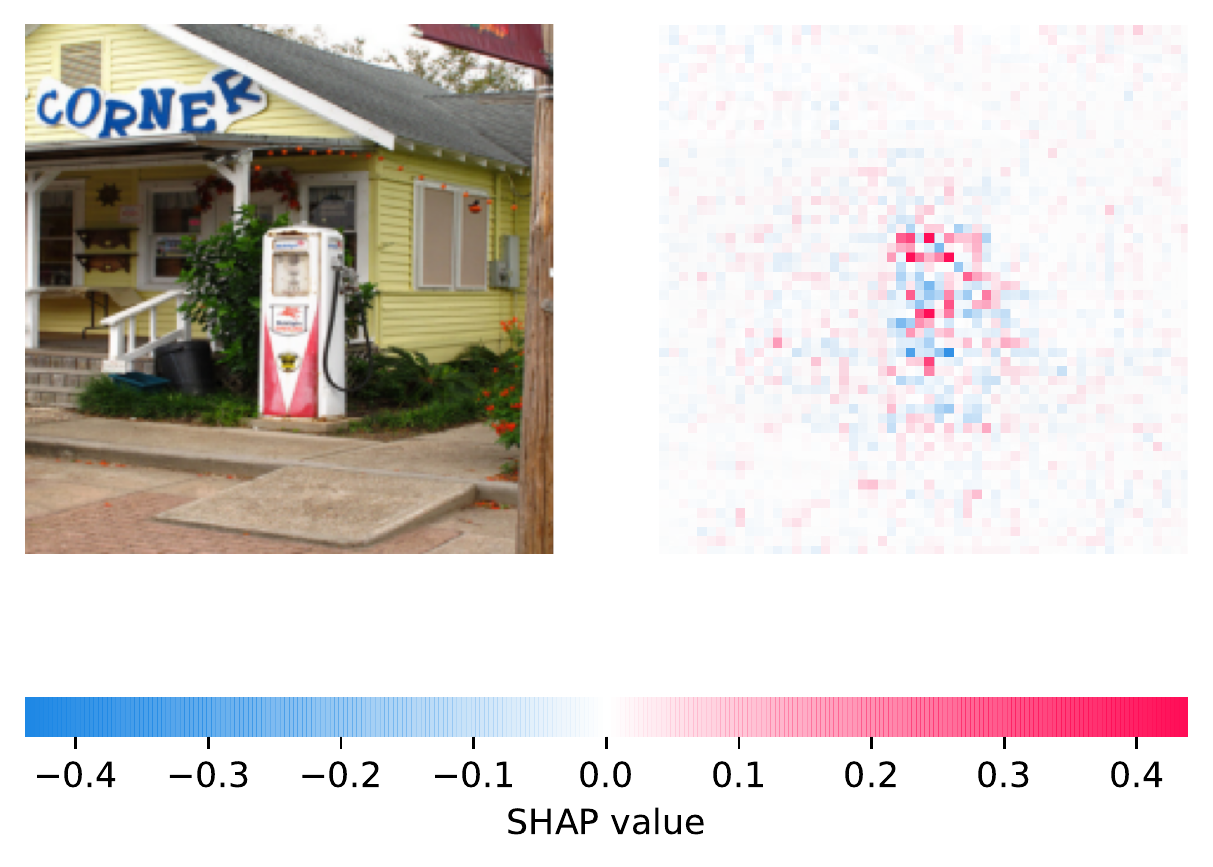} 
			\caption{}  \label{fig:25c}
		\end{subfigure}
		
		\centering
		\begin{subfigure}[t]{0.12\textwidth}
			\includegraphics[width=\linewidth]{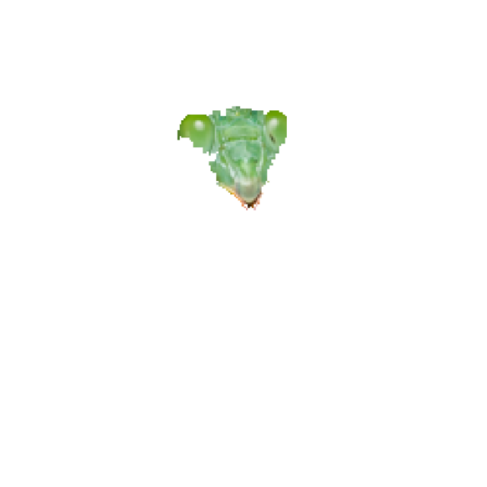} 
			\caption{} \label{fig:25d}
		\end{subfigure}
		\quad
		\begin{subfigure}[t]{0.12\textwidth}
			\includegraphics[width=\linewidth]{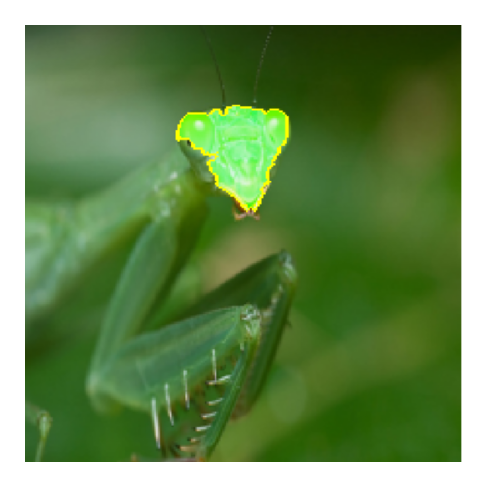} 
			\caption{} \label{fig:25e}
		\end{subfigure}
		\quad
		\begin{subfigure}[t]{0.12\textwidth}
			\includegraphics[width=\linewidth]{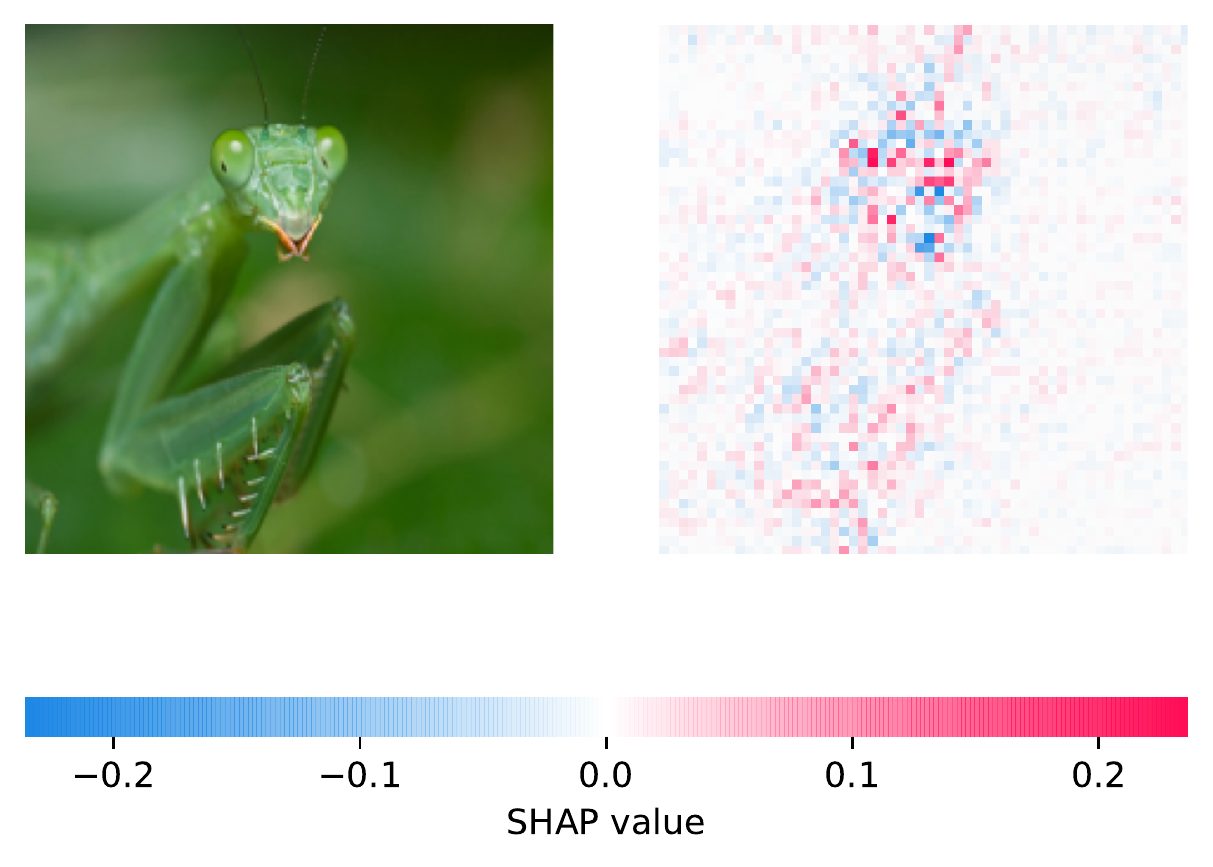} 
			\caption{} \label{fig:25f}
		\end{subfigure}
		
		\centering
		\begin{subfigure}[t]{0.12\textwidth}
			\includegraphics[width=\linewidth]{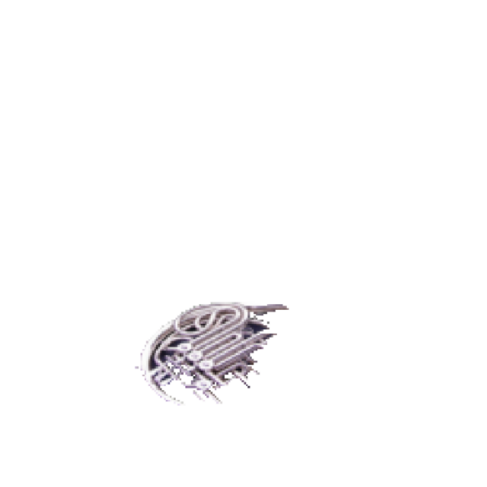} 
			\caption{} \label{fig:25g}
		\end{subfigure}
		\quad
		\begin{subfigure}[t]{0.12\textwidth}
			\includegraphics[width=\linewidth]{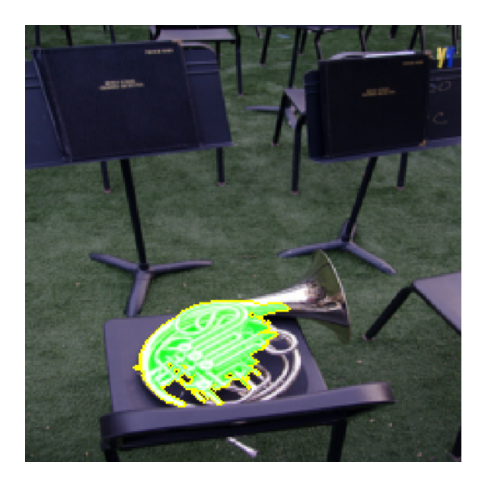} 
			\caption{} 	\label{fig:25h}
		\end{subfigure}
		\quad
		\begin{subfigure}[t]{0.12\textwidth}
			\includegraphics[width=\linewidth]{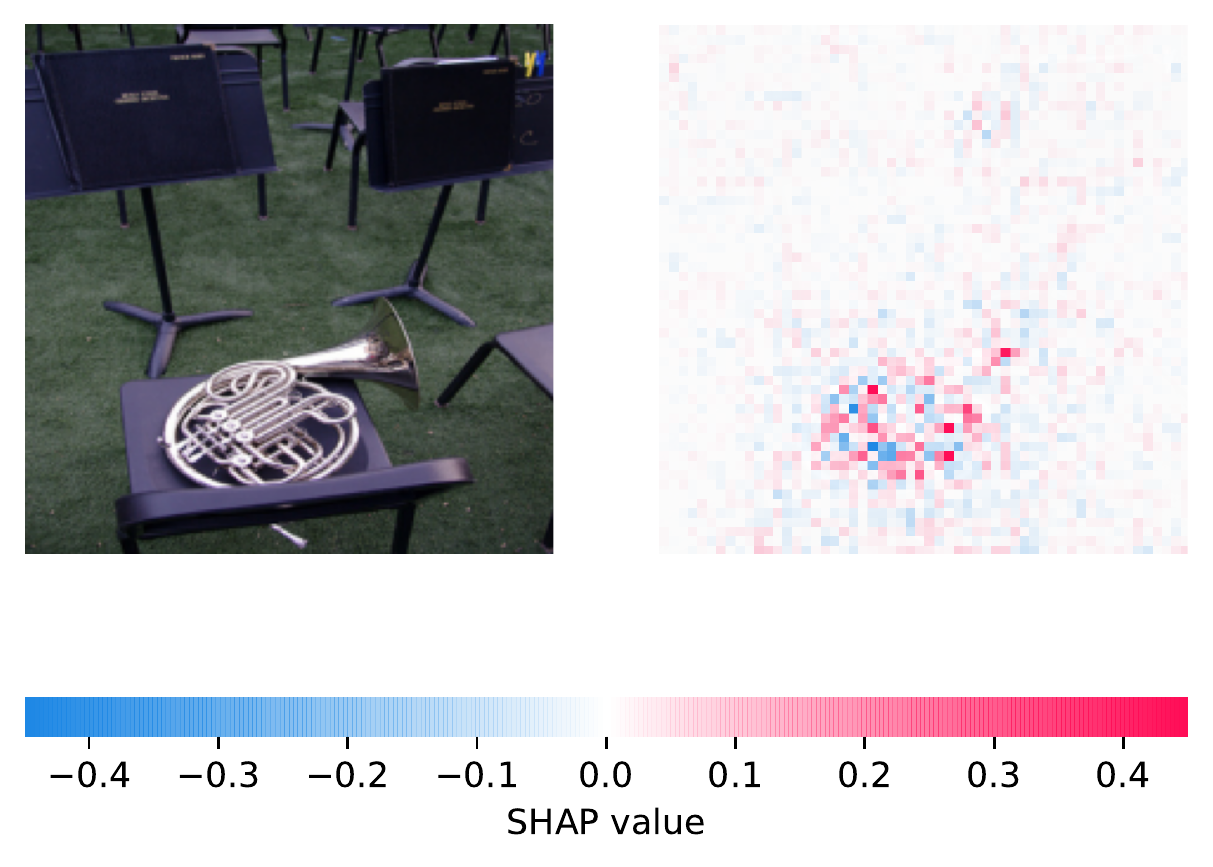} 
			\caption{} \label{fig:25i}
		\end{subfigure}
		\caption{Comparisons between our adversarial explainability approach (Left Column), LIME (Middle Column), and SHAP (Right Column). Explanations are produced for a ResNet34 trained on ImageNet.}  \label{fig:25}
	\end{figure}

	We test our algorithm against LIME and SHAP (Gradient Explainer). It is important to note that SHAP subsumes a number of prior approaches and provides a fair baseline. To show the consistency of our approach,  we present visualizations for 3 cases: 1) AlexNet, CIFAR10, 2) VGG16, CIFAR100, 3) ResNet34, ImageNet using the 3 explainability tools and provide more experiments in Appendix \ref{app_6}. The algorithms produce similar explanations where AXAI has fewer tune-able parameters and performs faster. LIME fails to produce good explanations for low-resolution CIFAR10 images. In Appendix \ref{app_6}, we provide examples showing that AXAI outperforms LIME for low-resolution inputs. We benchmark the running-time performance of AXAI, LIME and SHAP for ResNet34 trained on ImageNet on a single CPU (Intel Core i5-7360U) and single GPU (Tesla V100-SXM2) on the entire test dataset. The results are given in Table. \ref{tab:table44}. LIME is the slowest to produce explanations. This is because LIME needs to forward propagate the perturbed inputs through the DNN several times. SHAP is also slower to generate the results in comparison to AXAI. LIME works better on a GPU. AXAI maintains its  relative performance on the CPU and GPU. This is because the segmentation step which mainly uses the CPU is the main computational bottleneck for the algorithms (Appendix \ref{app_1}).  A few comparisons between AXAI, LIME, and SHAP are shown in Fig. \ref{fig:25}.
	\begin{table}[!]
		\centering
		\resizebox{1.\columnwidth}{!}{
			\begin{tabular}{|l|l|l|}
				\hline
				& Single CPU (Intel Core i5-7360U) & Single GPU (Tesla V100-SXM2) \\ \hline
				LIME                          & 105s                             & 5.8s                         \\ \hline
				SHAP (Gradient Explainer)     & 35s                              & 3.8s                         \\ \hline
				AXAI (PGD with 20 iters) & 6.6s                             & 1.7s                         \\ \hline
			\end{tabular}
		}
		\caption{Benchmark running-time experiments.}
		\label{tab:table44}
	\end{table}
	
	\begin{figure}[h]
		
		\centering
		\begin{subfigure}[!]{0.18\textwidth}
			\includegraphics[width=\linewidth]{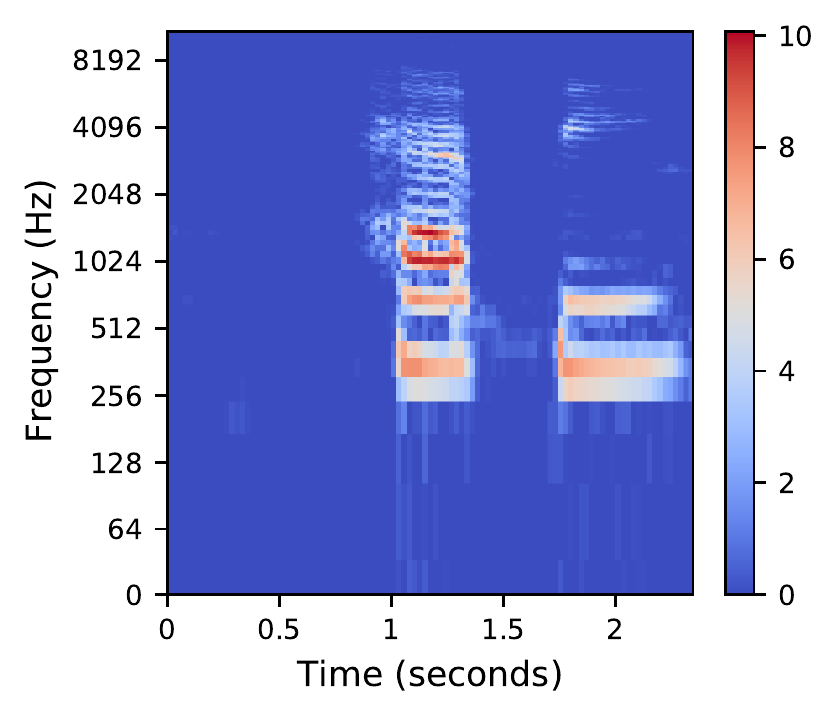} 
			\caption{"Happy"} \label{fig:13e}
		\end{subfigure}
		\quad
		\begin{subfigure}[!]{0.18\textwidth}
			\includegraphics[width=\linewidth]{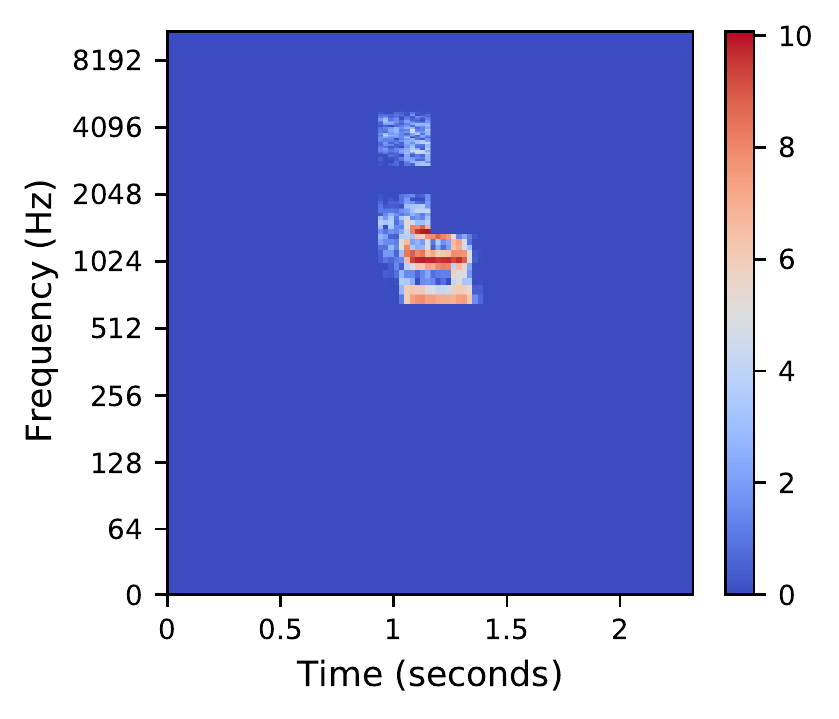} 
			\caption{Explanation} \label{fig:13f}
		\end{subfigure}
		\quad
		\begin{subfigure}[!]{0.18\textwidth}
			\includegraphics[width=\linewidth]{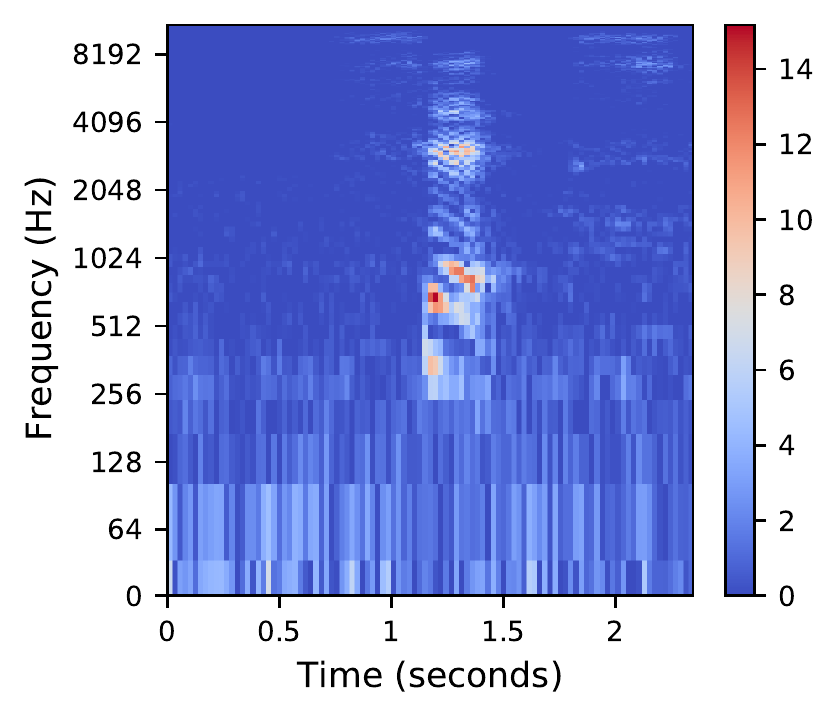} 
			\caption{"six"} \label{fig:13i}
		\end{subfigure}
		\quad
		\begin{subfigure}[!]{0.18\textwidth}
			\includegraphics[width=\linewidth]{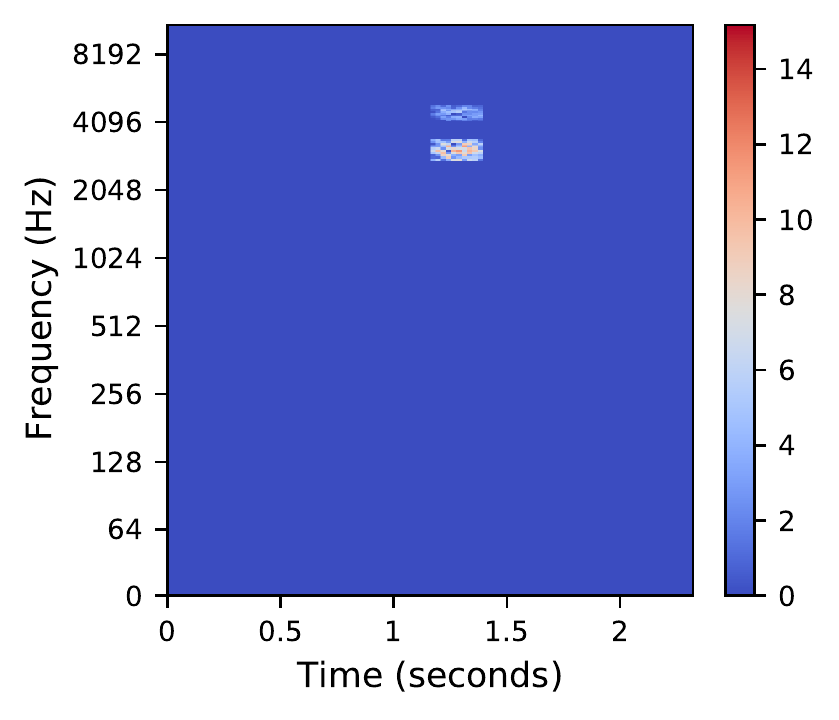} 
			\caption{Explanation} \label{fig:13j}
		\end{subfigure}
		\caption{Examples for the LeNet speech recognition model.}  \label{fig:26}
	\end{figure}
	
	\section{Final Remarks and Conclusion} \label{exp}
	In this paper, we proposed a new approach for explaining the predictions of DNNs. Interpretability is directly related to the readability of an explanation \cite{gilpin2018explaining}. An explanation relying on thousands of features is not interpretable. AXAI, similar to LIME, uses input segmentation to create human-readable explanations focused on important input features. Further, AXAI has the following properties,
	
	\textbf{Property 1 (Robustness):} Our approach is more robust to the changes in segmentation hyper-parameters in comparison to other segmentation based approaches such as LIME. This is because AXAI does not require a surrogate model trained on ``randomly perturbed inputs.'' AXAI uses the deterministic attack magnitudes as ``base explanations" for a given DNN and dataset, and uses segments as an ``aid'' to visualize the results. The segmentation affects the visualizations. We further explain this in Appendix \ref{app_1}. Robustness is identical to stability of explanations as defined in \cite{pub.1104451629}. A lower number of non-deterministic steps in the algorithm enhances stability. A carefully filtered explanation based on our approach simply removes the features that have a low impact on predictions. One can interpret this process as a de-noising step to create a sparse representation of explanations.
	
	\textbf{Property 2 (Local attribution):} Our algorithm is locally stable and uses local attributes to produce explanations. This is because an adversarial attack uses the most minimal amount of noise within an $\ell_2$ ball of some small $\epsilon$ to fool the DNN. Given the un-targeted nature of the attack used in AXAI, the distributions can be interpreted as estimations of the boundaries among neighboring classes. Thus, one can conclude that the attack magnitudes are a representation of feature contributions to the predictions on a local scale. A similar conclusion is made in \cite{ancona2017towards}, where it is argued that gradients can in fact point to important local attributions of a DNN. We explore this in details in Appendix \ref{app_4}.
	
	\textbf{Property 3 (Completeness):} Completeness as a property is described as the ability to accurately explain the operations of a DNN \cite{gilpin2018explaining}. An explanation is more complete when it can explain the behavior of the DNN for a larger set of inputs. \cite{sundararajan2017axiomatic} and \cite{smilkov2017smoothgrad} mention the problem of  sensitivity and lack of stability in gradient-based algorithms. In the literature, if a solution can reduce the gradient ``sensitivity'' problem, it can be described as having the “completeness” property \cite{gilpin2018explaining}. AXAI with PDG attack is complete in the same sense as SmoothGrad is \cite{smilkov2017smoothgrad}. SmoothGrad takes the average of saliency maps with added Gaussian noise to reduce sensitivity. The PGD attack behaves in a similar manner by adding adversarial noise at each iteration. Both solutions add perturbations to the input to smooth gradient fluctuations. While further research can be done on the power of iterative attacks in their gradient smoothing effects, we argue that AXAI with iterative PGD does have the desirable characteristic and produces stable sharpen visualizations of sensitivity maps for robust explanations.
	
	Lastly, as shown in Section \ref{sec:main}, our explainability algorithm exhibits a high-level of fidelity where the explainability outputs are both interpretable and also loyal to the decision making process of the DNN. The produced explainability segments directly point to the places in the input that affect the decision of the DNN. As a result, our solution can be used to explore the relationship between input features and predictions and to understand issues related to the training of DNNs, bias and robustness against adversarial attacks (Appendix \ref{app_5}).
	\section*{Potential Ethical Impact}
	Our work in this paper contributes to the fields of adversarial machine learning and  artificial intelligence explainability (AI Explainability). There is still a huge gap between building a model in Jupyter notebook and shipping it as a stand-alone product to the users. Advances in these two fields directly relate to the deployment of AI systems that behave in a robust and user-friendly manner after deployment. Building AI systems is hard. AI explainability can provide insights into how AI models behave, why they make the decision they make and the reasoning behind their incorrect predictions. Additionally, explaining the outcomes of a model can help reduce bias and contribute to improvements in accountability and ethics by providing beneficial insights into how AI models think and make their decisions. 
	
	Despite the hype, AI engineers struggle with deploying models which meet the users' performance expectations. A lack of robustness in the performance of trained model is a major impediment. We need to be able to design AI systems that both perform well and are robust. A robust model not only makes correct predictions in expected environment, but it also maintains an acceptable level of performance in unpredictable situations. Our work gives insights into how the adversary attacks an AI system trained to perform a specific task. Understanding how adversarial attacks behave can help AI engineers in development of AI systems that perform as expected while maintaining some level of robustness  in presence of external disturbances and adversarial noise. This type of information can help AI engineers in developing AI models that perform better. In short our paper can help AI researchers in their endeavor to design, develop and deploy explainable ethical AI systems that are robust and reliable.
	
	\begin{quote}
		\begin{small}
			\bibliographystyle{aaai}
			\bibliography{citations.bib}
		\end{small}
	\end{quote}
	
	\newpage
	\appendix
\onecolumn
\section{QuickShift Segmentation}  \label{app_1}
QuickShift is a mode seeking clustering algorithm proposed by  \cite{vedaldi2008quick}. QuickShift creates segments by repeatedly moving each data point to its closest neighbor point that has higher density calculated by a Parzen Estimator. The Kernel size argument in the QuickShift function controls the width of the gaussian kernel of the estimator. The path of moving points can be seen as a tree that connects data points. Eventually, the algorithm connects all data points into a single tree. To balance between under and over fragmentation of the image, a threshold, $\tau$, is served as a breaking point that limits the length of the branches in the QuickShift trees. The threshold, $\tau$, is the Max distance argument in the QuickShift function. Finally, the pre-processing step of QuickShift projects a given image into a 5D space, including color space $(r,g,b)$ and location $(x,y)$. A hyper-parameter, $\lambda$, takes a value between 0 and 1 and serves as a weight assigned to the color space, such that the feature space can be presented as $\{\lambda r, \lambda g, \lambda b, \lambda x, \lambda y\}$. 

\begin{figure}[htp]
	\centering
	\begin{subfigure}[t]{0.31\textwidth}
		\includegraphics[width=\linewidth]{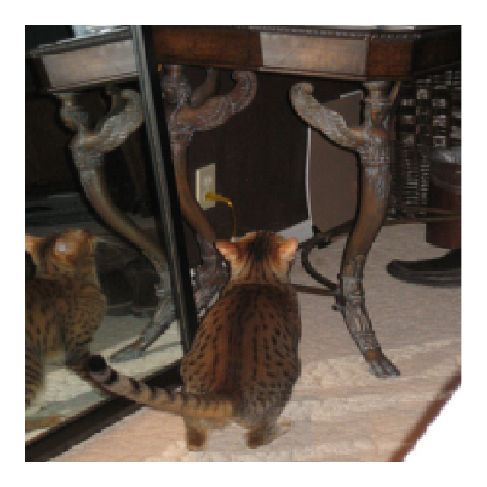} 
		\caption{Original image from CIFAR10} \label{fig:3a}
	\end{subfigure}
	\quad
	\begin{subfigure}[t]{0.31\textwidth}
		\includegraphics[width=\linewidth]{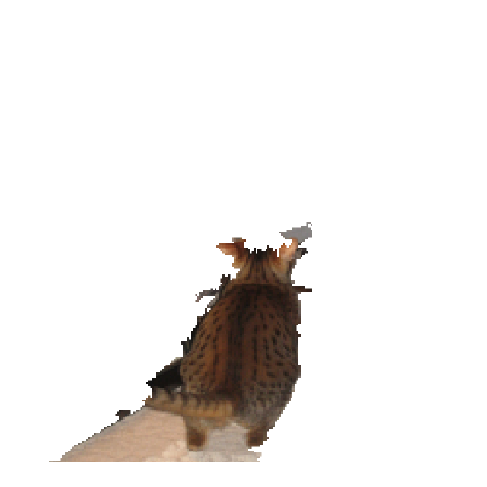} 
		\caption{Preferable segmentation} \label{fig:3b}
	\end{subfigure}
	\quad
	\begin{subfigure}[t]{0.31\textwidth}
		\includegraphics[width=\linewidth]{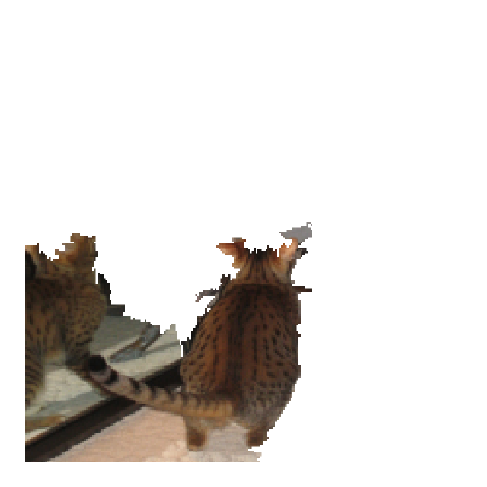} 
		\caption{Under segmentation} \label{fig:3c}
	\end{subfigure}
	\caption{Segmentation in low-resolution images.}  \label{fig:3}
\end{figure}

\begin{figure}[htp]
	\centering
	\begin{subfigure}[t]{0.31\textwidth}
		\includegraphics[width=\linewidth]{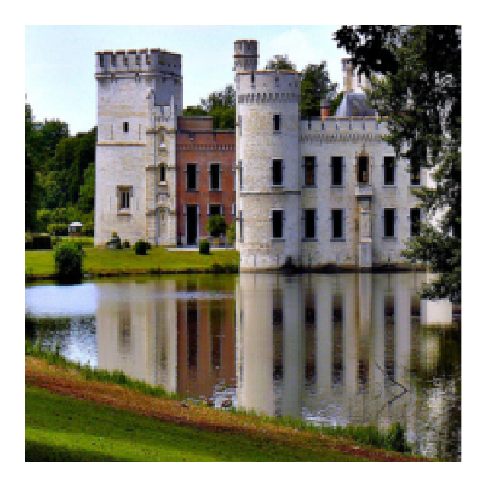} 
		\caption{Original image from ImageNet} \label{fig:4a}
	\end{subfigure}
	\quad
	\begin{subfigure}[t]{0.31\textwidth}
		\includegraphics[width=\linewidth]{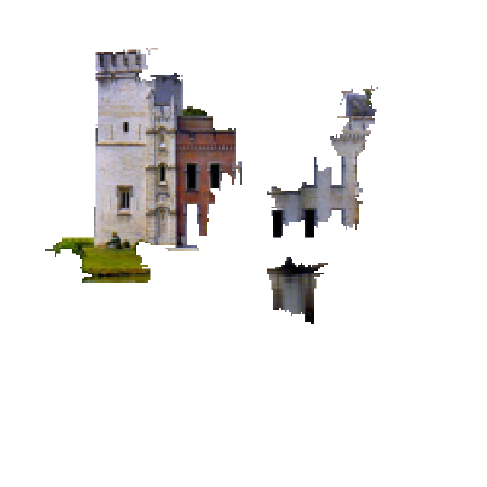} 
		\caption{Explainabilty length = 1} \label{fig:4b}
	\end{subfigure}
	\quad
	\begin{subfigure}[t]{0.31\textwidth}
		\includegraphics[width=\linewidth]{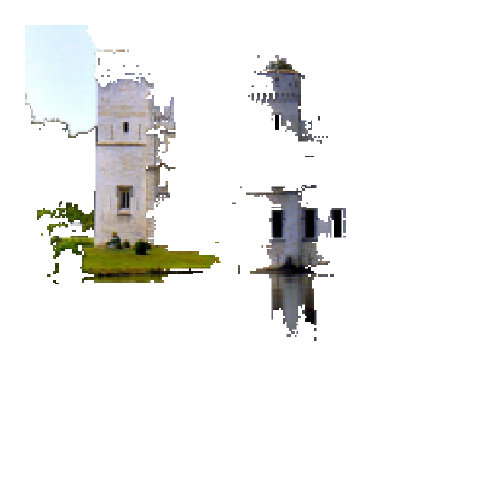} 
		\caption{Explainabilty length = 2} \label{fig:4c}
	\end{subfigure}
	\caption{Relationship between produced explanations and explainabilty length.}  \label{fig:4}
\end{figure}

LIME uses QuickShift for image segmentation where the default Kernel Size is 4, the Max distance is 200, and the threshold $\tau$  is 0.2. This combination prevents generating too many image segments. Even-though the image segmentation process is only performed once per image, we would like to point out that the parameter selection does change the explanation results slightly. First, increasing the kernel size increases the computation time while decreasing the number of image segments, making this parameter the major computational bottleneck in image segmentation. Second, extra care should be taken when it comes to low-resolution images, when the image is coarse and the number of image segments are low, because important and unimportant features can easily be merged together, as demonstrated in Fig. \ref{fig:3}. From the perspective of explainability, both accuracy and human-readability are needed. This is achieved as long as the important segments are not merged with unimportant ones. This problem can be solved by selecting a small kernel size. In our algorithm, we introduce a user tunable hyper-parameter, called explainabilty length, K, that allows users to decide the number of explainable segments. Human-readability is subjective, so we let the user decide the explainable length, Fig. \ref{fig:4}. We see that in Fig.  \ref{fig:4}, the wall of the castle on the left most side of the image is merged with the sky due to the similarity between colors.  In both case, we picked the top 10 segments as explanations, i.e., explainabilty length=10. It is important to note that unlike LIME and other explainability algorithms, the choice of a longer explainabilty length (more segments) does not increase the computational time of our algorithm.

\begin{figure}[htp]
	\centering
	\begin{subfigure}[t]{0.31\textwidth}
		\includegraphics[width=\linewidth]{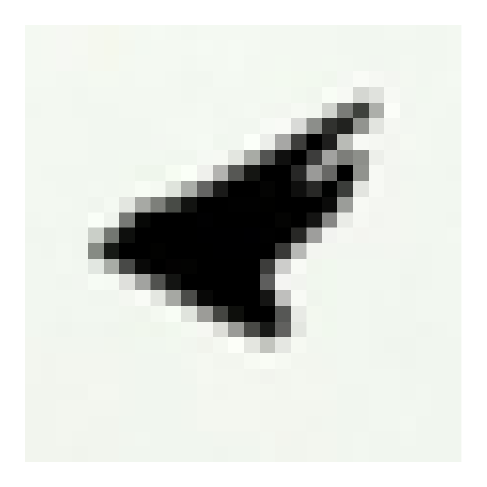} 
		\caption{Original image from ImageNet} \label{fig:2a}
	\end{subfigure}
	\quad
	\begin{subfigure}[t]{0.31\textwidth}
		\includegraphics[width=\linewidth]{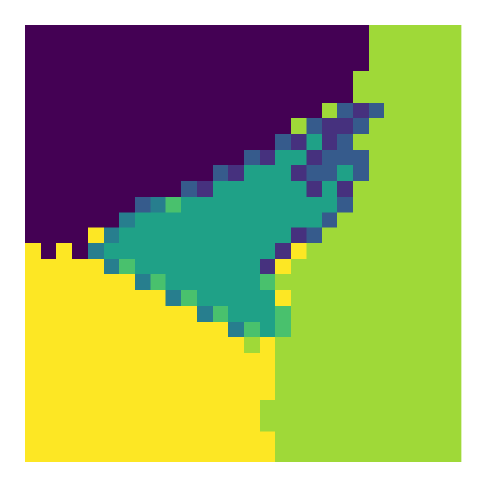} 
		\caption{$\lambda$=0.2} \label{fig:2b}
	\end{subfigure}
	\quad
	\begin{subfigure}[t]{0.31\textwidth}
		\includegraphics[width=\linewidth]{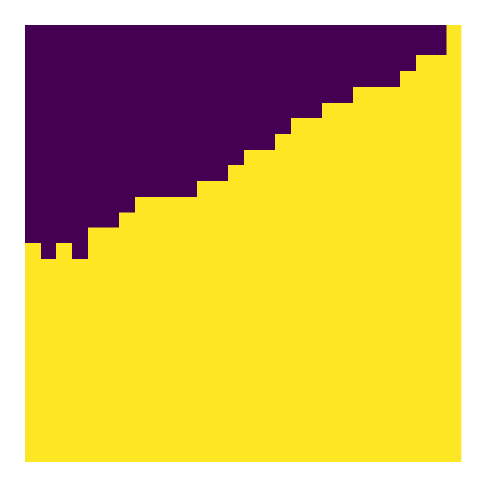} 
		\caption{$\lambda$=0.8} \label{fig:2c}
	\end{subfigure}
	\caption{The effects of $\lambda$ on the explanations produced.}  \label{fig:2}
\end{figure}

Deciding the tradeoff between the importance of the color (r,g,b) and spatial components (x,y) of the feature space, is especially important for high resolution images. Take a castle image in the ImgeNet dataset as an example (given in  Fig. \ref{fig:2}). We choose two different parameter combinations for comparison. The only difference between the two combinations is the $\lambda$ parameter. For the first combination, we used 0.2 (Fig. \ref{fig:2b}), for the second combination, we used 0.8 (Fig. \ref{fig:2c}). One can see that using a lower $\lambda$ prevents details from merging with irrelevant background information. In Fig. \ref{fig:2b} and Fig. \ref{fig:2c}, the total number of segments are nearly the same (73 and 81) but the explanations have different qualities.
\section{Convergence of Explanations across Adversarial Attacks}  \label{app_2}
\begin{figure}[htp]
	\centering
	\begin{subfigure}[t]{0.3\textwidth}
		\includegraphics[width=\linewidth]{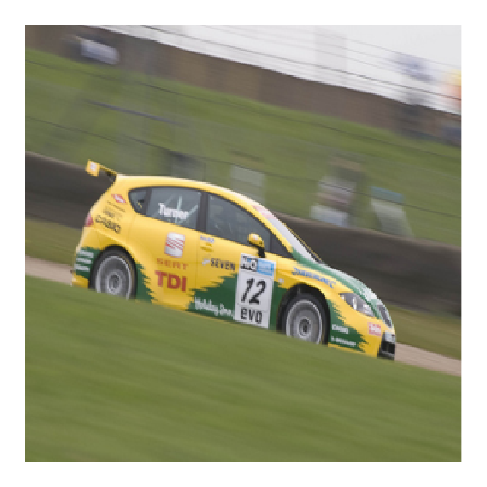} 
		\caption{Original Image from ImageNet} \label{fig:8f}
	\end{subfigure}
	\quad
	\begin{subfigure}[t]{0.3\textwidth}
		\includegraphics[width=\linewidth]{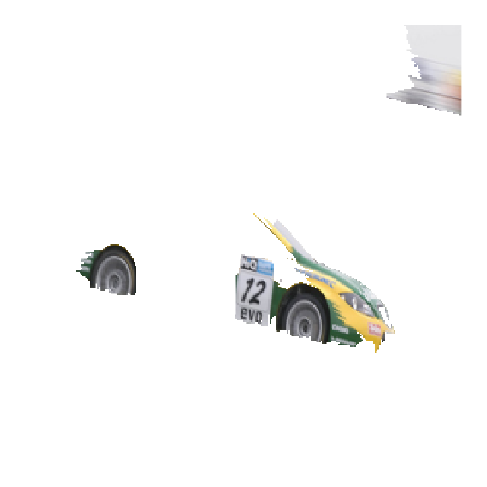} 
		\caption{$\ell_2$ FGM} \label{fig:8a}
	\end{subfigure}
	\quad
	\begin{subfigure}[t]{0.3\textwidth}
		\includegraphics[width=\linewidth]{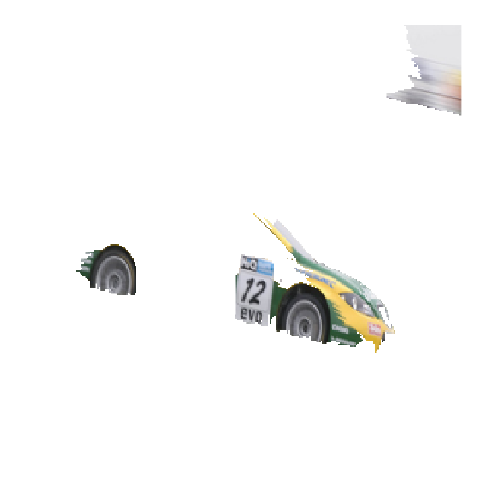} 
		\caption{PGD with 10 iterations} \label{fig:8b}
	\end{subfigure}
	\quad
	\begin{subfigure}[t]{0.3\textwidth}
		\includegraphics[width=\linewidth]{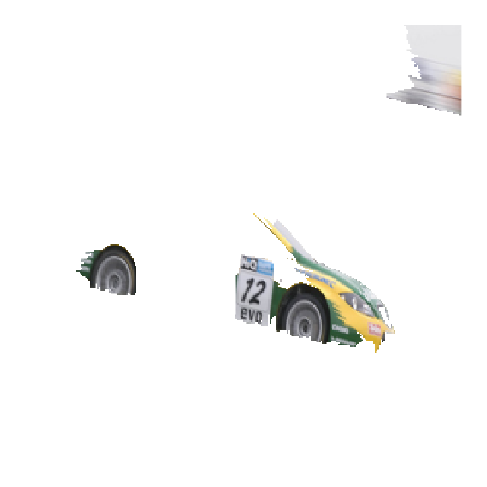} 
		\caption{PGD with 40 iterations} \label{fig:8c}
	\end{subfigure}
	\quad
	\begin{subfigure}[t]{0.3\textwidth}
		\includegraphics[width=\linewidth]{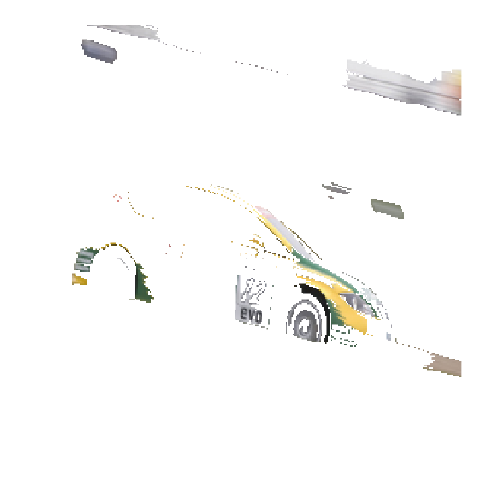} 
		\caption{PGD with 10 iterations} \label{fig:8d}
	\end{subfigure}
	\quad
	\begin{subfigure}[t]{0.3\textwidth}
		\includegraphics[width=\linewidth]{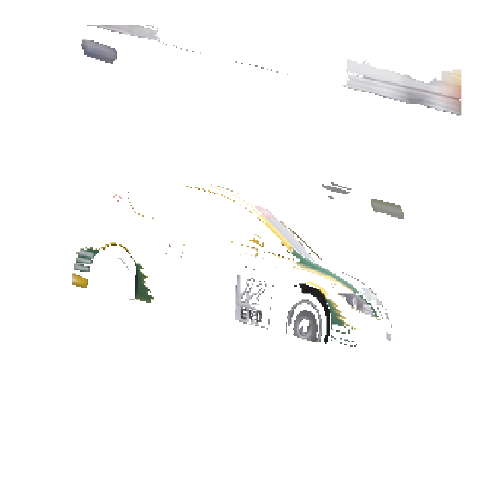} 
		\caption{PGD with 40 iterations} \label{fig:8e}
	\end{subfigure}
	\caption{Convergence of explanations for different adversarial attacks and number of segments (Architecture: ResNet34, Dataset: ImageNet).}  \label{fig:8}
\end{figure}

As a tool for explainability, efficiency, accuracy and consistency are of top priority. Our experiments show that $\ell_2$ PGD attacks  with different iterations create explanations similar to $\ell_2$ FGM attack. This points to consistency in explanations produced by our algorithm. PGD attack is an iterative version of FGM, while both attacks are subjected to an $\ell_2$ norm. Note that the distribution of the attacks can influence the explanation results. This also means that since the attack distributions of the first iteration and later iterations of the PGD attack are nearly identical, the overall explanations remain the same. In Fig. \ref{fig:8}, we provide an example from the ImageNet dataset to show the convergence of the attacks and consistency of  our explanations. Fig. \ref{fig:8a} shows the explanation results for an FGM based algorithm.  Fig. \ref{fig:8b}  and  Fig. \ref{fig:8c}  show the explanation results based on the PGD attack with different number of iterations. They both look exactly the same. This is because the slight changes on the attack distribution for different number of iterations, do not affect the overall density of pixel changes in each segment, thus the final explainability results do not change. This point to stability and consistency of our algorithm. To further explore the stability and consistency of our approach, we can segment the image into much smaller segments, as given in Fig. \ref{fig:8d}  and  Fig. \ref{fig:8e}, in this case using 50 times more segments than the previous case and then produce the explanations. In this case, we do see small differences between an explanation produced with a PGM attack with 10 iterations and one based on a PGM attack with 40 iterations. These small differences are caused by small differences in the attack distributions in each segment.  While it is interesting to further explore how different types of attacks can lead to more ``suitable'' explanations, it is important to note that one could explain the outcomes using our algorithm and with both types of attacks. Further, we can conclude that using FGM or PGD attacks in our algorithm satisfies consistency, accuracy and efficiency conditions for producing explanations.
\section{Further Details on the Statistical Analyses given in Subsection \ref{subsec:stat}}  \label{app_3}
\subsection{Further details on the statistical tests}
The Fisher-Pearson coefficient $g_1$ of a distribution $x$ with a sample size $N$ is calculated using the third moment $m_3$ and the second moment $m_2$ of the distribution,
\begin{equation}
g_1=\frac{m_3}{m_2^{\frac{3}{2}}},
\end{equation}
where,
\begin{equation}
m_i=\frac{1}{N} \sum_{n=1}^{N}(x[n]-\bar{x})^i 
\end{equation}

If skewness is 0, the data is perfectly symmetrical, if skewness is positive, then one interprets the distribution as skewed right, if skewness is negative, then the distribution is skewed left.  \cite{bulmer1979principles} pointed out that there are three levels of symmetricity, a) when skewness is between -0.5 to 0.5, the distribution is ``approximately symmetric,'' b) when skewness is within -1 and -0.5 or 0.5 and +1, the distribution is ``moderately skewed,'' c) when skewness falls out of the mentioned range, then the distribution is highly skewed. The Fisher-Pearson coefficient of all attack magnitudes are shown in Fig. \ref{fig:7}. It is seen that the skewness of all attack magnitudes falls within -0.5 an 0.5 showing the strong evidence that the distributions are approximately symmetric. 

The t-statistic test is represented as follows,
\begin{equation}
t=\frac{\bar{X_1}-\bar{X_2}}{s_p\sqrt{\frac{2}{ n}}}
\end{equation}
where,
\begin{equation}
s_p=\sqrt{\frac{s^2_{X_1}+s^2_{X_2}}{2}}
\end{equation}
Here $\bar{X_1}$, $\bar{X_2}$ and $s^2_{X_1}$, $s^2_{X_2}$ are the means and variances of the two distributions with size $n$.   The t-statistic can be interpreted as a kind of measurement for the ratio of the ``difference between groups'' over the ``difference within groups.'' Carrying out pair t-tests on all samples allows us to further be conservative on the similarity on means between the distributions. The results are shown in Table \ref{tab:table1}. Overall, there is no significant differences between the distributions. 

To show the similarity between the distributions produced for a dataset, we also use the one-way ANOVA test on all the samples to show that the means across different distributions are the same. Samples here are defined as intensity vs. frequency distributions for all adversarial test samples created by attacking a model trained on a specific dataset. For CIFAR10, we get the p-value of 0.9, and for a random subset of ImageNet test dataset we get the p-value of 0.94, indicating no significant differences between the distribution means. Similarly, a two-sample location t-test is used to determine if there is a significant difference between two groups where the null hypothesis is the equality of the means. Even-though ANOVA and t-tests are known for being robust on non-normal data, we further performed pair wise Mann–Whitney U test on all pair of distributions to test whether the mean ranks are similar. 

Mann–Whitney U test is a nonparametric test of the null hypothesis that two independent samples selected from population have the same distribution. The statistic U is calculated as following,
\begin{equation}
U_1=R_1-\frac{n_1(n_1-1)}{2},~
U_2=R_2-\frac{n_2(n_2)}{2}
\end{equation}
Where subscripts ``1'' and ``2'' denote the two distributions being compared. In the case of comparing two distributions ``sample 1'' and ``sample 2.'' One first combines ``sample 1'' and ``sample 2'' together to form an ordered set, and then one assigns ranks to the members of this set. Next, one adds up the ranks for the members of the set coming from ``sample 1'' and ``sample 2'' respectively. This is called the rank sum of $R_1$ and $R_2$. Once the rank sums are calculated The U statistic of the two distributions ($U_1$ and $U_2$) are calculated as above.  Finally, the U statistic is determined by the lower value between $U_1$ and $U_2$. If $U_1$ is lower than $U_2$, then $U_1$ is the U statistic of the Mann Whitney test between ``sample1'' and ``sample 2"'' and vice versa. We further perform the pair-wise Mann–Whitney U test on all pair of distributions to test whether the mean ranks are similar as well. If U is 0, it means that the two distributions are far away from each other where there are no overlaps between them. If the Rank sums are close enough, one can say the two distributions are highly overlapped. Thus, one can say the Mann–Whitney U test is a test comparing the Rank sums (or the mean ranks, calculated by dividing the Rank sums over the size of samples) of two distributions. The smaller values of $U_1$ and $U_2$ is the one used when consulting significance tables. 
\subsection{Quantile-Quantile plot}
Quantile-Quantile (Q-Q) plot allows us to show how the quantiles of a distribution deviates from a specified theoretical distribution. The theoretical distribution selected here is the normal distribution. Quantiles are cut points dividing the range of a probability distribution into continuous intervals with equal probabilities. A Q-Q plot is then a scatter-plot showing two sets of quantiles (a sample distribution and a theoretical distribution) against one another. The x-axis are the quantile values of the theoretical distribution while the y-axis are the quantile values of the sample distribution, i.e., the distribution of attack intensities vs. pixel frequencies. One can see that if the quantiles of the sample distribution perfectly match the theoretical quantiles, then one can see all the quantiles located on a straight line. While it is unlikely to have identical distributions that perfectly match the theoretical distribution, one can look at different sections of the Q-Q curves to distinguish the parts that two distributions share similarity and parts that they differ. Compared to a normal distribution, if the sample distribution has heavy or light tails, the Q-Q curve bends at the upper or lower portion based on side of the tails that deviates from the normal distribution. One can say that one purpose of Q-Q plots is to look at the “straightness” of the Q-Q curve. We took a subset that contains 1000 images from both ImageNet and CIFAR10 and plotted the distributions against a normal distribution as given in Fig. \ref{fig:11}. It is seen that all attack distributions plotted against the normal distribution have fairly straight lines at the middle portion of the Q-Q curve, while the curve bends at the upper part and the lower part. One can interpret this result as the attack magnitudes are similar to a normal distribution but differ in a way that the distributions have``heavy tails'' thus the upper part of the curve bends ``up'' and the lower part of the curve bends``down.''
\subsection{The beta distribution}
The beta distribution is a family of  distributions defined on the interval [a, b] parametrized by two positive shape parameters, denoted by $p$ and $q$. The general formula for the probability density function of the beta distribution can be written as,
\begin{equation}
f(x)=\frac{(x-a)^{p-1}(b-x)^{q-1}}{B(p,q)(b-x)^{p+q-1}}
\end{equation}
where,
\begin{equation}
B(\alpha,\beta) = \int_{0}^{1} {t^{\alpha-1}(1-t)^{\beta-1}dt} 
\end{equation}

The beta distribution is often used to describe different types of data, such as rainfall, traffic and financial data. In this paper, estimate the parameters of a beta distribution for our distributions. The method of moments estimation is employed to calculate the shape parameters, $p$,$q$, of the two-parameter beta distribution. As the interval [a, b] is known, the method of moments estimates of $p$ and $q$ are
\begin{equation} 
p = \bar{x}(\frac{\bar{x}(1 - \bar{x})}{s^2} - 1)
\end{equation}
\begin{equation} 
q = (1 - \bar{x})(\frac{\bar{x}(1 - \bar{x})}{s^2} - 1)
\end{equation}

When the interval [a, b] is [0, 1]. This is called the standard beta distribution. Since in most cases the interval [a, b] is not bounded between [0, 1], one can replace $\bar{x}$ with $\frac{\bar{x} - a}{b-a}$ and $s^2$ with $\frac{s^2}{(b-a)^2}$. Finally the estimated $p$ and $q$ of the beta distribution is listed in Table \ref{tab:table5}.
\subsection{Statistical analysis of distributions for DNNs with text or audio input types}
\begin{figure}[!]
	\centering
	\begin{subfigure}[!]{0.3\textwidth}
		\includegraphics[width=\linewidth]{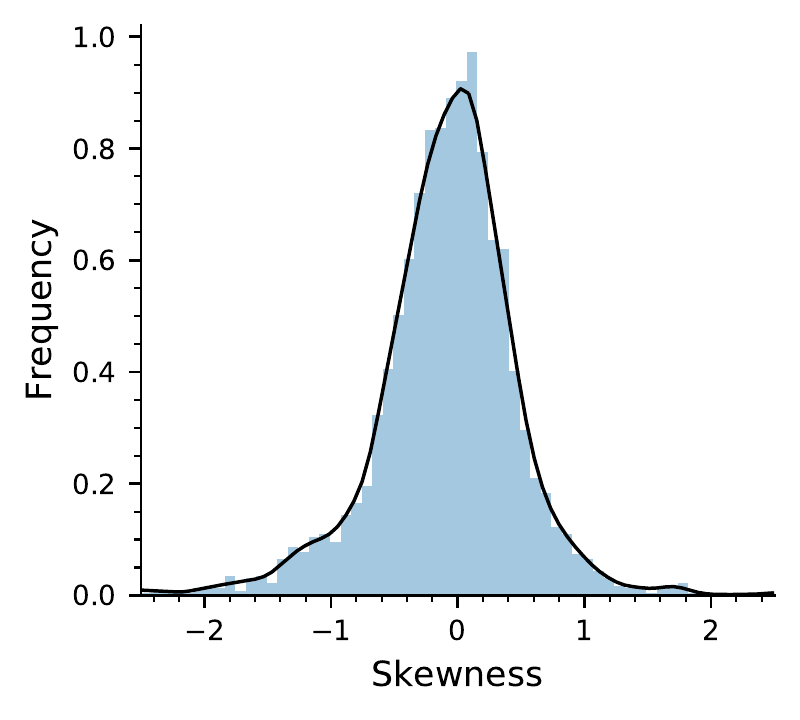} 
		\caption{PGD, LeNet, SpeechCommands} \label{fig:a10d}
	\end{subfigure}
	\quad
	\begin{subfigure}[!]{0.3\textwidth}
		\includegraphics[width=\linewidth]{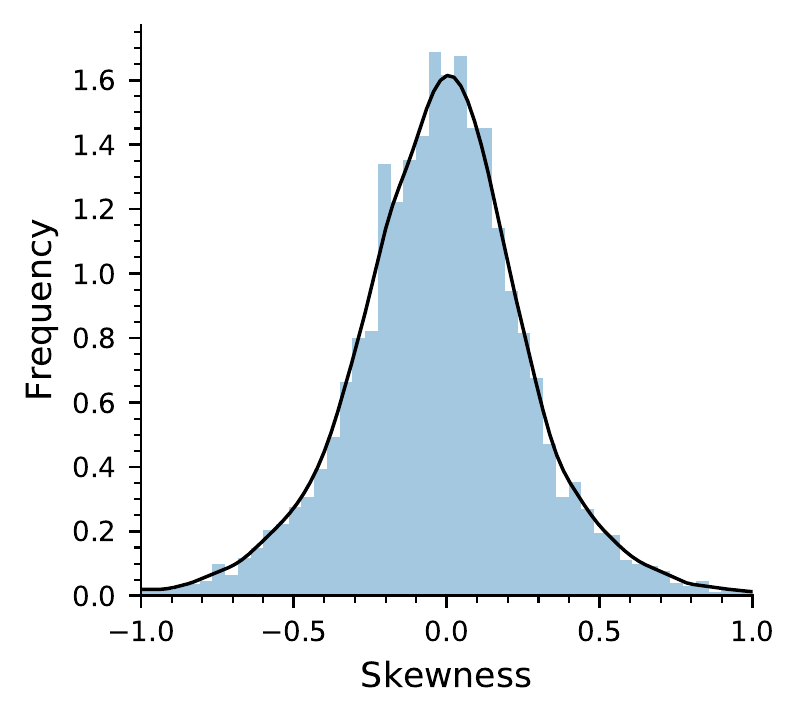} 
		\caption{PGD, CNN, Sentence Polarity} \label{fig:a10e}
	\end{subfigure}
	\caption{The Fisher-Pearson coefficient of attack magnitudes vs. frequency  distributions.}  \label{fig:a10}
\end{figure}

We test the symmetricity of distributions by calculating the Fisher-Pearson coefficient of skewness for LeNet trained on Speech Commands dataset, and  a convolutional neural network (CNN) given in \cite{kim2014convolutional} on Polarity dataset.  The Fisher-Pearson coefficients of the attack magnitudes vs. frequency  distributions for all 3 cases are shown in Fig. \ref{fig:a10}. It is seen that the skewness of all distributions falls within the $[-0.5, 0.5]$ range showing strong evidence that they are approximately symmetric \cite{bulmer1979principles}.

We perform the two-sample location t-test and Mann-Whitney U test to determine if there is a significant difference between two groups where the null hypothesis is the equality of the means. The results reported in Table \ref{tab:tablea3} indicate no significant difference between the means. Further, the Mann-Whitney U test results indicate that all pairs are similar to each other on the mean ranks. Under the assumption of two distributions having similar shapes, one could further state that Mann-Whitney test can be considered as a test of medians \cite{mcdonald2009handbook}. Since, we have shown that the shapes are similar, we can conclude that there are no significant difference between the medians of the distributions. 
\begin{table}[!]
	\centering
	\resizebox{.65\columnwidth}{!}{
		\begin{tabular}{|l|l|l|l|l|}
			\hline
			Dataset&\multicolumn{2}{c|}{LeNet, SpeechCommands, PGD}&\multicolumn{2}{c|}{CNN, Sentence Polarity, PGD}\\ \hline
			Test& t-test & Mann-Whitney& t-test & Mann-Whitne \\ \hline
			p-value 	&	0.30	&	0.25	&	0.47	&	0.42	\\ \hline
	\end{tabular}}
	\caption{p-values for the mean similarity statistical tests at significance level 0.05.}
	\label{tab:tablea3}
\end{table}

\begin{table}[]
	\centering
	\resizebox{.65\columnwidth}{!}{
		\begin{tabular}{|l|l|l|}
			\hline
			&  LeNet, SpeechCommands, PGD& CNN, Sentence Polarity, PGD \\ \hline
			15th Quantile &$ (-4.110e-3, -4.049e-3)$&$(-2.753e-1, -2.673e-1)$\\ \hline
			25th Quantile &$ (-1.150e-3, -1.109e-3)$&$(-1.472e-1, -1.414e-1)$\\ \hline
			Mean   &$(1.749e-5, 2.245e-5)$&$(-4.165e-3, -2.492e-3)$\\ \hline
			Median   &$(-4.181e-09, 1.356e-09)$&$(-2.142e-3, -6.219e-4)$\\ \hline
			75th Quantile &$(1.145e-3, 1.204e-3)$&$(1.365e-1, 1.421e-1)$\\ \hline
			85th Quantile&$(4.153e-3, 4.220e-3)$&$(2.599e-1, 2.677e-1)$\\ \hline
		\end{tabular}
	}
	\caption{Estimations for mean, median, 15th , 25th, 75th and 85th quantiles at 95\% confidence level.}
	\label{tab:tablea4}
\end{table}
\begin{table}[]
	\centering
	\resizebox{.65\columnwidth}{!}{
		\begin{tabular}{|l|l|l|l|l|l|}
			\hline
			&  LeNet, SpeechCommands, PGD & CNN, Sentence Polarity, PGD \\ \hline
			$p$ & $(5.282e+1, 5.451e+1)$ & $(1.322e+1, 1.368e+1)$ \\ \hline
			$q$  & $(5.144e+1, 5.309e+1)$ & $(1.346e+1, 1.393e+1)$ \\ \hline
		\end{tabular}
	}
	\caption{Statistical estimations for parameters of  beta distribution at 95\% confidence level.}
	\label{tab:tablea5}
\end{table}
Next, to show consistency across distributions for a given model, dataset and attack, we estimate the values of quantiles, means and medians. We do this by estimating the statistics of the distributions and constructing confidences intervals. For each experiment, we estimate the mean, median, 15th, 25th, 75th and 85th quantiles of each attack magnitude vs. frequency distribution for the entire test dataset. The statistical confidence interval estimations at confidence level of $95\%$ are reported in Table \ref{tab:tablea4}. Our results show that the confidence intervals have narrow ranges and the estimations are consistent. The estimates for the 15th, 25th, 75th and 85th quantiles indicate a strong  symmetricity with respect to the origin in all cases. Another observation is that the confidence interval of the mean and medians are pretty narrow, supporting the results of the t-tests and Mann-Whitney U test. Finally, we can  show with high confidence that the distributions consistently follow a beta distribution. The beta distribution is a family of  distributions defined by two positive shape parameters, denoted by $p$ and $q$.  The estimated $p$ and $q$ of the beta distribution are reported in Table \ref{tab:tablea5}.

\section{Explanations and Class Boundaries}  \label{app_4}
\begin{figure}[htp]
	\centering
	\begin{subfigure}[t]{0.31\textwidth}
		\includegraphics[width=\linewidth]{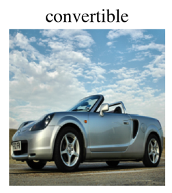} 
		\caption{} \label{fig:14a}
	\end{subfigure}
	\quad
	\begin{subfigure}[t]{0.31\textwidth}
		\includegraphics[width=\linewidth]{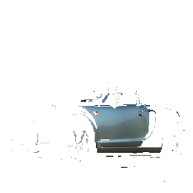} 
		\caption{} \label{fig:14b}
	\end{subfigure}
	\quad
	\begin{subfigure}[t]{0.31\textwidth}
		\includegraphics[width=\linewidth]{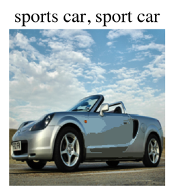} 
		\caption{}  \label{fig:14c}
		
	\end{subfigure}
	\caption{Left: Original image sample from ImageNet, Middle: The intensely attacked segments. Right: The original image with the explainable parts, i.e., the doors, blurred.}  \label{fig:14}
\end{figure}

Explaining how important features affect the predictions made by the model depends on the set of classes the model was trained to predict. Un-targeted attacks change the prediction label of an input to the label of its closest neighbor. Based on the different datasets that a model may have been trained on, the label changes after attack may be significantly different. For example, given an image of a ``Beagle'' and a model that is trained on a dataset consisting of labels \{Cats and Dogs\}, after attacking the model, the label of the image can change from ``Dog'' to ``Cat.'' But if the same model is trained on a dataset composing of ``Beagle, Golden retriever, and Egyptian Cat'', the label of the image can change from ``Beagle'' to ``Golden retriever,'' which is a more granule change. When an image is attacked, the features of the image will be directed to the nearest class with a similar probability distribution in the decision layer. Let’s look at an example from ImageNet where the input image is classified as a ``convertible" by ResNet34 trained on ImageNet (given in Fig. \ref{fig:14}). There are multiple classes such as minivan, sports car, race car etc., under the ``car" category in ImageNet. After attacking the model, the label changes from ``convertible" to ``sports car.'' This indicates that ``sports car'' may be the nearest neighbor class to the ``convertible'' class. If we look at the produced explanations we see that segments including the door are intensely attacked as given in Fig. \ref{fig:14b}.  The fact is that the model thinks that the doors are the ‘most’ important features for switching the label from ``convertible'' to ``sports car.''  Both classes, ``convertible''  and ``sports car,'' have similar wheels but different doors. In order to fool the model, attacking the wheels is not of top priority, it’s the doors that makes the difference between two classes. The fact is that the model thinks that the doors are the most important features for classifying the original image as ``convertible'' and not ``sports car.'' Both classes, have similar wheels but different doors. In order to fool the model, attacking the wheels is not of top priority, it’s the doors that make the difference between two classes. After bluring the segments of interest to the model, i.e. the door segment\textemdash Fig. \ref{fig:14c}, and feeding the image to the model, the predicted label changes from ``convertible'' to ``sports car'' which proves that the doors are the major features supporting the predictions made by the model. Using adversarial attacks as the force behind producing the explanations helps with finding the important features that are not only globally important to the model (doors are important features of cars, other classes do not have doors similar to cars), but also locally important to the model (within the car class, doors are the important features that make a difference between a convertible and a sports car).

There are also some explainable features that humans hardly understand but models do, these can be called ``non-robust features." \cite{tsipras2018robustness} introduced the concept of robust and non-robust features, where the authors indicated that there are features that humans ignore but the models are sensitive to. They call these the non-robust features. Non-robust features are the features can easily be manipulated by the attacker in order to fool the model. Robust features are features that are both important to the model and also humans and at the same time invincible to small adversarial manipulations. 
\section{Further Experiment Results} \label{app_5}
\subsection{Explaining an image classification model}
\begin{figure}[htp]
	\centering
	\begin{subfigure}[t]{0.22\textwidth}
		\includegraphics[width=\linewidth]{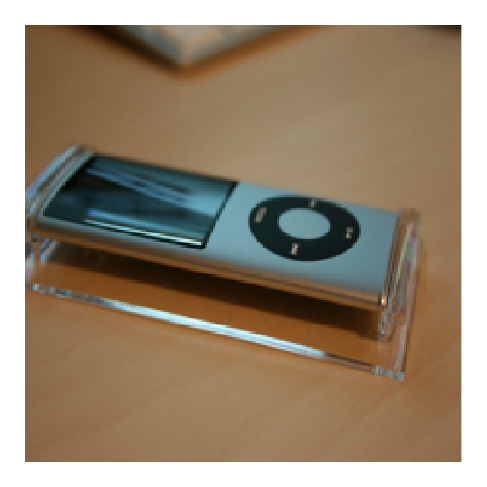} 
		\caption{An image of an iPod} \label{fig:5a}
	\end{subfigure}
	\quad
	\begin{subfigure}[t]{0.22\textwidth}
		\includegraphics[width=\linewidth]{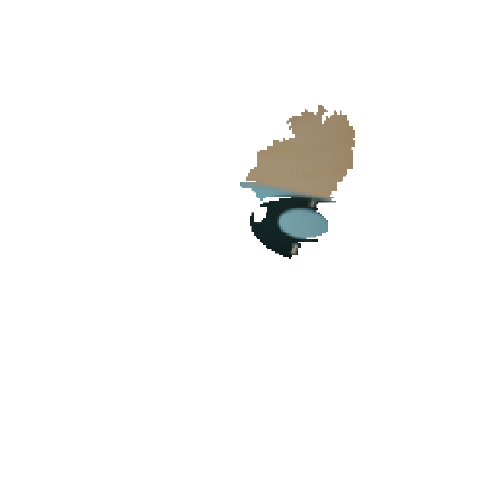} 
		\caption{Explanation} \label{fig:5b}
	\end{subfigure}
	\quad
	\begin{subfigure}[t]{0.22\textwidth}
		\includegraphics[width=\linewidth]{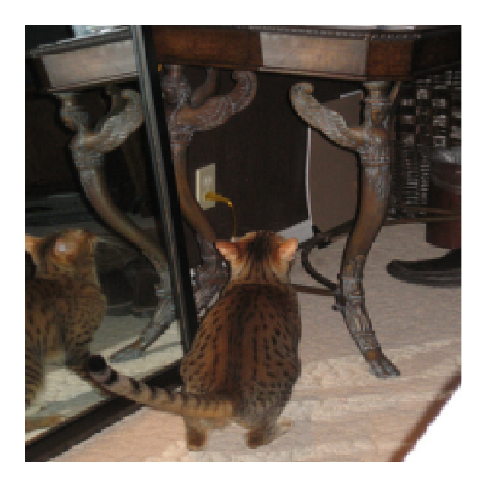} 
		\caption{An image of a cat} \label{fig:5c}
	\end{subfigure}
	\quad
	\begin{subfigure}[t]{0.22\textwidth}
		\includegraphics[width=\linewidth]{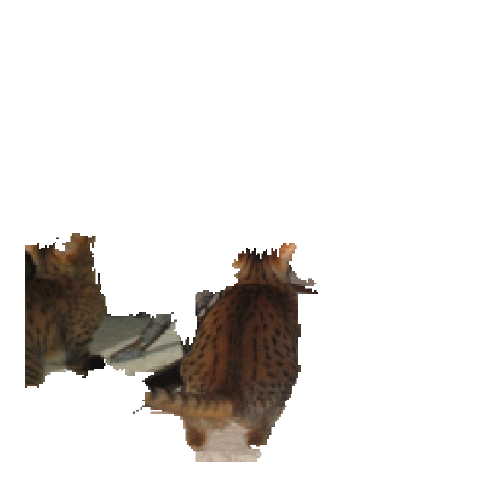} 
		\caption{Explanation} \label{fig:5d}
	\end{subfigure}
	\caption{The explanation results for a ResNet34 image classification model trained on ImageNet.}  \label{fig:5}
\end{figure}

Fig. \ref{fig:5} shows two examples of  the explanations produced using AXAI for image samples from ImageNet  \cite{deng2009imagenet} test dataset for a Resnet34 trained on ImageNet training dataset. In the first example, Fig. \ref{fig:5a}, the explanation results clearly show that the round control panel on an iPod is an important feature that helps the model identify an IPod in the image. The second example,  Fig. \ref{fig:5c}, shows how the model recognizes that there are two cats in the image (one is the reflection of the cat in the mirror).

\begin{figure}[htp]
	\centering
	\begin{subfigure}[t]{0.22\textwidth}
		\includegraphics[width=\linewidth]{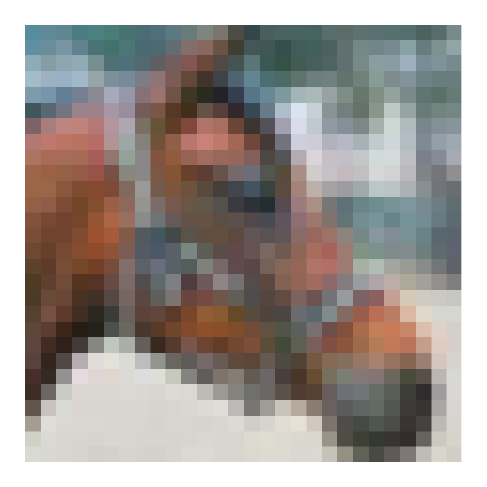} 
		\caption{An image of a horse} \label{fig:6a}
	\end{subfigure}
	\quad
	\begin{subfigure}[t]{0.22\textwidth}
		\includegraphics[width=\linewidth]{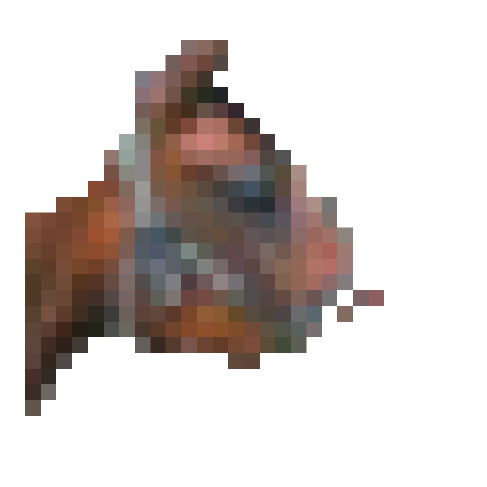} 
		\caption{Explanation} \label{fig:6b}
	\end{subfigure}
	\quad
	\begin{subfigure}[t]{0.22\textwidth}
		\includegraphics[width=\linewidth]{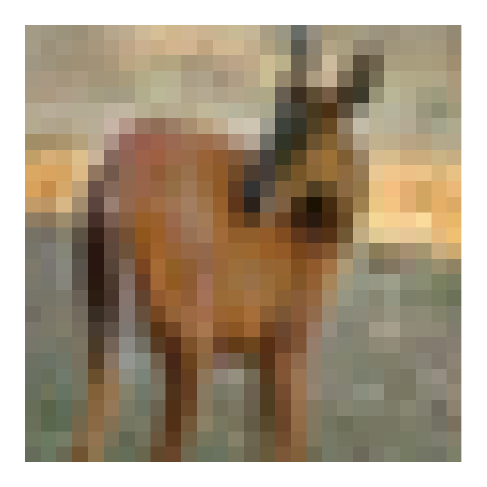} 
		\caption{An image of a deer} \label{fig:6c}
	\end{subfigure}
	\quad
	\begin{subfigure}[t]{0.22\textwidth}
		\includegraphics[width=\linewidth]{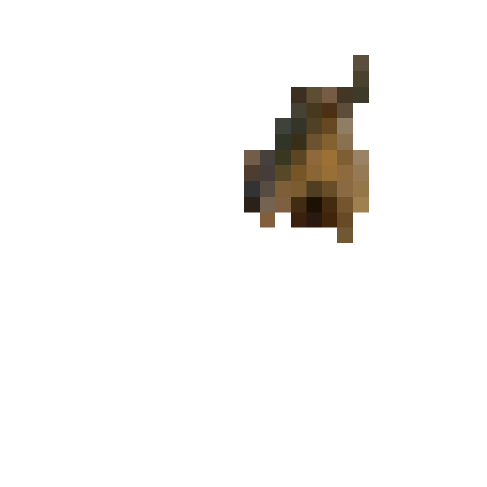} 
		\caption{Explanation} \label{fig:6d}
	\end{subfigure}
	\caption{The explanation results for an AlexNet image classification model trained on CIFAR10.}  \label{fig:6}
\end{figure}

CIFAR10 dataset \cite{kaur2018convolutional} consists of images of size $32\times32$ pixels, compared to ImageNet, these images are low-resolution images. Fig. \ref{fig:6} shows the explanations produced by AXAI  for sample images from CIFAR10 dataset for an AlexNet image classification model trained on CIFAR10 training dataset. For CIFAR10, our explanations clearly separate the background and capture the target object. The explanation given in Fig. \ref{fig:6b} shows that the head of the horse with the leather halter is recognized by the model, and the white fence behind the horse is completely ignored by the model. This indicates that the model is well-trained. Similarly in Fig. \ref{fig:6d} the ear and head of deer in the image helps the model to classify the image correctly into the deer class. Images from CIFAR10 dataset are easily explained due to the nature of the dataset with most objects in the images being located in the middle of the image and the lack of noisy background in most images.
\subsection{Explaining an object detection model}
\begin{figure}[htp]
	\centering
	\begin{subfigure}[t]{0.22\textwidth}
		\includegraphics[width=\linewidth]{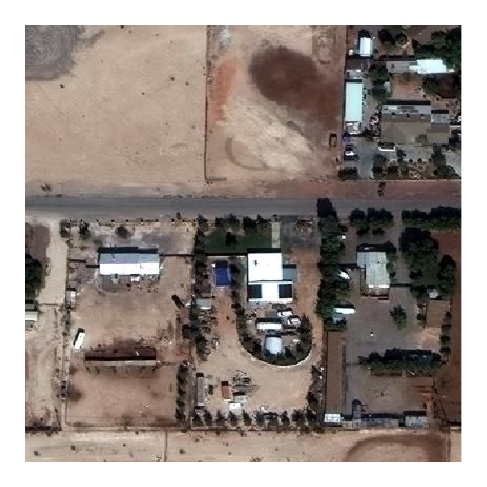} 
		\caption{Image sample 1} \label{fig:7a}
	\end{subfigure}
	\quad
	\begin{subfigure}[t]{0.22\textwidth}
		\includegraphics[width=\linewidth]{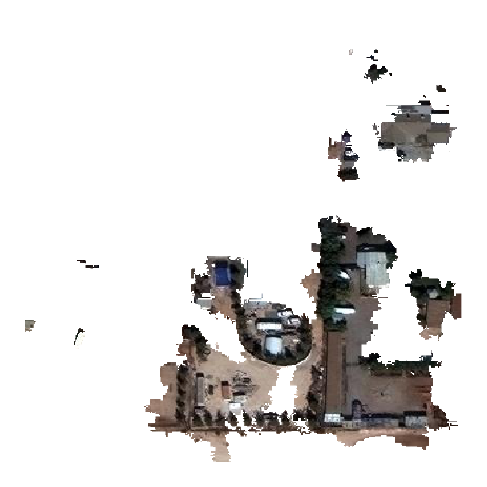} 
		\caption{Explanation} \label{fig:7b}
	\end{subfigure}
	\quad
	\begin{subfigure}[t]{0.22\textwidth}
		\includegraphics[width=\linewidth]{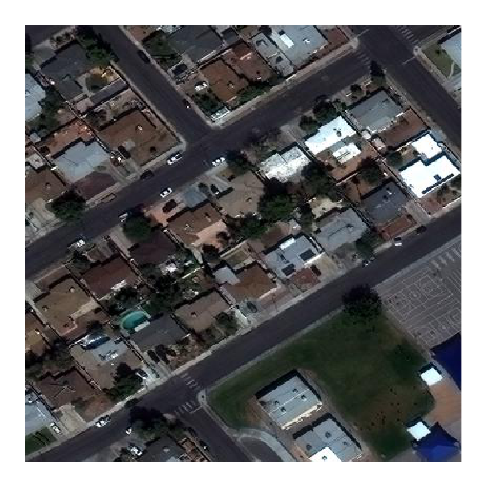} 
		\caption{Image sample 2} \label{fig:7c}
	\end{subfigure}
	\quad
	\begin{subfigure}[t]{0.22\textwidth}
		\includegraphics[width=\linewidth]{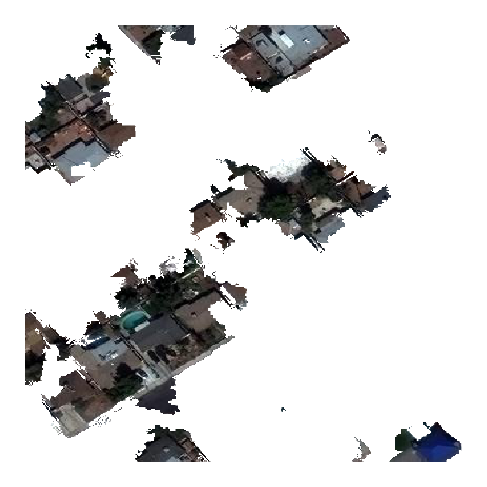} 
		\caption{Explanation} \label{fig:7d}
	\end{subfigure}
	\caption{The explanation results for a YOLOv3 object detection model trained on SpaceNet Building Dataset .}  \label{fig:7}
\end{figure}

We present two examples of explanations produced by our algorithm for a YOLOv3 object detection model trained on the SpaceNet Building Dataset \cite{van2018spacenet} to detect buildings in overhead imagery. The produced explanation  are clearly focused on areas where buildings are located and ignore empty spaces in the images such as the top left corner of Fig. \ref{fig:7b}. Further, as seen in Fig. \ref{fig:7d}, the roads are ignored and only buildings and their contours affect the predictions made by the object detector.

\subsection{Further details on the speech recognition experiment}
The Speech Commands Dataset \cite{warden2018speech} is an audio dataset of short spoken words, such as ``Right,'' ``Three,'' ``Bed,'' etc. The audio files are converted to spectograms and are used to train a LeNet for a command recognition task. A spectrogram is a visual representation of the spectrum of frequencies of a signal as it varies with time. Fig. \ref{fig:13a} is an example of a spectrogram. The y-axis, the frequency, of the spectograms are presented on a log-scale, the x-axis represent the time-scale, and the color bar shows the magnitude. Fig. \ref{fig:13a} is the frequency spectrum of a human speaking the word ``Right.'' It is seen that in the time interval 0.4s to 1.1s, high magnitude is presented in the spectrum. In other words, the speaker pronounces the word "Right" around  0.4s to 1.1s into the recorded audio file. This is how one reads a spectrogram. Our explainable solution uses audio files as input, converts them into spectrograms, and then generates the corresponding explanations. So if one feeds AXAI with an audio file of a human speaking ``Right,'' AXAI first transforms the audio into a spectrogram shown in Fig. \ref{fig:13a}, and produces the explanations in Fig. \ref{fig:13b}. The explanation will have the exact same scale as the input, and simply masks out the unimportant parts of the spectrogram. To read the explanations, one can refer to the original spectrogram input Fig. \ref{fig:13a} and find where the audio is located in the spectrogram (for example looking at the magnitudes), and then look at the corresponding location of the explanations in Fig. \ref{fig:13b}. 

The explanations of two examples are presented in Fig. \ref{fig:13}. The spectrogram of the first example ``Right'' and its explanation are shown in Fig. \ref{fig:13a} and Fig. \ref{fig:13b}. One can see from Fig. \ref{fig:13a} that the spoken word ``'Right'' appears between 0.4s to 1.1s in the spectrogram of the audio file. If one looks at its corresponding explanation, it is seen that only time-intervals of 0.4s to 0.5s, 0.5s to 0.6s and 1.0s to 1.2s are not masked out by AXAI. This means that these intervals in the audio have great importance for the prediction made by the model. if we look back at Fig. \ref{fig:13a}, one then realizes that the explanation shows that the first few and the last few seconds of the spoken word ``Right'' are important to the model, and the middle part is not. Why is that? The neighboring class of  ``Right'' is ``Five.'' ``Right'' and ``Five'' differ in how "R" \& "F" and "t" \& "ve" are pronounced. The middle part of ``Five'' and ``Right'' is highly similar and does not affect the model's prediction on deciding whether the spoken word is ``Five'' or ``Right.'' The second example is ``Three.'' As seen in the spectrogram, Fig. \ref{fig:13c}, ``Three'' is expressed around the time-interval 1.4s to 2.2s in the spectrogram of the audio file. The corresponding explanation is shown in Fig. \ref{fig:13d}. The explanation masks out almost everywhere except 1.4s to 1.6s and a small part in 1.6s to 1.7s and 1.9s to 2.2s. Now, let's look at the original spectrgram of ``Three'' and understand what the explanation means. Since The explanation highlights 1.4s to 1.6s, which is the first few seconds of the spoken word. To understand why, one can learn that if we attack the model, then ``Three'' is miss-classified as ``Tree.''  This indicates that the model has  learned to recognize ``Three'' and not ``Three''  by learning the difference between ``Thr'' and ``Tr.'' The explanation tells us that the first few seconds of the audio are important (the utterance of ``Thr'').
\subsection{Ablation study}
If a feature or a group of features is important to a model, then completely removing those features from the input would decrease the probability of a correct prediction. Accordingly, we performed an ablation study confirming  that the explanations produced by AXAI contain important features. This ablation method can be used to test the accuracy of an explainability solution. If the generated explanation is faithful to the model, then removing the explanations would decrease the accuracy of the predictions. In this section, we demonstrate a simple experiment to validate our algorithm. Our experiment is performed as follows: 1) Generate the explanation of a targeted image $X$ via AXAI, where the explanation length $K=10$ is selected in this experiment, 2) Blur the top 5 explanations/segments of the targeted image according to the produced explanations, feed the modified image to the model and obtain its label, 3) repeat this process throughout the test dataset 4) Calculate the total decrease in accuracy. We use a ResNet34 training on ImageNet for this experiment and report the results for the entire ImageNet test dataset. Our results show that the prediction accuracy  of the DNN decreases to $\%43$ after blurring the top 5 explanation/segments. To further investigate, instead of blurring the top 5 explanations, we blur only the 6th to10th explanations. This results in a $\%22$  drop in total accuracy. Hence, we can conclude 1) AXAI generates faithful explanations so that blurring the top explanations (the 1st-5th explanations) lead to a strong decrease in model prediction accuracy, and 2) AXAI generates faithful explanations in order of importance, i.e., the generated 6th to 10th explanations are also important to the model but their influence on model predictions is relatively less than the first 5 generated explanations.
\subsection{AXAI explanations for a robust model trained with adversarial training}
\begin{figure}[htp]
	\centering
	\begin{subfigure}[t]{0.31\textwidth}
		\includegraphics[width=\linewidth]{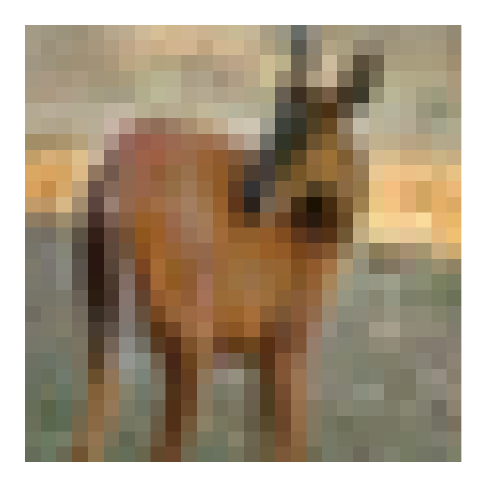} 
		\caption{Image of a deer} \label{fig:a16a}
	\end{subfigure}
	\quad
	\begin{subfigure}[t]{0.31\textwidth}
		\includegraphics[width=\linewidth]{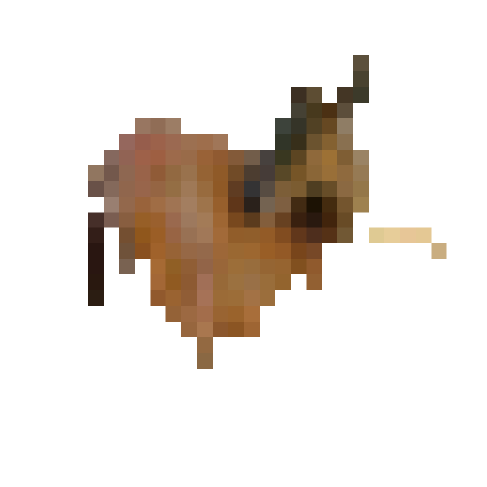} 
		\caption{Non-robust AlexNet} \label{fig:a16b}
	\end{subfigure}
	\quad
	\begin{subfigure}[t]{0.31\textwidth}
		\includegraphics[width=\linewidth]{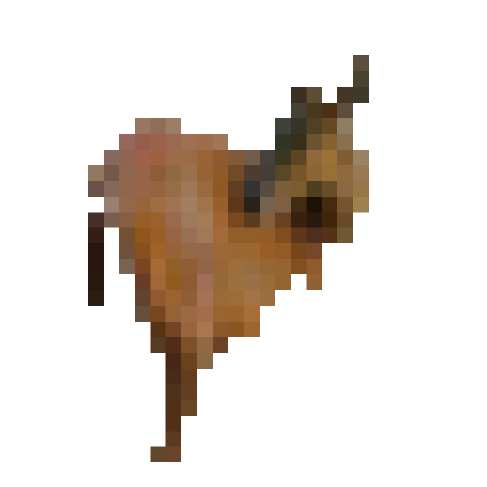} 
		\caption{Robust AlexNet} \label{fig:a16c}
	\end{subfigure}
	\caption{Comparison between explanations produced by AXAI ($K=10$) for AlexNet trained on CIFAR10 with and without adversarial training}  \label{fig:a16}
\end{figure}

In this subsection, we compare the explanations produced for a robust model to explanations produced for a non-robust model. In our experiment, a robust model is a model trained on an adversarial dataset in addition to the training dataset so that the final trained model is more robust against adversarial attacks. Hypothetically, a robust model should focus more on robust important input features when making predictions. We have trained a non-robust AlexNet and a robust AlexNet on CIFAR10 and produced the explanations using AXAI for test inputs. Fig. \ref{fig:a16} shows the AXAI produced explanations for the DNN given a sample input. It is seen that a small part of the background is included in the explanations produced for the non-robust AlexNet. However, the AXAI generated explanations for the robust model includes only the important features pertaining to the object in the image. In addition, the leg of the deer is now included in the explanations as well. It is concluded that explanations produced for the robust DNN are sharper, clearer and more robust than the ones generated for the regularly trained DNN.

\subsection{Additional Examples} 
\begin{figure}[hp]
	\centering
	\begin{subfigure}[t]{0.22\textwidth}
		\includegraphics[width=\linewidth]{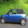} 
		\caption{An image of a car} \label{fig:21a}
	\end{subfigure}
	\quad
	\begin{subfigure}[t]{0.22\textwidth}
		\includegraphics[width=\linewidth]{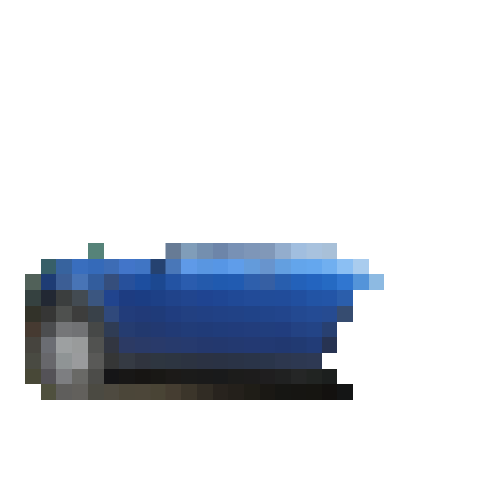} 
		\caption{Explanation} \label{fig:21b}
	\end{subfigure}
	\quad
	\begin{subfigure}[t]{0.22\textwidth}
		\includegraphics[width=\linewidth]{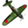} 
		\caption{An image of a plane} \label{fig:21c}
	\end{subfigure}
	\quad
	\begin{subfigure}[t]{0.22\textwidth}
		\includegraphics[width=\linewidth]{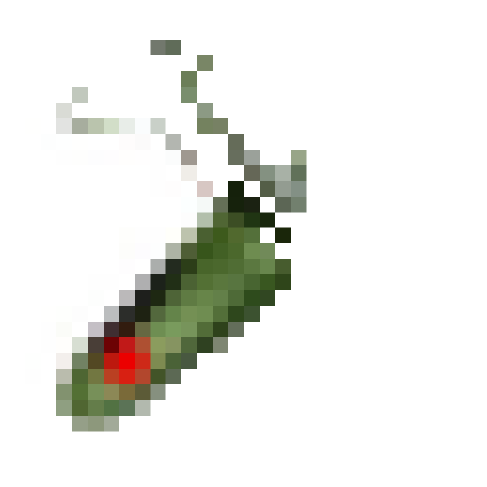} 
		\caption{Explanation} \label{fig:21d}
	\end{subfigure}
	
	\quad
	\begin{subfigure}[t]{0.22\textwidth}
		\includegraphics[width=\linewidth]{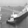} 
		\caption{An image of a ship} \label{fig:21e}
	\end{subfigure}
	\quad
	\begin{subfigure}[t]{0.22\textwidth}
		\includegraphics[width=\linewidth]{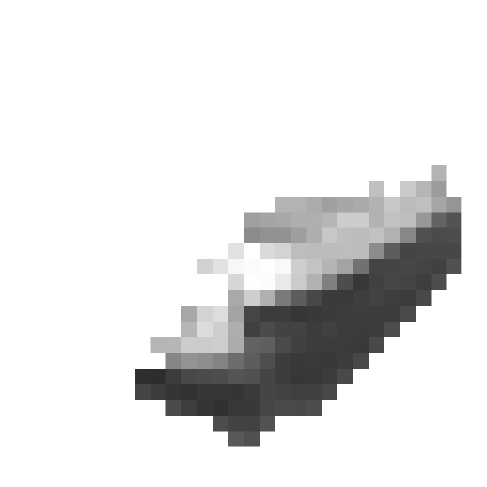} 
		\caption{Explanation} \label{fig:21f}
	\end{subfigure}
	\quad
	\begin{subfigure}[t]{0.22\textwidth}
		\includegraphics[width=\linewidth]{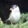} 
		\caption{An image of a bird} \label{fig:21g}
	\end{subfigure}
	\quad
	\begin{subfigure}[t]{0.22\textwidth}
		\includegraphics[width=\linewidth]{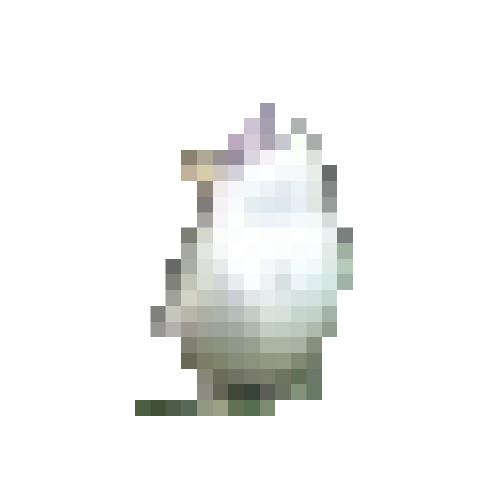} 
		\caption{Explanation} \label{fig:21h}
	\end{subfigure}
	
	\caption{Additional explainability results produced by AXAI for an AlexNet image classification model trained on CIFAR10.}  \label{fig:21}
\end{figure}

\begin{figure}[hp]
	\centering
	\begin{subfigure}[t]{0.22\textwidth}
		\includegraphics[width=\linewidth]{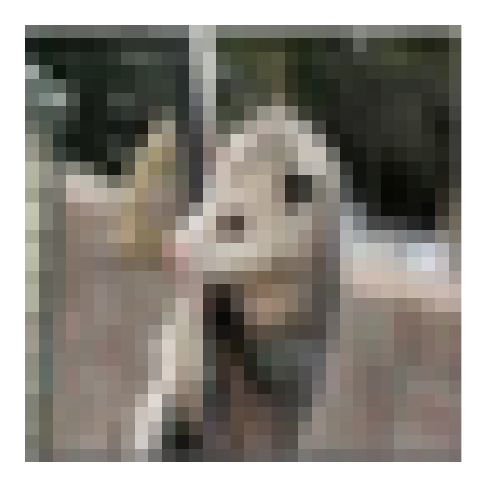} 
		\caption{An image of a possum} \label{fig:28a}
	\end{subfigure}
	\quad
	\begin{subfigure}[t]{0.22\textwidth}
		\includegraphics[width=\linewidth]{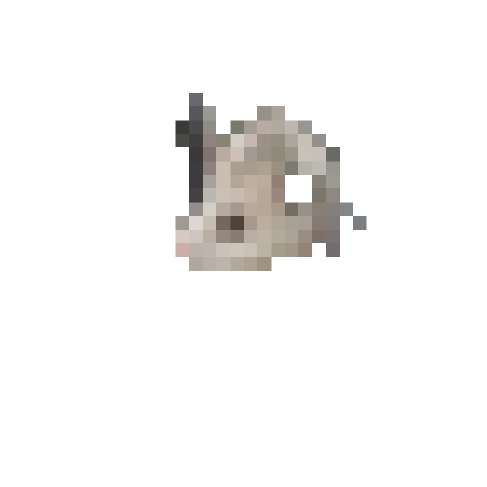} 
		\caption{Explanation} \label{fig:28b}
	\end{subfigure}
	\quad
	\begin{subfigure}[t]{0.22\textwidth}
		\includegraphics[width=\linewidth]{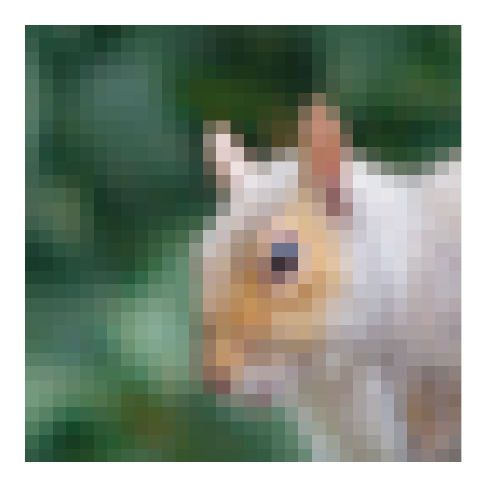} 
		\caption{An image of a squirrel} \label{fig:28c}
	\end{subfigure}
	\quad
	\begin{subfigure}[t]{0.22\textwidth}
		\includegraphics[width=\linewidth]{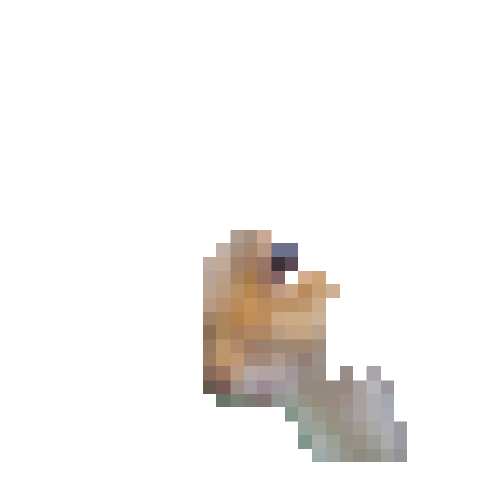} 
		\caption{Explanation} \label{fig:28d}
	\end{subfigure}
	
	\quad
	\begin{subfigure}[t]{0.22\textwidth}
		\includegraphics[width=\linewidth]{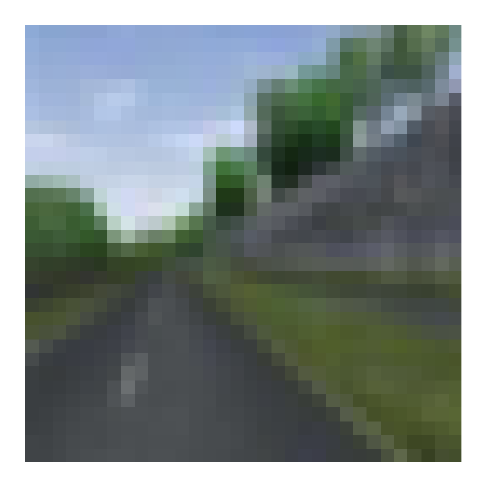} 
		\caption{An image of a road} \label{fig:28e}
	\end{subfigure}
	\quad
	\begin{subfigure}[t]{0.22\textwidth}
		\includegraphics[width=\linewidth]{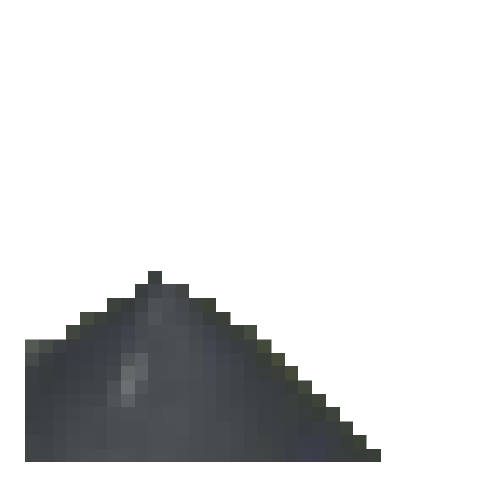} 
		\caption{Explanation} \label{fig:28f}
	\end{subfigure}
	\quad
	\begin{subfigure}[t]{0.22\textwidth}
		\includegraphics[width=\linewidth]{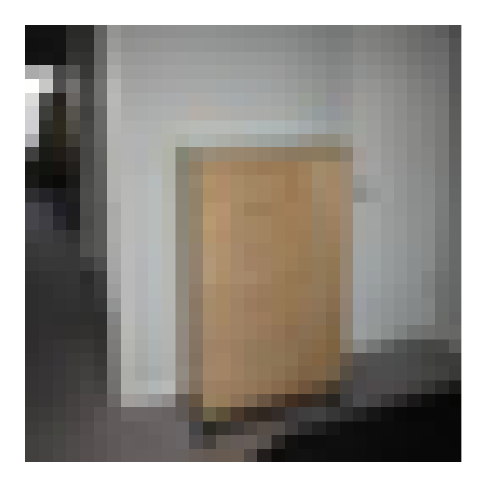} 
		\caption{An image of a wardrobe} \label{fig:28g}
	\end{subfigure}
	\quad
	\begin{subfigure}[t]{0.22\textwidth}
		\includegraphics[width=\linewidth]{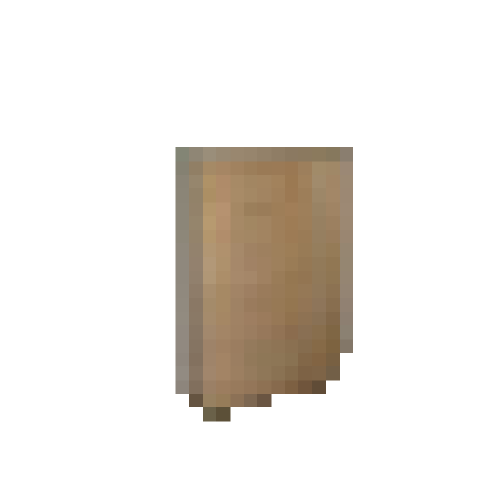} 
		\caption{Explanation} \label{fig:28h}
	\end{subfigure}
	
	\caption{Additional explainability results produced by AXAI  for a VGG16 image classification model trained on CIFAR100.}  \label{fig:28}
\end{figure}

\begin{figure}[htp]
	\centering
	\begin{subfigure}[t]{0.22\textwidth}
		\includegraphics[width=\linewidth]{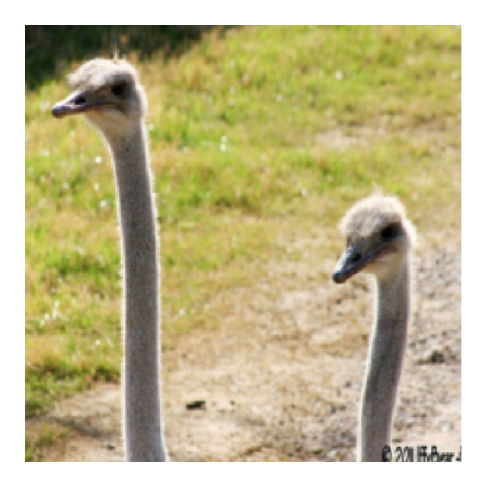} 
		\caption{An image of an ostrich} \label{fig:24a}
	\end{subfigure}
	\quad
	\begin{subfigure}[t]{0.22\textwidth}
		\includegraphics[width=\linewidth]{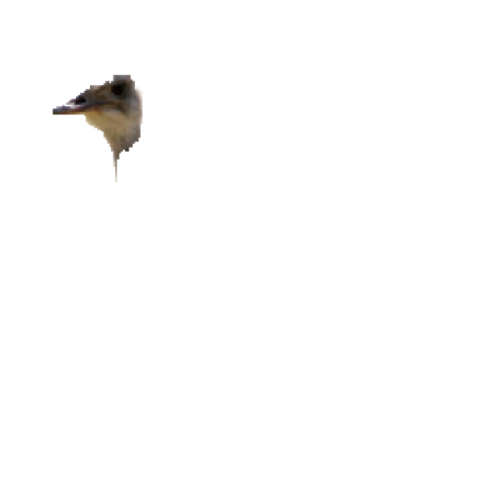} 
		\caption{Explanation} \label{fig:24b}
	\end{subfigure}
	\quad
	\begin{subfigure}[t]{0.22\textwidth}
		\includegraphics[width=\linewidth]{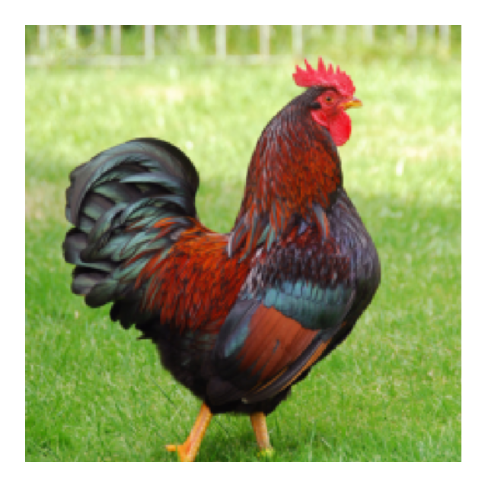} 
		\caption{An image of a cock} \label{fig:24c}
	\end{subfigure}
	\quad
	\begin{subfigure}[t]{0.22\textwidth}
		\includegraphics[width=\linewidth]{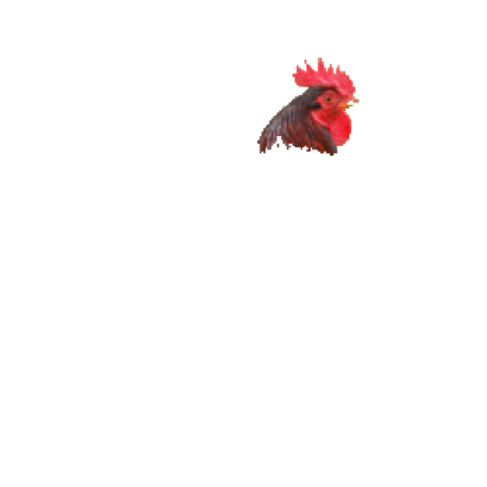} 
		\caption{Explanation} \label{fig:24d}
	\end{subfigure}
	
	\quad
	\begin{subfigure}[t]{0.22\textwidth}
		\includegraphics[width=\linewidth]{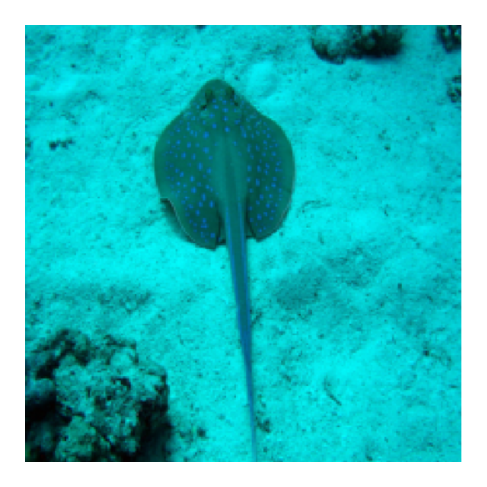} 
		\caption{An image of a stingray} \label{fig:24e}
	\end{subfigure}
	\quad
	\begin{subfigure}[t]{0.22\textwidth}
		\includegraphics[width=\linewidth]{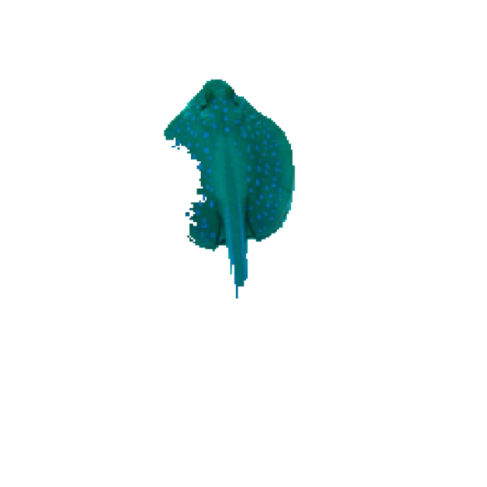} 
		\caption{Explanation} \label{fig:24f}
	\end{subfigure}
	\quad
	\begin{subfigure}[t]{0.22\textwidth}
		\includegraphics[width=\linewidth]{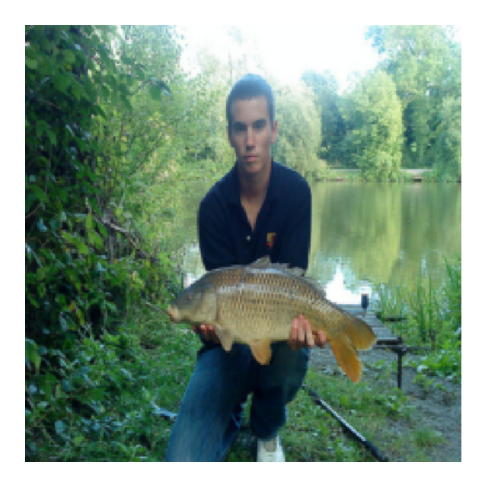} 
		\caption{An image of a tench} \label{fig:24g}
	\end{subfigure}
	\quad
	\begin{subfigure}[t]{0.22\textwidth}
		\includegraphics[width=\linewidth]{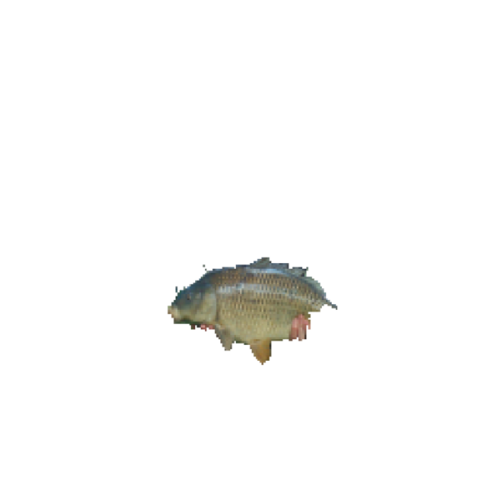} 
		\caption{Explanation} \label{fig:24h}
	\end{subfigure}
	
	\caption{Additional explainability results produced by AXAI for a ResNet34 image classification model trained on ImageNet.}  \label{fig:24}
\end{figure}

\begin{figure}[!]
	\centering
	\begin{subfigure}[!]{0.22\textwidth}
		\includegraphics[width=\linewidth]{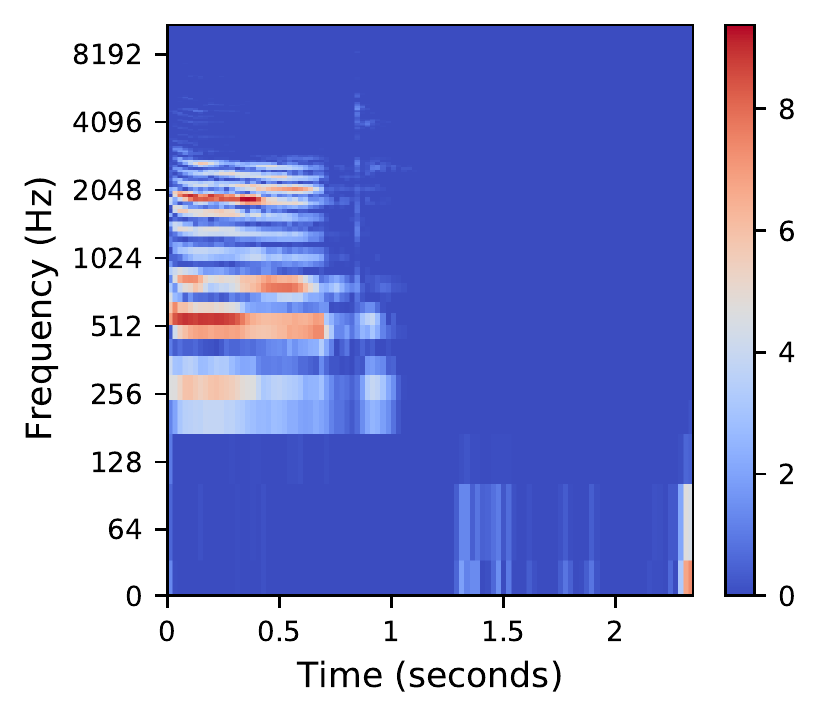} 
		\caption{"bird"} \label{fig:22a}
	\end{subfigure}
	\quad
	\begin{subfigure}[!]{0.22\textwidth}
		\includegraphics[width=\linewidth]{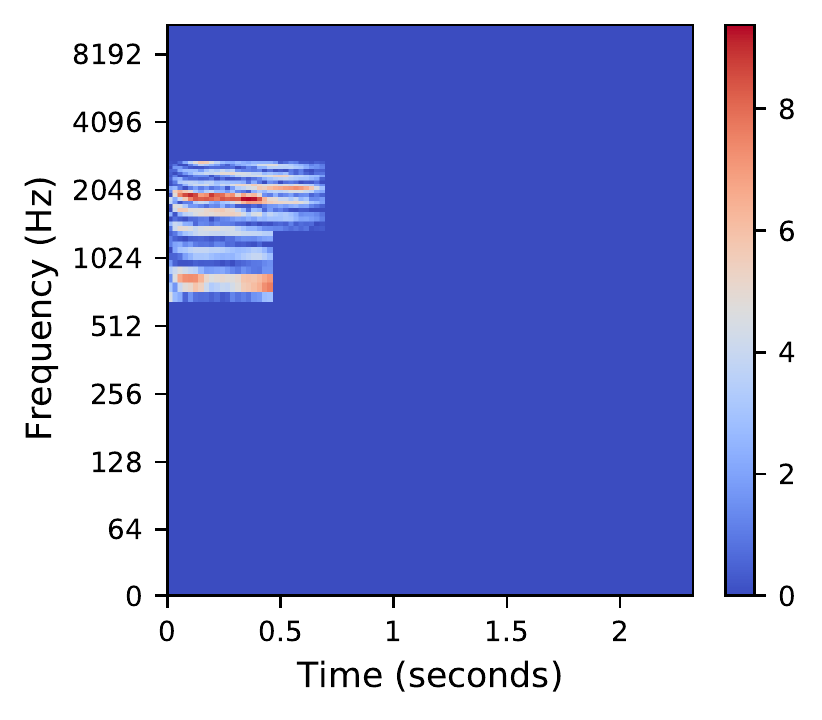} 
		\caption{Explanation} \label{fig:22b}
	\end{subfigure}
	\quad
	\begin{subfigure}[!]{0.22\textwidth}
		\includegraphics[width=\linewidth]{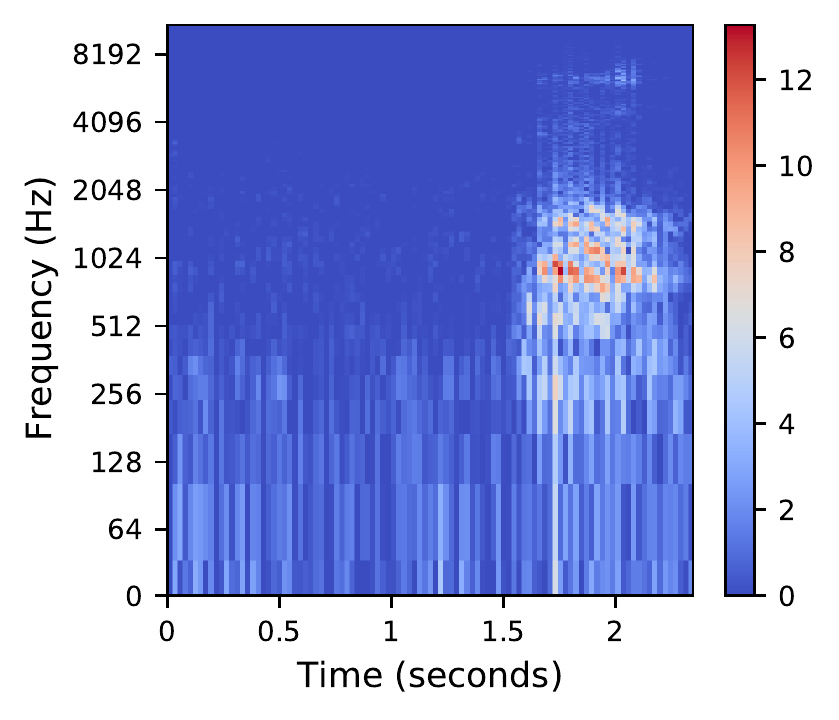} 
		\caption{"dog"} \label{fig:22c}
	\end{subfigure}
	\quad
	\begin{subfigure}[!]{0.22\textwidth}
		\includegraphics[width=\linewidth]{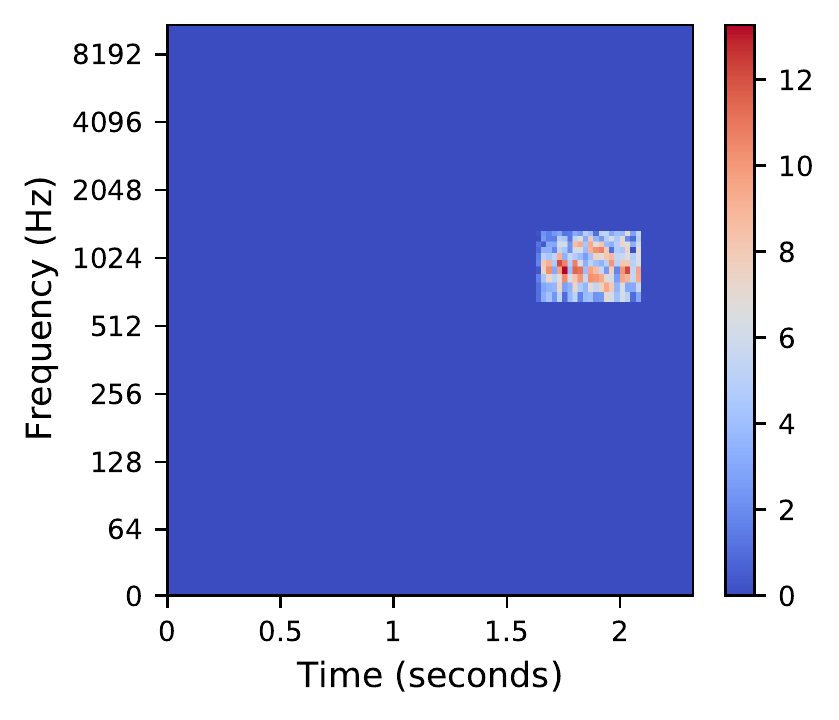} 
		\caption{Explanation} \label{fig:22d}
	\end{subfigure}
	\centering
	\begin{subfigure}[!]{0.22\textwidth}
		\includegraphics[width=\linewidth]{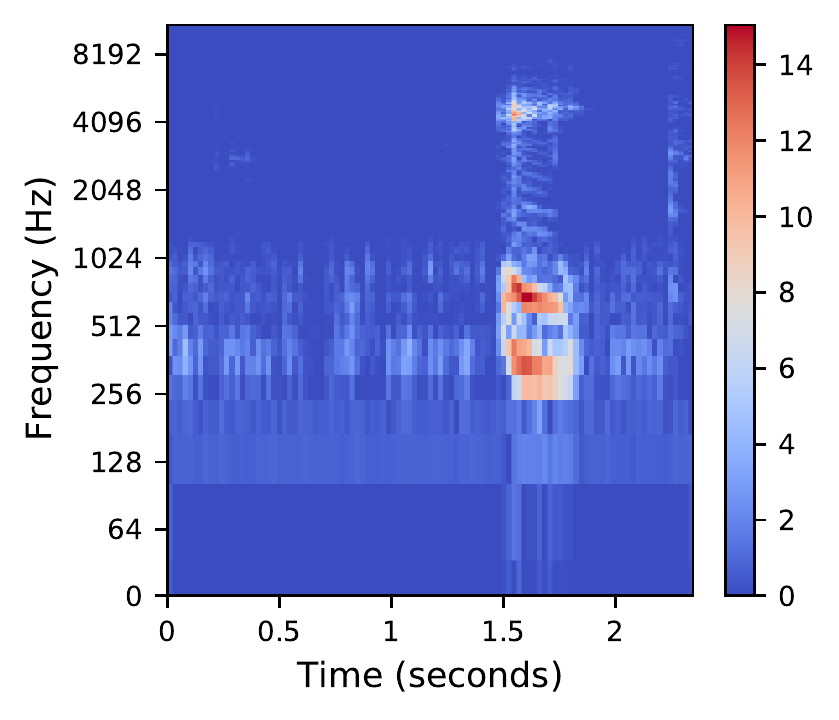} 
		\caption{"eight"} \label{fig:22e}
	\end{subfigure}
	\quad
	\begin{subfigure}[!]{0.22\textwidth}
		\includegraphics[width=\linewidth]{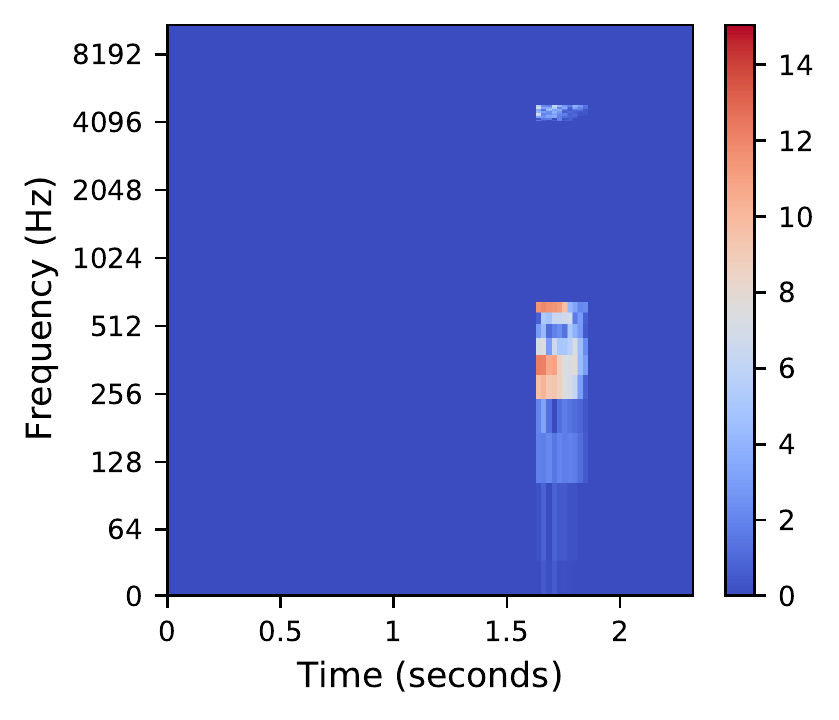} 
		\caption{Explanation} \label{fig:22f}
	\end{subfigure}
	\quad
	\begin{subfigure}[!]{0.22\textwidth}
		\includegraphics[width=\linewidth]{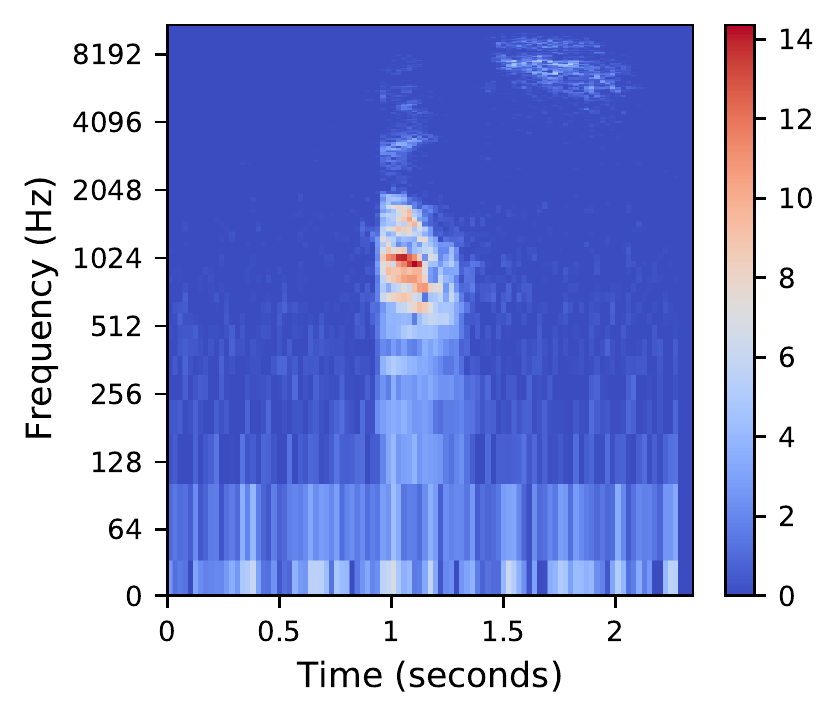} 
		\caption{"house"} \label{fig:22g}
	\end{subfigure}
	\quad
	\begin{subfigure}[!]{0.22\textwidth}
		\includegraphics[width=\linewidth]{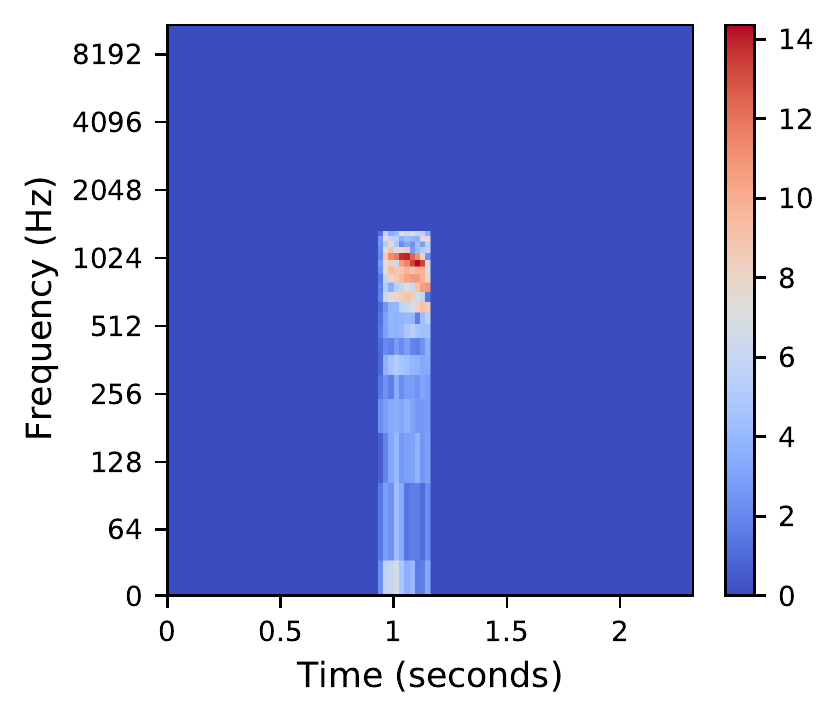} 
		\caption{Explanation} \label{fig:22h}
	\end{subfigure}
	
	\caption{Additional explainability results produced by AXAI for a LeNet speech recognition model trained on the Speech Commands Dataset.}  \label{fig:22}
\end{figure}

\begin{figure}[!]
	\centering
	\begin{subfigure}[!]{0.4\textwidth}
		\includegraphics[width=\linewidth]{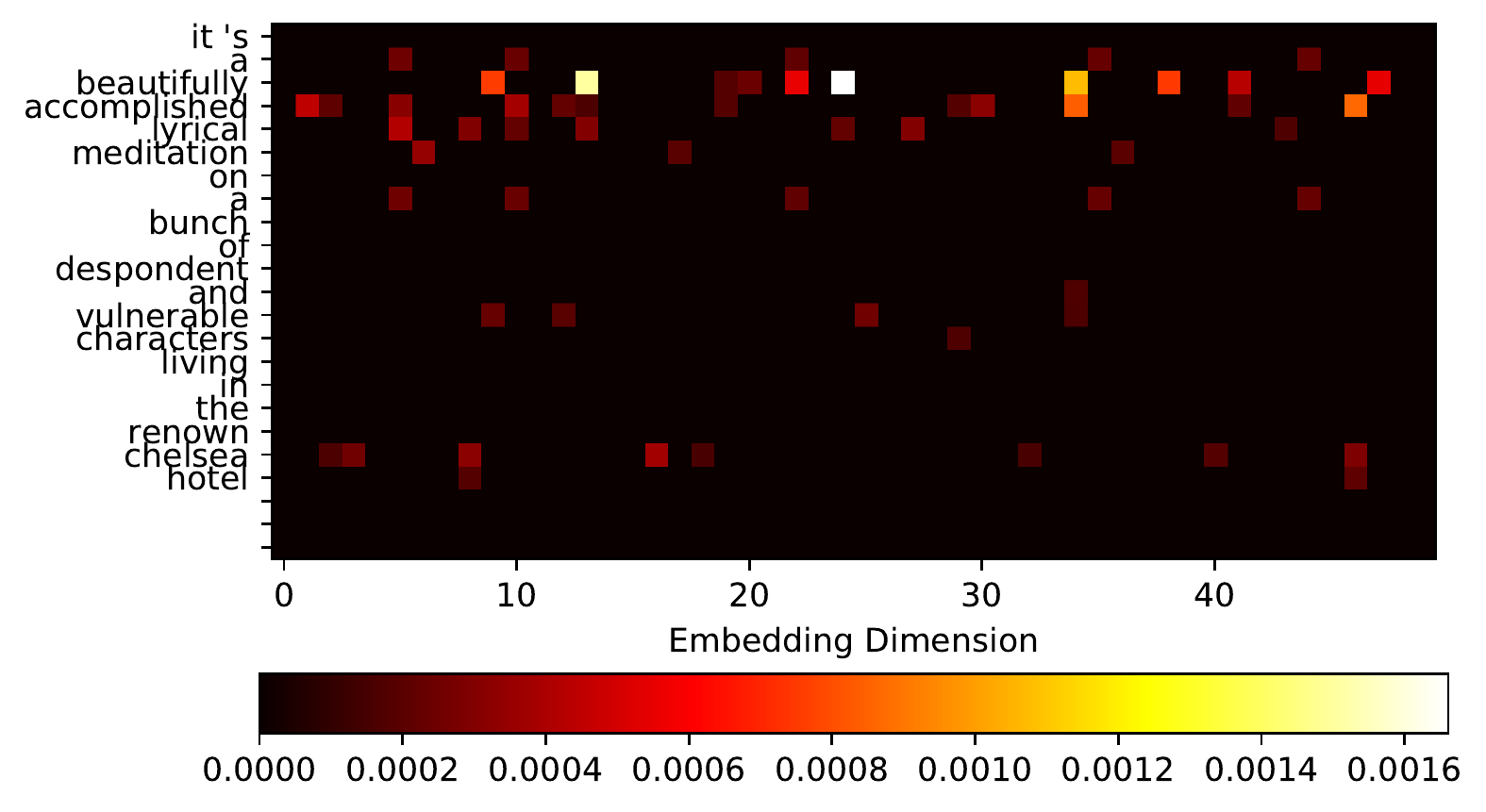} 
		\caption{Text example 1} \label{fig:23a}
	\end{subfigure}
	\quad
	\begin{subfigure}[!]{0.4\textwidth}
		\includegraphics[width=\linewidth]{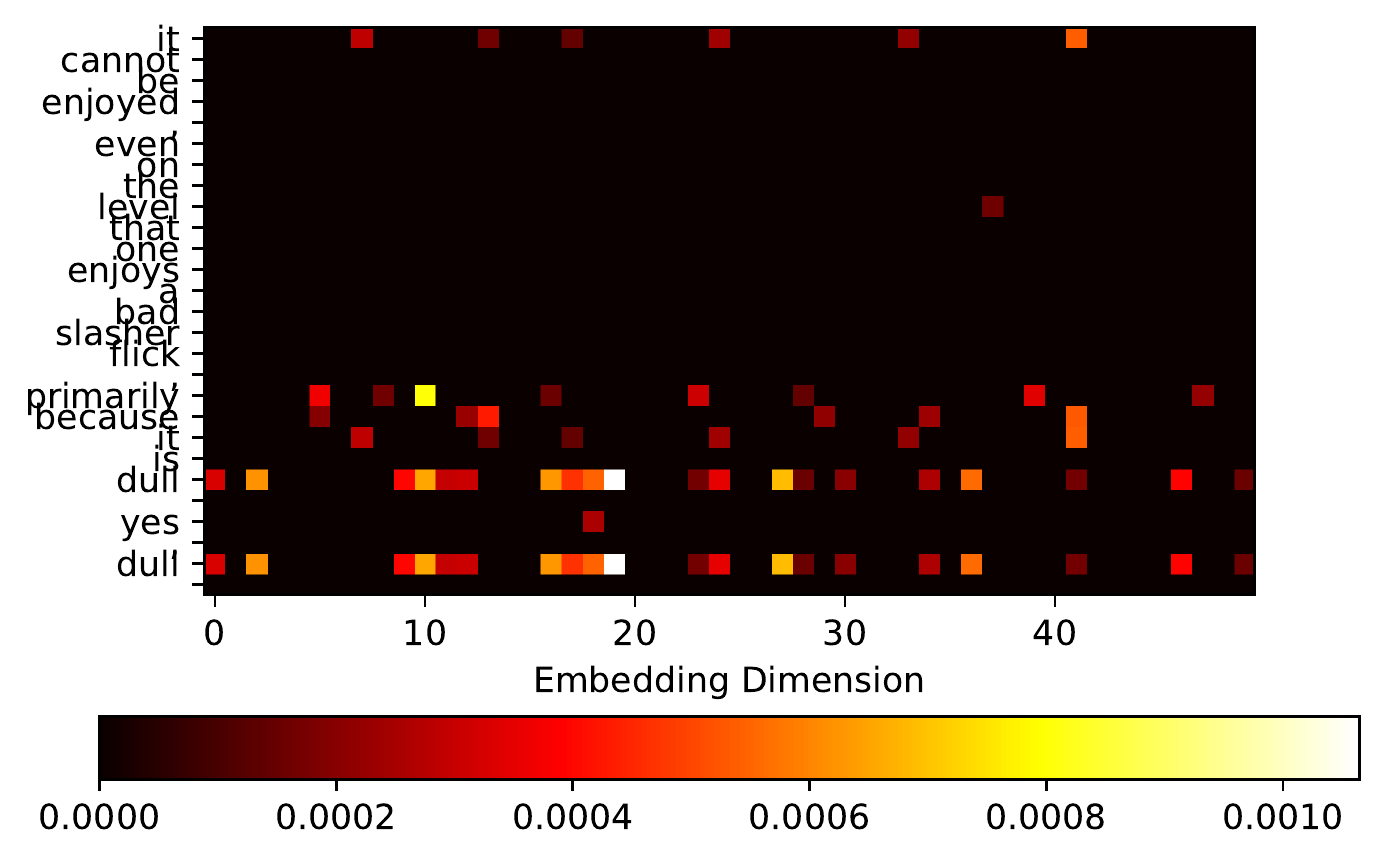} 
		\caption{Text example 2} \label{fig:23b}
	\end{subfigure}
	\quad
	\begin{subfigure}[!]{0.4\textwidth}
		\includegraphics[width=\linewidth]{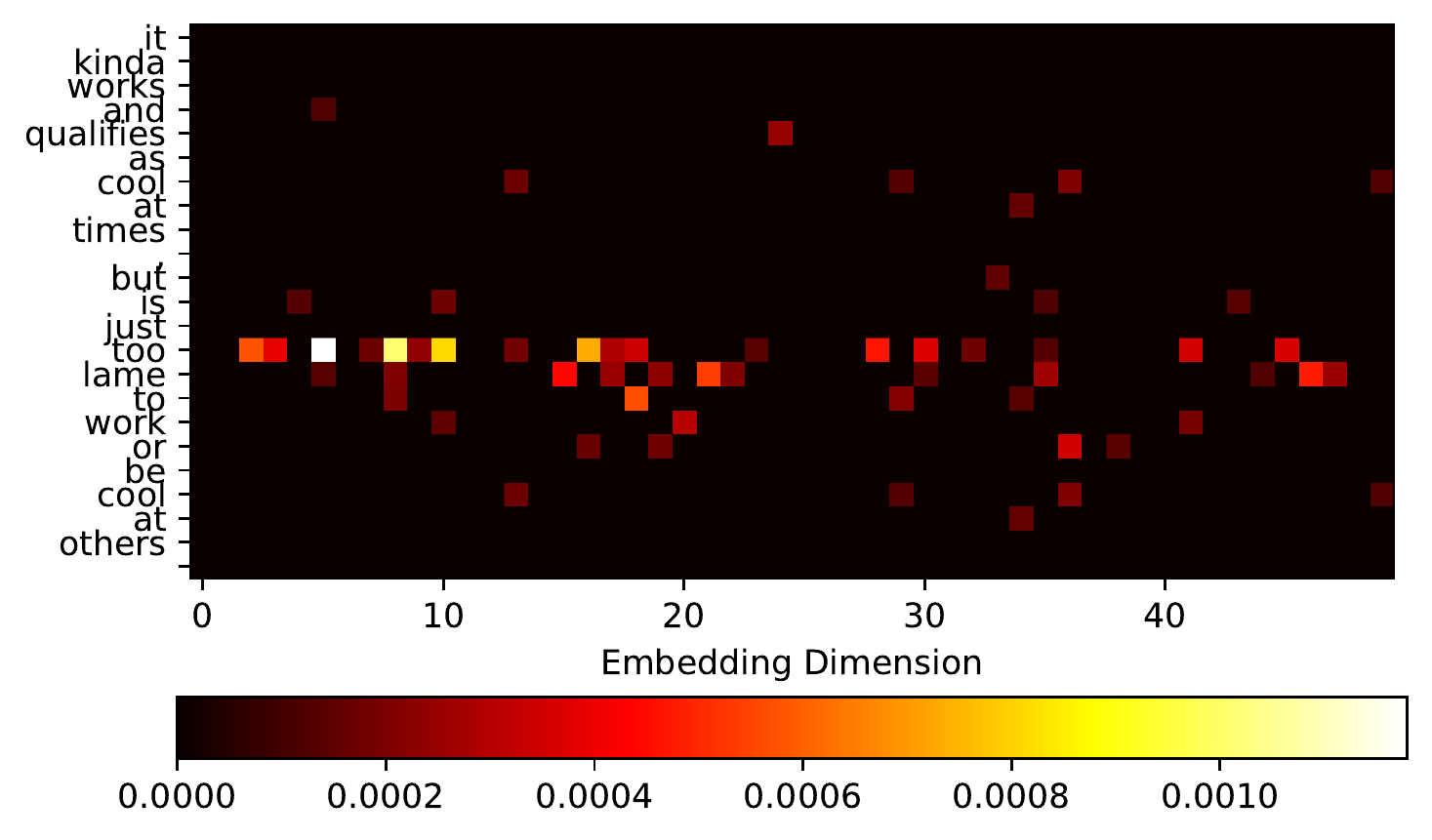} 
		\caption{Text example 3} \label{fig:23c}
	\end{subfigure}
	\quad
	\begin{subfigure}[!]{0.4\textwidth}
		\includegraphics[width=\linewidth]{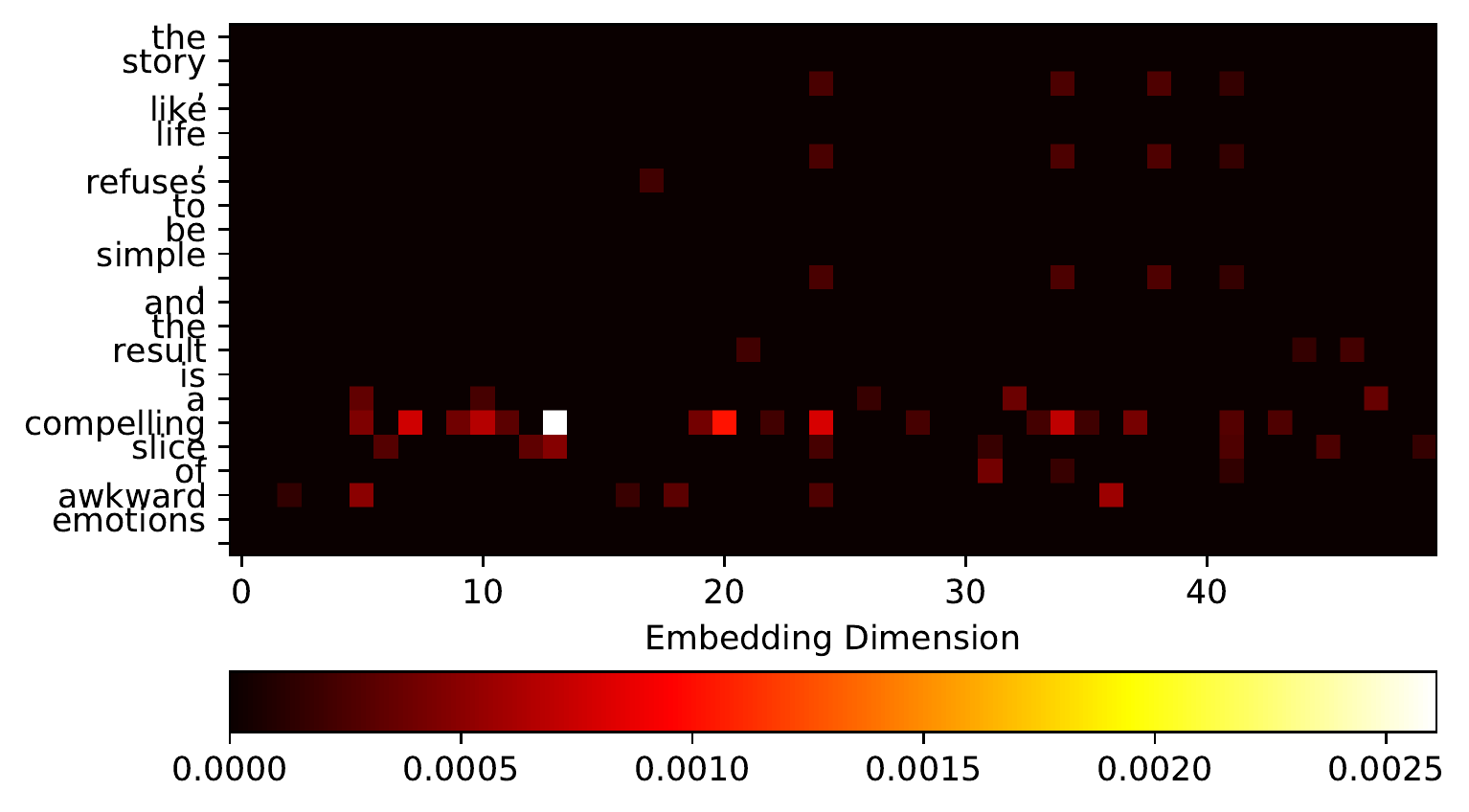} 
		\caption{Text example 3} \label{fig:23d}
	\end{subfigure}
	
	\caption{Additional explainability results produced by AXAI for a sentence classification model trained on the Sentence Polarity Dataset.}  \label{fig:23}
\end{figure}

In this section, we provide additional explainability  results from using AXAI on an AlexNet image classification model trained on CIFAR10, a VGG16 image classification model trained on CIFAR100, a ResNet34 image classification model trained on ImageNet, the LeNet speech recognition model and the sentence classification model, Fig. \ref{fig:21},  Fig. \ref{fig:28}, Fig. \ref{fig:24}, Fig. \ref{fig:22}, and Fig. \ref{fig:23}. 


\section{Benchmark Tests} \label{app_6}
\begin{figure}[htp]
	\centering
	\begin{subfigure}[t]{0.26\textwidth}
		\includegraphics[width=\linewidth]{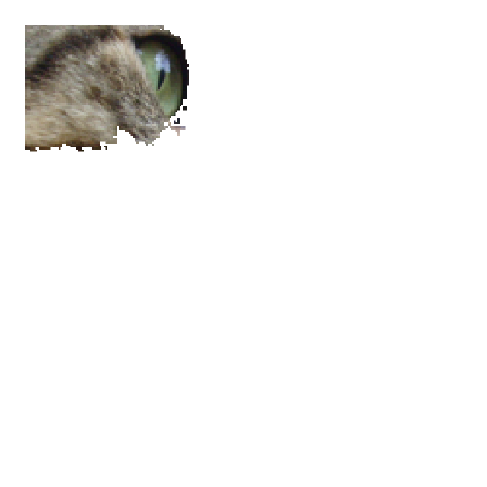} 
		\caption{} \label{fig:20a}
	\end{subfigure}
	\quad
	\begin{subfigure}[t]{0.26\textwidth}
		\includegraphics[width=\linewidth]{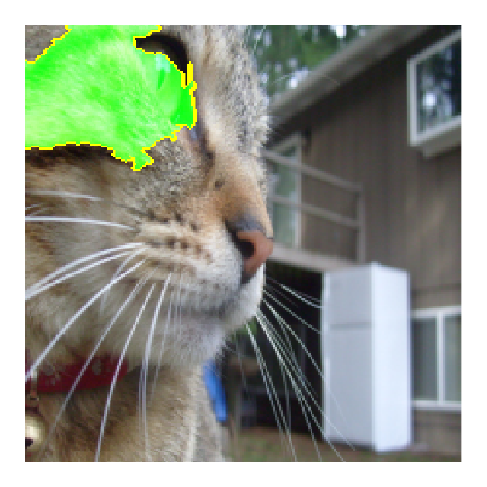} 
		\caption{} \label{fig:20b}
	\end{subfigure}
	\quad
	\begin{subfigure}[t]{0.26\textwidth}
		\includegraphics[width=\linewidth]{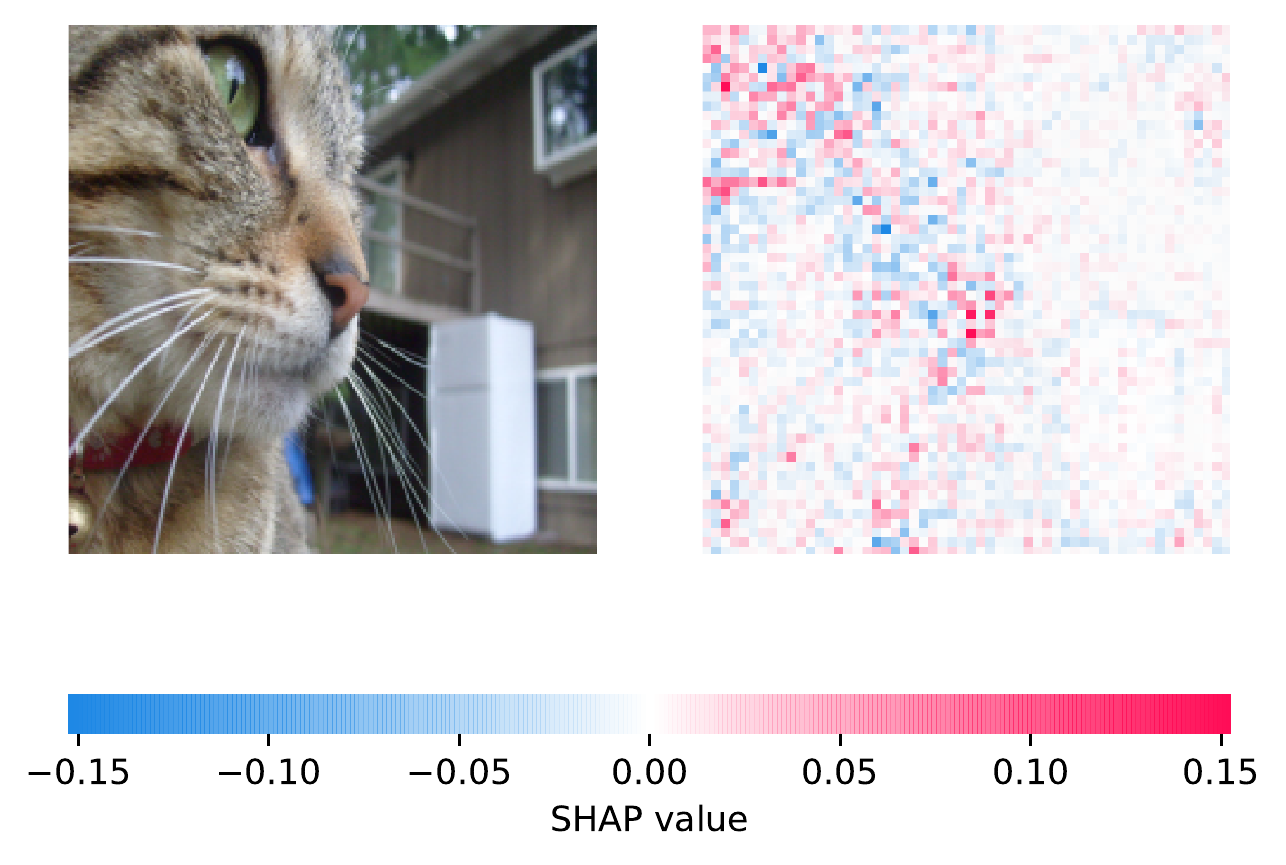} 
		\caption{}  \label{fig:20c}
		
	\end{subfigure}
	\begin{subfigure}[t]{0.26\textwidth}
		\includegraphics[width=\linewidth]{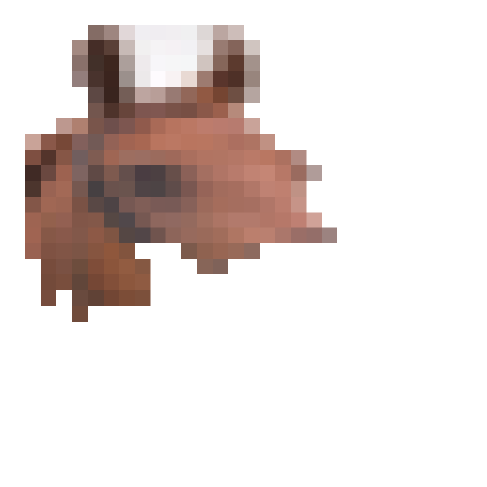} 
		\caption{} \label{fig:20d}
	\end{subfigure}
	\quad
	\begin{subfigure}[t]{0.26\textwidth}
		\includegraphics[width=\linewidth]{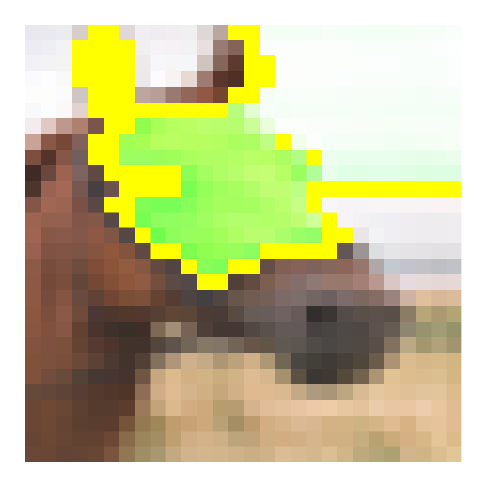} 
		\caption{} \label{fig:20e}
	\end{subfigure}
	\quad
	\begin{subfigure}[t]{0.26\textwidth}
		\includegraphics[width=\linewidth]{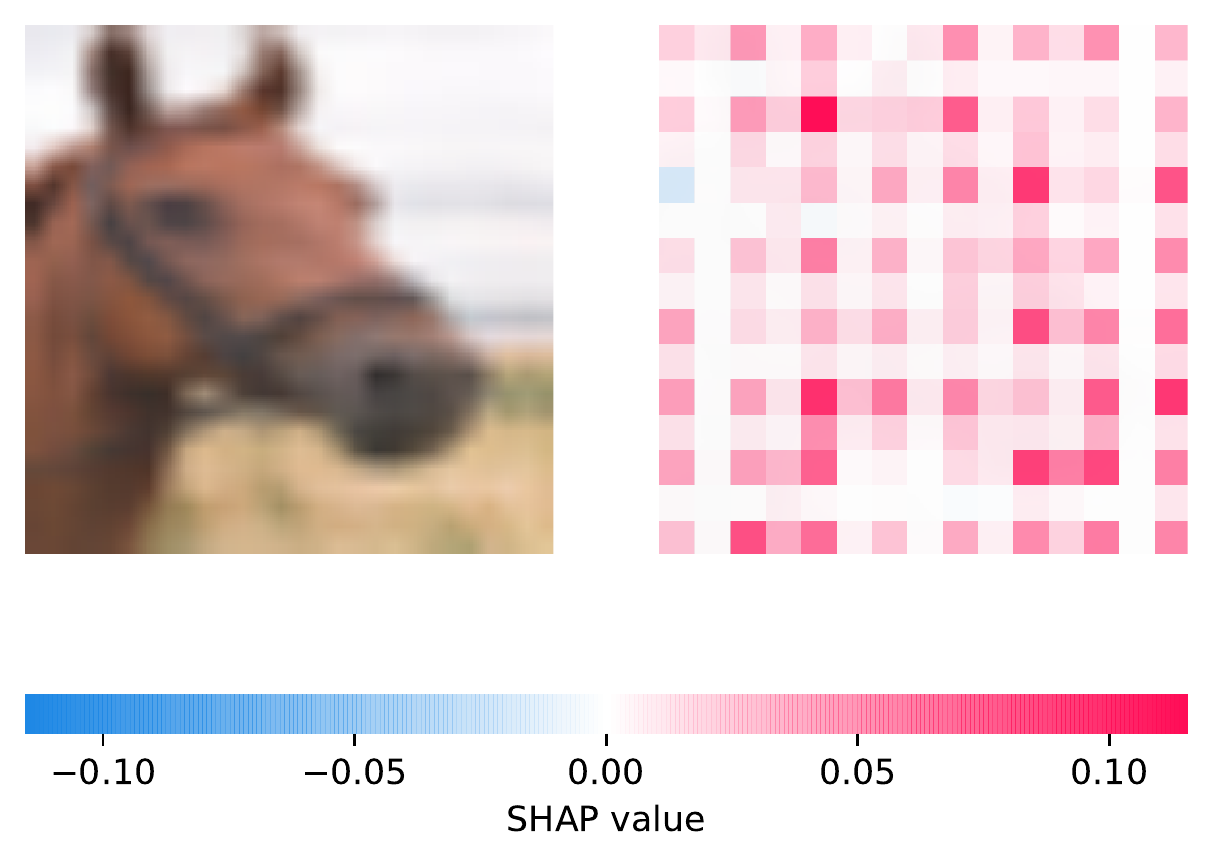} 
		\caption{} \label{fig:20f}
	\end{subfigure}
	\centering
	\begin{subfigure}[t]{0.26\textwidth}
		\includegraphics[width=\linewidth]{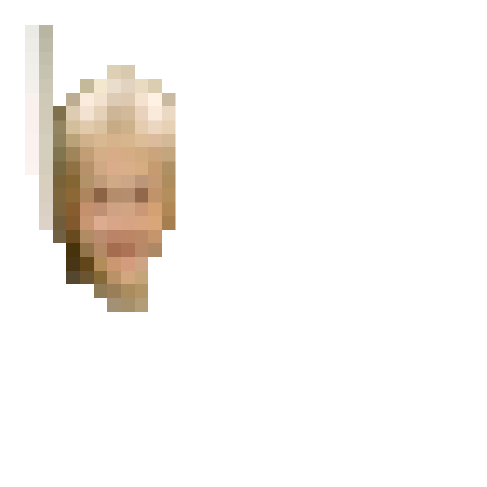} 
		\caption{} \label{fig:20g}
	\end{subfigure}
	\quad
	\begin{subfigure}[t]{0.26\textwidth}
		\includegraphics[width=\linewidth]{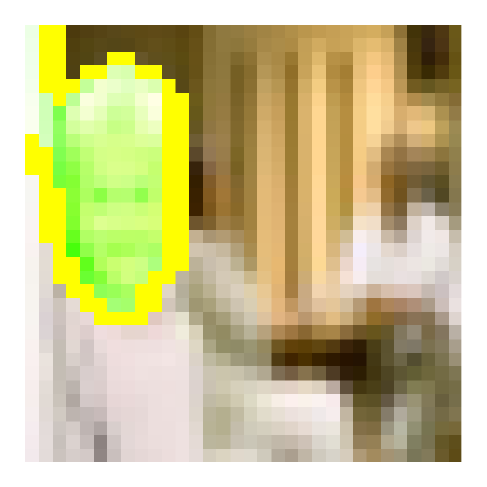} 
		\caption{} 	\label{fig:20h}
	\end{subfigure}
	\quad
	\begin{subfigure}[t]{0.26\textwidth}
		\includegraphics[width=\linewidth]{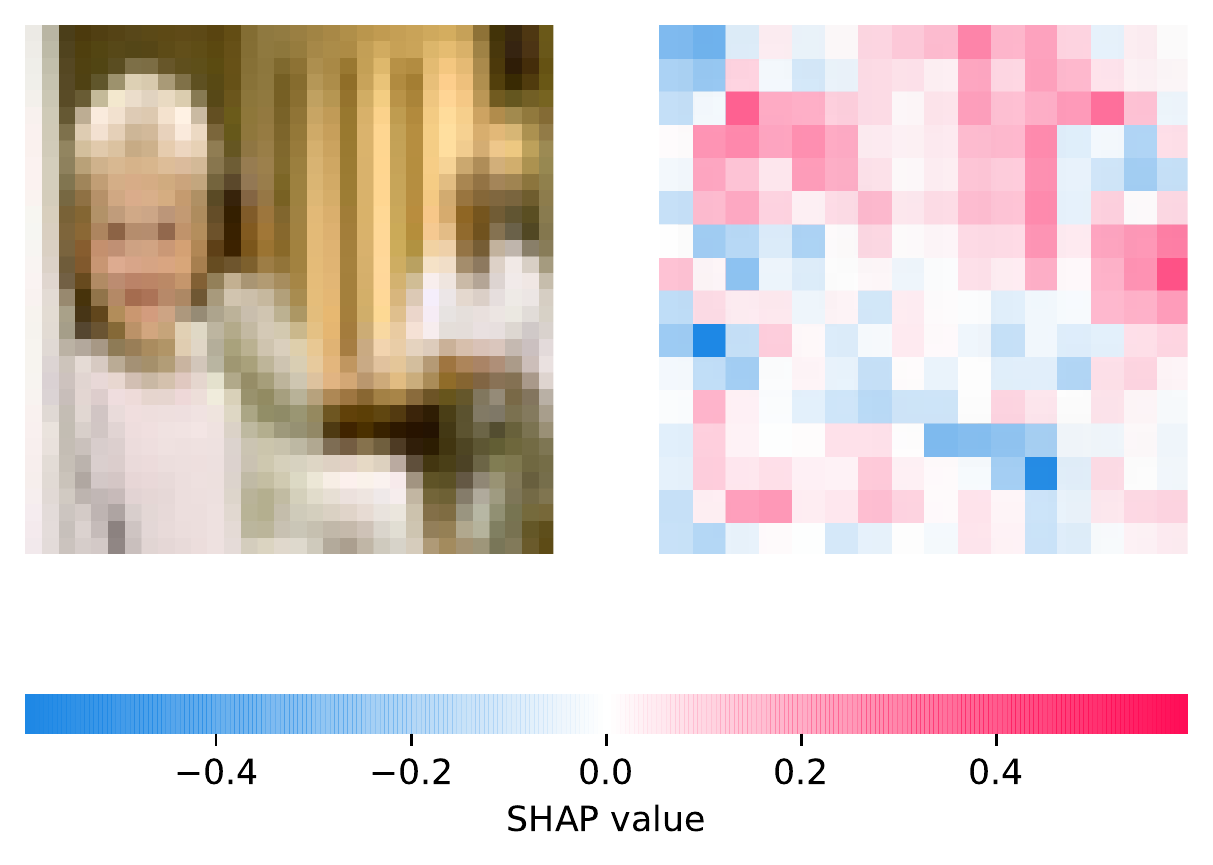} 
		\caption{} \label{fig:20i}
	\end{subfigure}
	\caption{Comparisons between our adversarial explainability approach (Left Column), LIME (Middle Column), and SHAP (Right Column). LIME parameters: number of perturbed samples $N=1000$, number of features $M=5$. First row: ResNet34 trained on ImageNet, Second row: AlexNet trained on CIFAR10, Third row: VGG16 trained on CIFAR100}  \label{fig:20}
\end{figure}

We test our algorithm against LIME and SHAP. We use ``Gradient Explainer" in SHAP, which integrates the f Integrated gradients algorithm with SHAP. Fig. \ref{fig:20} shows some sample comparisons among the 3 algorithms for 3 cases: 1) AlexNet trained on CIFAR10, 2) ResNet34 trained on ImageNet, 3) VGG16 trained on CIFAR100. PGDM with 20 iterations is used in our algorithm. For ImageNet, explanations for a sample test picture belonging to ``Egyptian cat" are shown in Fig. \ref{fig:20a}, Fig. \ref{fig:20b}, and Fig. \ref{fig:20c}. One can see the similarity between the explanations. The explanations produced by the 3 algorithms focused on the upper left of the image which contains the eyes of the ``Egyptian cat."  Both LIME and our algorithms point to the same segment as explanations. SHAP (Gradient Explainer) locates pixels of interest. The important pixels shown in this case aligns with the results of LIME and AXAI. Since the default image segmentation parameters LIME chooses do not allow for a suitable number of segments for explanation for CIFAR10 and CIFAR100 due to the resolutions of images, we lowered the Kernel size parameter to 1. The default Kernel size parameters LIME uses for QuickShift is too large for low-resolution images. As we mentioned before, this leads to a few very large segments in the image and neglects all the granular details in the image. For CIFAR10, both our approach and LIME capture the upper portion of the head of the horse including the ears and eyes (Fig. \ref{fig:20d},  Fig. \ref{fig:20e}). The results of SHAP point out the important pixels located on the head, the nose and some pixels in the background (Fig. \ref{fig:20f}). For CIFAR100, the explanations produced by the 3 algorithms are once again highly similar (Fig. \ref{fig:20g},  Fig. \ref{fig:20h}, and  Fig. \ref{fig:20i}). One can see that in many cases, pixel explanations do not serve as the best solution. Without the segments, it is hard to grasp the meaning behind explanations, this is because the human brain tends to comprehend image segments better than individual pixels. 

\end{document}